\PassOptionsToPackage{usenames,dvipsnames}{xcolor}
\documentclass[10pt,twocolumn,letterpaper]{article}

\usepackage[pagenumbers]{cvpr} 

\usepackage{graphicx}
\usepackage{amsmath}
\usepackage{amssymb}
\usepackage{booktabs}
\usepackage{bbding}
\usepackage{comment}
\usepackage{enumitem}
\usepackage{wrapfig}
\usepackage{xspace}
\usepackage{xargs}                      
\usepackage{graphicx}
\usepackage{multirow}
\usepackage{caption}
\usepackage{subcaption}
\usepackage{xcolor, soul}
\sethlcolor{green}

\newcommand{\DTD}{\textsc{DTD}}
\newcommand{\FGVC}{\textsc{FGVC}}
\newcommand{\CUB}{\textsc{CUB}}
\newcommand{\Flowers}{\textsc{Flowers}}
\newcommand{\Pets}{\textsc{Pets}}
\newcommand{\Cars}{\textsc{Cars}}
\newcommand{\Food}{\textsc{Food}}
\newcommand{\Imagenet}{\textsc{ImgNet-1k}}

\newcommand{\Oracle}{\textsc{Oracle}}
\newcommand{\GAC}{\textsc{GAC}}
\newcommand{\NNC}{\textsc{NNC}}
\newcommand{\LogitAvg}{\textsc{Log-Avg}}
\newcommand{\Best}{\textsc{Best}}
\newcommand{\VoteOne}{\textsc{Vote T-1}}
\newcommand{\VoteThree}{\textsc{Vote T-3}}
\newcommand{\Confidence}{\textsc{Conf}}

\newcommand{\CalibratedConfidence}{\textsc{C-Conf}}
\newcommand{\CLogitAvg}{\textsc{C-Log-Avg}}

\newcommand{\UN}{\textsc{UN}}
\newcommand{\DN}{\textsc{DN}}
\newcommand{\LTwo}{$\textsc{L}_2$}

\usepackage{amsmath,amsfonts,bm}

\def\eqref#1{equation~\ref{#1}}

\def\1{\bm{1}}

\def\vt{{\bm{t}}}

\def\vz{{\bm{z}}}

\def\evz{{z}}

\DeclareMathAlphabet{\mathsfit}{\encodingdefault}{\sfdefault}{m}{sl}
\SetMathAlphabet{\mathsfit}{bold}{\encodingdefault}{\sfdefault}{bx}{n}

\def\sB{{\mathbb{B}}}
\def\sC{{\mathbb{C}}}

\def\sL{{\mathbb{L}}}

\def\sR{{\mathbb{R}}}

\def\sX{{\mathbb{X}}}
\def\sY{{\mathbb{Y}}}

\usepackage[pagebackref,breaklinks,colorlinks]{hyperref}

\usepackage[capitalize]{cleveref}
\crefname{section}{Sec.}{Secs.}
\Crefname{section}{Section}{Sections}
\Crefname{table}{Table}{Tables}
\crefname{table}{Tab.}{Tabs.}

\def\cvprPaperID{5319} 
\def\confName{CVPR}
\def\confYear{2024}

\newcommand{\upg}[1]{{\scriptsize \color{ForestGreen}$\uparrow$#1}}
\newcommand{\downr}[1]{{\scriptsize \color{BrickRed}$\downarrow$#1}} 
\newcommand{\samey}[1]{{\scriptsize \color{Dandelion}$-$#1}} 

\makeatletter
\newcommand{\xfill}[1]{%
  \leavevmode\leaders\hrule height \dimexpr.7ex+#1\relax depth -.7ex\hfill\kern\z@%
}
\makeatother

\usepackage{eqparbox,collcell}

\begin{document}

\title{Unveiling Backbone Effects in CLIP: \\ Exploring Representational Synergies and Variances}

\author{
Cristian Rodriguez-Opazo$^1$
\and
Edison Marrese-Taylor$^2$
\and
Ehsan Abbasnejad$^1$
\and
Hamed Damirchi$^1$
\and
Ignacio M. Jara$^3$
\and
Felipe Bravo-Marquez$^3$
\and
Anton van den Hengel$^{1,4}$\\
\\
\vspace{-14mm}
\and
$^1$Australian Institute for Machine Learning, University of Adelaide,
\and
\vspace{-8mm}
$^2$The University of Tokyo, AIST
$^3$University of Chile, CENIA \& IMFD
$^4$Amazon
}
\maketitle

\begin{abstract}
Contrastive Language-Image Pretraining (CLIP) stands out as a prominent method for image representation learning. Various neural architectures, spanning Transformer-based models like Vision Transformers (ViTs) to Convolutional Networks (ConvNets) like ResNets, are trained with CLIP and serve as universal backbones across diverse vision tasks.
Despite utilizing the same data and training objectives, the effectiveness of representations learned by these architectures raises a critical question. Our investigation explores the differences in CLIP performance among these backbone architectures, revealing significant disparities in their classifications. Notably, normalizing these representations results in substantial performance variations. 
Our findings showcase a remarkable possible synergy between backbone predictions that could reach an improvement of over 20\% through informed selection of the appropriate backbone. 
Moreover, we propose a simple, yet effective approach to combine predictions from multiple backbones, leading to a notable performance boost of up to 6.34\%. We will release the code for reproducing the results.
\end{abstract}

\vspace{-2mm}

\section{Introduction}
\label{sec:intro}
Large pre-trained models, also known as Foundation Models (FMs), are reshaping the landscape of learning, particularly in computer vision \cite{radford2021learning, he2022masked, kirillov2023segment, liu2023improved} and natural language processing \cite{devlin2018bert,brown2020language,Touvron2023LLaMAOA}. 
These models are trained through self-supervised and contrastive learning objectives on internet-scale data, eliminating the need for manual annotations \cite{jia2021scaling,schuhmann2022laion}. 
Contrastive Language Image Pre-training (CLIP)~\cite{radford2021learning} stands as a pioneering contribution in the ongoing research endeavors within this domain, establishing itself as the state-of-the-art across diverse downstream tasks. 
The training objective is to learn to align text and image representations in an internet-scale dataset. This versatile framework is applicable to a wide range of commonly used architectures, facilitating the training of general image representation learning models.
These representation learning models have significantly advanced the field by acting as \emph{backbones} for other applications, producing state-of-the-art performance in various cross-modal alignment tasks, including but not limited to zero-shot classification \cite{Zhai_2022_CVPR,li2022elevater,jia2021scaling}, cross-modal retrieval \cite{li2020unicoder,li2020oscar,yu2022coca}, and the evaluation of machine-generated captions \cite{lee2020vilbertscore,jiang2019tiger}.

Despite the extensive body of prior research on CLIP, there remains a notable gap when it comes to thoroughly examining these image {backbones}. While we find extensive work evaluating the backbones employed across various tasks to showcase the generalization capabilities of specific methods \cite{goldblum2023battle, li2022elevater, zhang2021tip, gao2021clip}, they typically exhibit general performance improvements when using larger models. This supports the prevailing perception that larger models inherently lead to better representations, implying that they not only learn the same patterns but also more of them. If that is true, increasing the size of the model alone is enough to improve generalization. However, even within the same family of backbones, architectural selections exhibit nuanced inductive biases, capturing distinct patterns by different representations~\cite{inductive_bias}. Nevertheless, there is a conspicuous absence in thoroughly examining the synergies and variances among these backbones as learned by CLIP, particularly considering their shared training dataset and learning objective.



In light of these challenges, this paper conducts a comprehensive empirical investigation to unravel the intricacies of backbones within the CLIP framework. Our qualitative and quantitative evaluations reveal a notable ``orthogonality" in the behaviour of these backbones. These backbones adeptly respond to diverse patterns in the input, manifested through different representations, and exhibit distinct confidence levels in their predictions. We observe that employing different normalizations on the representations obtained from these backbones results in varied performance, complementing previous studies \cite{clip-understanding,zhou2023test}. 

Based on these findings, we consider combining predictions from various backbones for the final predictions. To that end, we first consider a set of baseline methods, following \cite{guoCalibrationModernNeural2017}--motivated by calibration techniques \cite{platt1999probabilistic}-- and techniques used in ensemble methods \cite{lakshminarayanan2017simple,dietterich2000ensemble}. We illustrate that these methods do not consistently enhance generalization performance. Instead, we introduce a straightforward parametric approach, where a set of temperatures is learned to adjust the logits for each backbone. The inspiration for this temperature parameter stems from the observation of diverse performance outcomes when manipulating the logits in CLIP~\cite{clip-understanding}.

To determine these temperatures, we explore two approaches: (1) utilizing a genetic algorithm as a search strategy and identifying a dataset-specific value, and (2) employing a multi-layer perceptron (MLP) to learn and adjust them individually for each sample. We observe that MLP efficiently learns to predict these temperatures with few samples.
This method allows for the seamless integration of information from different backbones, contributing to improved predictive outcomes.

In summary, our contributions are as follows:
\begin{itemize}
\item We perform extensive experiments across eight datasets, revealing that the representations produced by CLIP backbones and their subsequent predictions exhibit a notable diversity rather than uniformity. For instance, we observe a slight majority, with just over half of the ImageNet test-set correct predictions aligning within the ViT family. The remaining instances only exhibit partial agreement with other backbones.

\item We underscore discerning the optimal backbone alone for each individual test instance can yield up to 13\% improvement in zero-shot accuracy.

\item We demonstrate $L_2$ normalization of the representations generally outperforms the alternatives for zero-shot classification, complementing the previous findings of \cite{zhou2023test}.

\item Leveraging these insights, we introduce a simple approach to aggregate predictions from multiple backbones, enhancing the performance of the best single model by up to 6.34\%. Our neural network combination (NNC) approach demonstrates remarkable adaptability, requiring as few as a single instance to outperform the best 
of CLIP backbones. 
This performance boost is achieved solely by employing a scalar per backbone.

\end{itemize}

\section{Related work}
\label{sec:related_work}

\textbf{Vision Foundation Models} have extended the paradigm of pre-training to encompass wide-ranging datasets, ranging from hundreds of millions to billions of images. This expansion was significantly influenced by the introduction of Vision Transformers (ViTs) \cite{dosovitskiy2020vit}, which have highlighted the feasibility of training Transformers \cite{NIPS2017_3f5ee243} on such massive datasets within the field of computer vision. Subsequently, various large-scale pre-training methods have surfaced in the domain of computer vision \cite{radford2021learning,yu2022coca,he2022masked}. Among these, a notable category of Vision Foundational Models is exemplified by CLIP~\cite{radford2021learning, schuhmann2022laion, gadre2023datacomp}, which specializes in aligning noisy image-text pairs extracted from web sources. 
The prominence of CLIP stems not only from its scalability but also from its capability to generate meaningful alignments with prompts that facilitate zero-shot classification \cite{gao2021clip,zhang2021tip,li2022elevater}.
There are multiple versions of CLIP trained on different datasets and backbones. Well-studied convolutional backbones like Residual Networks \cite{He_2016_CVPR} (ResNet) and ViTs \cite{dosovitskiy2020vit} with different architectures setup. The focus of our current work centers on studying the difference in image classification of each backbone used to train CLIP.

\textbf{Model Ensembling} Improving performance by combining the output of multiple models is a foundational technique in machine learning, with studies on their effectiveness dating back to at least three decades ago \cite{dietterich2000ensemble,bauer1999empirical,breiman1996bagging,lakshminarayanan2017simple} 
More recent work has shown that foundational ensembling techniques can also be combined with deep neural networks, leading to what is known as deep ensembles \cite{lakshminarayanan2017simple}. In this context, we find previous work showing that deep ensembles exhibit high accuracy under distribution shift \cite{ovadia2019can}, and that higher divergence in training methodology leads to uncorrelated errors and better ensemble accuracy 
In contrast to techniques utilizing the same model with varied random initializations to achieve orthogonal predictions, our approach involves combining predictions from pre-trained models with distinct backbones. These backbones inherently embody diverse biases and properties \cite{naseer2021intriguing,abello2021dissecting,hermann2020origins}, facilitating the characterization of different properties associated with each label. By leveraging the inherent variations among these pre-trained backbones, our methodology provides a nuanced and effective means of capturing and utilizing diverse label-related features.

\textbf{Backbone studies}  Goldblum et al. \cite{goldblum2023battle} undertake a comprehensive comparison involving a diverse array of widely utilized pretrained backbones and randomly initialized baselines. Their evaluation spans extensive downstream tasks, encompassing image classification across diverse domains such as natural, medical, and satellite images. Additionally, the study delves into tasks like object detection and segmentation, evaluates models for out-of-distribution generalization, and assesses performance in image retrieval. Compared to these study, our work focuses specifically on revealing the orthogonality in predictions among the backbones utilized in the CLIP framework for the task of image classification. We present a method that explores and leverages this orthogonality to combine predictions effectively, aiming to boost performance straightforwardly.



\section{Proposed Approach}
\label{sec:method}


CLIP is trained using a self-supervised objective to align the output (\ie, representations) obtained from both language and visual encoders, which we denote as $\phi_l$ and $\phi_v$ respective. This alignment serves a critical purpose, as we can map textual or visual inputs using their corresponding neural encoders to the same semantic space, assuming the outputs are normalized. 

One way to use these models is to consider the joint probability of image $x$ and its corresponding description in natural language $l$ is proportional to a compatibility score between them, which we compute as the inner product between the encoded inputs, as Equation \ref{equation:score} shows below.
\begin{equation}
    \text{score}(x,l) = \phi_v(x)^\top \phi_l(l)
    \label{equation:score}
\end{equation}
To construct a probability distribution from these scores, we use the exponential function, thus allowing us to write the relationship shown below.
\begin{equation}
    p(x,l) \propto \exp\big(\phi_v(x)^\top \phi_l(l)\big)  
    \label{equation:jointprobs}
\end{equation}
In Equation \ref{equation:jointprobs}, $\sX$ and $\sL$ denote the visual and language spaces of possible inputs to the model, respectively.  Then, given a downstream multi-class classification task with the label set $\sY \subset \sL$ and $|\sY|=C$, we can write the following.
\begin{align}
    p(y \mid x) &= \frac{\exp\big(\phi_l(y)^\top \phi_v(x)\big)}{\sum_{y \in Y} \exp\big(\phi_l(y)^\top \phi_v(x)\big)} \label{eq:clip_prediction} \\
    \quad y^\star &= \arg\max_{y'\in Y} p(y' \mid x) \label{eq:clip_prediction_star}
\end{align}

\begin{table*}[ht!]
\centering
\resizebox{0.8\textwidth}{!}{%
\begin{tabular}{rcccccccccccc}
\toprule
\multirow{2}{*}{\bf Dataset} & \multicolumn{4}{c}{RN50} & \multicolumn{4}{c}{RN101} & \multicolumn{4}{c}{ViT-B-32} \\ 
\cmidrule(lr){2-5} \cmidrule(lr){6-9} \cmidrule(lr){10-13}
& \UN & \DN & \LTwo & \DN $+$ \LTwo & \UN & \DN & \LTwo & \DN $+$ \LTwo & \UN & \DN  & \LTwo & \DN $+$ \LTwo \\
\midrule
\Pets & 58.44 & 83.43 & \textbf{85.80} & 83.07 & 29.11 & 84.38 & \textbf{86.86} & 84.90 & 39.68 & 83.07 & \textbf{87.46} & 83.70 \\
\Cars & 48.59 & 53.67 & \textbf{54.22} & 53.00 & 50.29 & 59.26 & \textbf{61.12} & 60.73 & 47.23 & 57.18 & \textbf{59.73} & 58.05 \\
\CUB & 17.45 & 44.79 & 46.57 & \textbf{49.79} & 4.00 & 38.95 & 49.64 & \textbf{50.72} & 7.56 & 46.46 & \textbf{52.99} & 52.16 \\
\DTD & 34.36 & 40.43 & 41.22 & \textbf{41.38} & 27.39 & 41.33 & \textbf{43.67} & 42.98 & 34.57 & \textbf{45.96} & 43.99 & 43.03 \\
\FGVC & 10.89 & \textbf{17.49} & 17.07 & \textbf{17.49} & 9.12 & 18.33 & 18.63 & \textbf{19.23} & 14.04 & \textbf{19.74} & 19.65 & 19.32 \\
\Food & 63.40 & 74.48 & \textbf{77.91} & 77.23 & 52.77 & 74.23 & \textbf{81.86} & 80.96 & 64.48 & 78.13 & 82.58 & \textbf{82.80} \\
\Flowers & 28.04 & 63.41 & \textbf{66.12} & 64.35 & 0.89 & 48.71 & \textbf{65.20} & 63.12 & 20.07 & 62.43 & \textbf{66.48} & 64.40 \\
\Imagenet & 52.11 & 58.06 & \textbf{59.84} & 58.05 & 29.58 & 58.05 & \textbf{62.28} & 61.19 & 49.62 & 60.62 & \textbf{63.35} & 61.40 \\
\bottomrule
\multirow{14}{*}{}\\
\toprule
\multirow{2}{*}{\bf Dataset} & \multicolumn{4}{c}{ViT-B-16} & \multicolumn{4}{c}{ViT-L-14} & \multicolumn{4}{c}{\Oracle} \\ 
\cmidrule(lr){2-5} \cmidrule(lr){6-9} \cmidrule(lr){10-13}
& \UN & \DN & \LTwo & \DN $+$ \LTwo & \UN & \DN & \LTwo & \DN $+$ \LTwo & \UN & \DN  & \LTwo & \DN $+$ \LTwo \\ \midrule
\Pets & 52.96 & 86.32 & \textbf{89.07} & 87.14 & 81.93 & 92.15 & \textbf{93.59} & 91.88 & 87.24 & 97.47 & \textbf{98.06} & 95.88 \\
\Cars & 54.76 & 62.04 & \textbf{64.61} & 63.16 & 73.49 & 75.59 & \textbf{77.75} & 76.83 & 86.93 & 88.09 & \textbf{90.85} & 89.54 \\
\CUB & 16.86 & 49.10 & \textbf{55.28} & 55.25 & 45.22 & 60.44 & 62.06 & \textbf{62.55} & 54.49 & 76.01 & \textbf{81.20} & 79.36 \\
\DTD & 39.10 & \textbf{47.23} & 45.11 & 45.16 & 47.45 & \textbf{55.69} & 55.32 & 55.21 & 67.18 & \textbf{71.54} & 69.63 & 68.30 \\
\FGVC & 18.48 & 24.36 & 24.39 & \textbf{24.99} & 30.93 & \textbf{34.29} & 31.71 & 33.42 & 46.20 & \textbf{52.87} & 52.09 & 50.83 \\
\Food & 77.96 & 86.55 & \textbf{87.91} & 87.82 & 85.80 & 91.91 & 92.32 & \textbf{92.48} & 92.52 & 96.16 & 96.60 & \textbf{96.63} \\
\Flowers & 44.53 & 68.52 & \textbf{71.43} & 68.91 & 69.28 & 77.98 & \textbf{79.05} & 76.83 & 74.78 & 84.37 & \textbf{86.32} & 83.59 \\
\Imagenet & 57.18 & 65.78 & \textbf{68.34} & 66.83 & 72.01 & 74.37 & \textbf{75.54} & 74.20 & 80.68 & 84.25 & \textbf{85.30} & 83.76 \\ 
\bottomrule
\end{tabular}%
}
\caption{Zero-shot performance of CLIP backbones using different normalization techniques, Distribution Normalization (\DN), L2 Normalization (\LTwo), the combination of both (\DN $+$ \LTwo), as well as an unnormalized version (\UN). We also compare performance against upper-bound prediction if we could perfectly combine these backbones, denoted as \Oracle.}
\label{tab:zs-all-table}
\vspace{-3mm}
\end{table*}
\begin{figure*}[ht!]
    \centering
    \underline{\textbf{ImageNet-1K}}
    \includegraphics[width=0.99\textwidth]{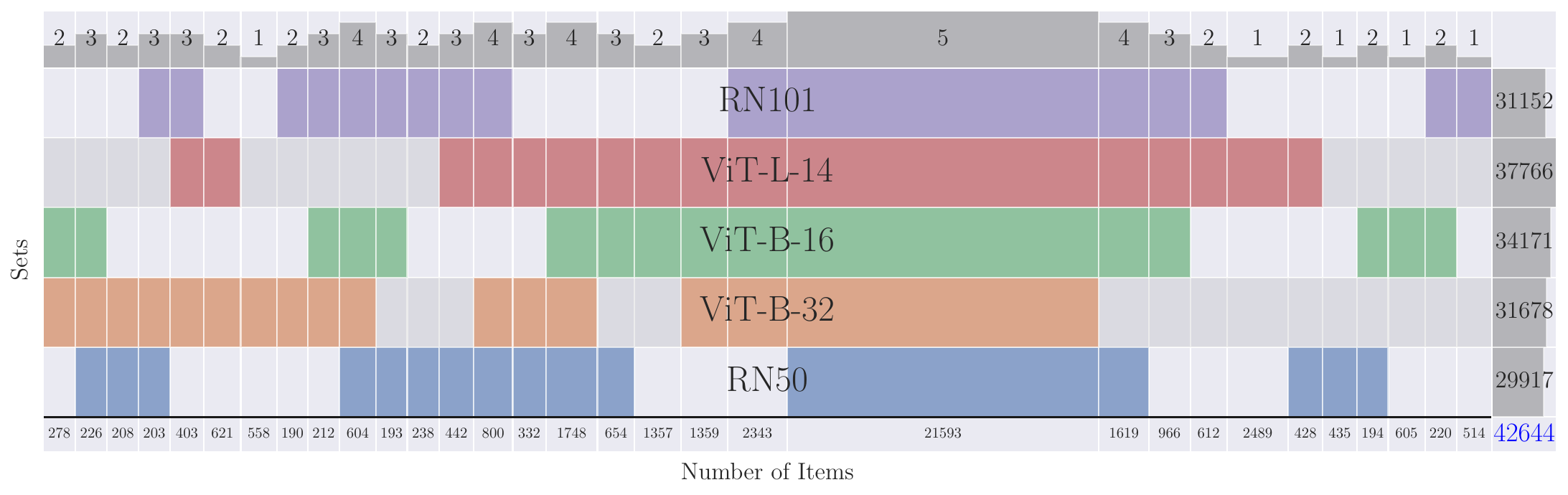}\vspace{-3mm}
    \caption{ImageNet-1k Venn diagram with the correct prediction of each backbone. The Top part of the Venn diagram shows the number of backbones that are predicting correctly a set of images. Each column represents a set of image instances that are predicted correctly by some group of backbones. Each row in the diagram shows in colour the backbone that correctly predicts a certain set of image instances, in grey when the backbone is not correctly predicting those instances. The bottom part of the Venn diagram shows the number of images in a certain set. The right part is the total amount of correctly predicted images per backbone.}
    \label{fig:venn_diagram}
    \vspace{-4mm}
\end{figure*}
Based on these principles, to perform zero-shot image classification using CLIP, we follow previous work and generate prompts that encapsulate the labels of the target dataset (e.g., \texttt{an image of \{label\}, which is a pet}). Subsequently, we calculate the similarity between the image feature and these prompts, and the prompt exhibiting the highest similarity is then chosen as the label for the given image. Importantly, we note that while performing zero-shot classification in this fashion, normalization defined in Equation \ref{eq:clip_prediction} is not required, which in practice means that the step is often not performed in order to reduce computational requirements.

Since our interest is to combine multiple CLIP backbones operating simultaneously in this fashion, our first key insight is to note that Equation \ref{eq:clip_prediction} is essentially equivalent to computing a softmax over a vector of logit-like values $\bar{\vz} \in \sR^{C}$, whose entries are constructed using on the inner product between a given input $x$ and the set of labels $\sY$. Concretely, let $\bar{\evz}_c$ denote the cth entry in vector $\bar{\vz}$, then we define the following.
\begin{equation}
    \bar{\evz}_c =  \phi_l(y_c)^\top \phi_v(x) \quad \forall c = [1, \ldots, C]
    \label{equation:logit-like}
\end{equation}
We can use the equation above to rewrite Equation \ref{eq:clip_prediction} simply as $p(y|x) = \text{softmax}(\bar{\vz})$, meaning that we can now essentially equate the output of our zero-shot CLIP classification pipeline, to that of a regular classifier, and interpret the output as a probability distribution over the label space $\sC$.




Since we are interested in combining backbones, we note that a key goal when constructing traditional model ensembles, a technique often used to combine different models, is to increase robustness by increasing the overall confidence of the predictions. In the supervised setting, confidence is generally measured as the probability associated with the predicted label. Though it is expected that a network should provide a calibrated confidence measure in addition to its prediction, it has been shown that deep neural nets are poorly confidence calibrated \cite{guoCalibrationModernNeural2017}.

In order to shed light into this issue for the case of CLIP models, we first perform an in-depth study of the output of 5 different single-backbone models when in zero-shot classification in 8 benchmark datasets (please see Section \ref{sec:setup} for details on backbones and datasets). Concretely, we propose to assess the ``\textit{orthogonality}'' of zero-shot predictions by means of two distinct approaches. Firstly, we establish an \Oracle{} prediction that is informed about the appropriate backbone to select for each image sample in the target dataset. While a more sophisticated combination of probabilities may yield enhanced performance, we consider this oracle prediction a pragmatic upper bound for evaluating the potential of combining these backbones. Secondly, we propose to employ Venn diagrams to visually analyze the discrepancies and overlaps in predictions, which allows us to qualitatively characterize the nature of the \textit{orthogonality} across models.

At this point, it is also important to highlight that the features extracted from the image and language encoders must first inhabit the same embedding space for effective comparison between prompts and the corresponding image. While different methods for doing this exist, a prevalent technique for achieving this alignment involves normalizing the features using the \LTwo-Norm{}.  However, Zhou et al. \cite{zhou2023test} recently argued that a discrepancy exists between the pre-training objective and the subsequent application in downstream tasks and proposed Distribution Normalization (\DN{}), which consists of removing the mean of the data to each of the features,
\begin{align}
    \label{eq:dn}
    \bar{\evz}^{\prime}_c = \Big(\phi_l(y_c)-\frac{1}{2}\mu_y\Big)^T\Big(\phi_v(x)-\frac{1}{2}\mu_x\Big)
\end{align}
where $\mu_x$ and $\mu_y$ are the means of a subset of the target dataset. 
We, therefore, also study the impact of different normalization techniques across our selected backbones and datasets, comparing against the option of not performing any normalization (\UN{}). 


Table \ref{tab:zs-all-table} summarizes the results of our quantitative results. We first see that while the necessity for normalization for aligning the vision and language spaces is evident, determining the optimal normalization for each backbone and dataset proves less straightforward. In contrast to findings by Zhou et al. \cite{zhou2023test}, our results indicate that, overall L2-Norm consistently demonstrates superior performance accuracy in our experimental setup. Despite the potential insights that could be derived from other properties of Distribution Normalizations \cite{zhou2023test}.

Furthermore, within the ResNet family, it is observed that, for the Flowers dataset, the deeper model, ResNet-101, exhibits inferior performance compared to ResNet-50. Conversely, the ViT models consistently demonstrate an improvement in performance with increasing model size, and ViT-L-14 emerges as the superior performing backbone across all datasets by a significant margin.


Finally, we see that our \Oracle{} consistently outperforms the best individual backbone. Specifically, we see that on  \FGVC, \CUB, and \Cars{} the oracle exhibits improvements of 20.38\%, 19.14\%, and 13.11\%, respectively, for the L2 Normalization cases. It is important to notice that regardless of the normalization method used, a distinct \textit{orthogonality} is evident when assessing the performance of the proposed \Oracle.

In qualitative terms, as shown in Figure \ref{fig:venn_diagram} via a Venn diagram with the correct predictions of each backbone on ImageNet-1k \cite{deng2009imagenet}, we see that ViT-L-14 accurately predicts 37,766 images from the test set, with 21,593 correctly predicted across all five backbones (please refer to the Supplementary Material for results of other datasets). Leveraging the informed \Oracle{} increases the overall correct predictions to 42,644 samples of 50,000. Notably, each backbone exhibits correct predictions that are unique to that specific backbone; RN50 and RN101 contribute 435 and 514 exclusive predictions, respectively. ViTs B-32, B-16, and L-14 present 558, 605, and 2,489 distinctive predictions, respectively. We highlight that predictions within each backbone family do not constitute a subset of the best model's predictions in each family

Based on these empirical findings, we hypothesize that we can effectively improve the classification performance by learning to combine CLIP backbones. Thus, we propose an approach that is directly inspired by previous work on network calibration, but extended to multiple model combinations. Concretely, our framework can be regarded as a set of variations of Platt scaling \cite{platt1999probabilistic}. In Platt scaling, the logits of a model are used as features for a logistic regression model, which is trained on the validation set to return probabilities. More concretely, given logit scores $z_i$ for an example $i$, Platt scaling learns scalar parameters $a, b \in \sR$ and outputs $\hat q_i = \sigma(a z_i + b)$ as the calibrated probability. Parameters $a$ and $b$ can be optimized using the NLL loss over the validation set, but during this time, the parameter of the original model are fixed. 

In this work, we adapt the simplest extension of Platt scaling, also known as temperature scaling \cite{guoCalibrationModernNeural2017}, to our backbone combination setting. In temperature scaling, a single scalar parameter $t > 0$ is set for all classes of a given model. The new, calibrated confidence prediction is given by Equation \ref{equation:temperature}, below, where $t$ is called the temperature, $z_i$ is the logits, for example $i$ returned by the uncalibrated model.
\begin{equation}
    \hat q_i = \max_k{\text{softmax}(z_i/t)^{(k)}}
    \label{equation:temperature}
\end{equation}

In the original setting, $t$ is optimized with respect to the NLL on the validation set aiming to reduce the overconfidence of the model on its predictions and produce more reliable predictions, but because the parameter $t$ does not change the maximum of the softmax function, the class prediction $\hat{y}_i$ remains unchanged, meaning that the performance of a given model remains the same. 

Different from the above, in this work, we aim to jointly optimize a set of temperature parameters $t_b$ with $b \in [1, \ldots, B]$ for a set of $B$ CLIP backbones aiming to combine the predictions of the backbones to produce a final prediction that change the confidence of each backbone depending on the confidence of the others and the input. To the best of our knowledge, we are unaware of any prior use in the context of calibrating mixtures of probabilistic models such as the one proposed above. Thus, we aim to improve the classification performance by learning the temperatures $t_b$ that weigh the logit $z_i^b$ for a backbone $b$ and example $i$, using the cross-entropy loss as expressed in the following equation: 
{\small\begin{align}
\vspace{-2mm}
    p(y \mid x) &= \text{softmax}\Big(\sum_{b \in B} t_b z^b_i\Big) \label{eq:combination_of_clip}\\
    \mathcal{L}_c &= \sum_{y \in Y} y_i \text{log} p(y|x)\nonumber
\end{align}}
Following the original proposals, which rely on data to find the optimal calibration temperature $t$, we propose utilising two existing machine learning techniques to find the set of $B$ parameters that optimize the joint confidence (the temperatures for each backbone considered in the mix). 


\begin{itemize}[leftmargin=*,topsep=1pt,itemsep=2ex,parsep=-1ex]
    \item We utilize a genetic algorithm (\GAC) to find the set of temperatures that better combine the backbones to improve the classification performance by regulating the confidence of each expert (backbone) when compared with the others. We use mutation and crossover operators to find the best temperatures that minimize the classification error.  
    \item We train a simple neural network combination (\NNC) (a single-layer MLP) to predict the set of temperatures that best calibrate our backbone mixture. As input, this model receives the concatenated representations obtained by passing the images through the encoder $\phi_v$, for each backbone $b \in \sB$. The neural net directly produces a vector of temperatures $\vt \in \sR^B$ and is trained using a cross-entropy loss. 

\end{itemize}

\begin{table*}[t]
\centering
\resizebox{0.92\textwidth}{!}{
\begin{tabular}{rccccccccc}
\toprule
\multirow{3}{*}{\bf Dataset} & \multirow{3}{*}{\Best} & \multicolumn{4}{c}{\bf Non-Parametric} & \multicolumn{4}{c}{\bf Parametric} \\
\cmidrule(lr){3-6} \cmidrule(lr){7-10} 
 & & \VoteOne & \VoteThree & \Confidence & \LogitAvg & \CalibratedConfidence & \CLogitAvg & \GAC & \NNC \\
\midrule
\Pets & 93.59 & 91.63 \downr{-1.96} & 93.00 \downr{-0.59} & 92.26 \downr{-1.33} & 92.86 \downr{-0.73} & 92.89 \downr{-0.70} & 92.94 \downr{-0.65} & 93.30 \downr{-0.29} & \textbf{94.58} \upg{0.99} \\
\Cars & 77.75 & 73.67 \downr{-4.08} & 75.70 \downr{-2.05} & 75.56 \downr{-2.19} & 75.75 \downr{-2.00} & 76.78 \downr{-0.97}& 75.96 \downr{-1.79} & 78.96 \upg{1.21} & \textbf{80.30} \upg{2.55} \\
\CUB & 62.06 & 63.32 \upg{1.26} & 64.38 \upg{2.32} & 53.71 \downr{-8.35} & 65.53 \upg{3.47} & 61.79 \downr{-0.27} & 65.01 \upg{2.95} & 66.29 \upg{4.23} & \textbf{68.40} \upg{6.34} \\
\DTD & 55.32 & 53.72 \downr{-1.60} & 55.69 \upg{0.37} & 42.55 \downr{-12.77} & 54.36 \downr{-0.96} & 55.59 \upg{0.27} & 55.21 \downr{-0.11} & 56.12 \upg{0.80} & \textbf{58.94} \upg{3.62} \\
\FGVC & 31.71 & 28.95 \downr{-2.76} & 30.96 \downr{-0.75} & 30.63 \downr{-1.08} & 31.53 \downr{-0.18} & 31.29 \downr{-0.42} & 31.89 \upg{0.18} & 33.18 \upg{1.47} & \textbf{35.88} \upg{4.17} \\
\Food & 92.32 & 90.01 \downr{-2.31} & 90.87 \downr{-1.45} & 89.62 \downr{-2.70} & 91.09 \downr{-1.23} & 91.43 \downr{-0.89} & 90.91 \downr{-1.41} & 92.91 \upg{0.59} & \textbf{93.07} \upg{0.75} \\
\Flowers & 79.05 & 75.70 \downr{-3.35} & 76.28 \downr{-2.77} & 75.85 \downr{-3.20} & 75.54 \downr{-3.51} & 77.60 \downr{-1.45} & 76.25 \downr{-2.80} & 78.16 \downr{-0.89} & \textbf{81.10} \upg{2.05} \\
\Imagenet & 75.54 & 72.67 \downr{-2.87} & 73.95 \downr{-1.59} & 67.46 \downr{-8.08} & 73.89 \downr{-1.65} & 74.86 \downr{-0.68} & 73.86 \downr{-1.68} & 76.22 \upg{0.68} & \textbf{76.59} \upg{1.05} \\ 
\midrule
Mean $\Delta$ & - & \multicolumn{1}{c}{-2.21} & \multicolumn{1}{c}{-0.81} & \multicolumn{1}{c}{-4.96}  & \multicolumn{1}{c}{-0.85} & \multicolumn{1}{c}{-0.64} & \multicolumn{1}{c}{-0.66} & \multicolumn{1}{c}{0.98}  & \multicolumn{1}{c}{2.69} \\
Max $\Delta$ & - & \multicolumn{1}{c}{1.26}   & \multicolumn{1}{c}{2.32}  & \multicolumn{1}{c}{-1.08}  & \multicolumn{1}{c}{3.47}  & \multicolumn{1}{c}{0.27}  & \multicolumn{1}{c}{2.95}  & \multicolumn{1}{c}{4.23}  & \multicolumn{1}{c}{6.34} \\ 
Min $\Delta$ & - & \multicolumn{1}{c}{-4.08}  & \multicolumn{1}{c}{-2.77} & \multicolumn{1}{c}{-12.77} & \multicolumn{1}{c}{-3.51} & \multicolumn{1}{c}{-1.45} & \multicolumn{1}{c}{-2.80} & \multicolumn{1}{c}{-0.89} & \multicolumn{1}{c}{0.75} \\ 
\bottomrule
\end{tabular}
}
\caption{Our results on combining the zero-shot predictions of CLIP backbones, which we group intro non-parametric and parametric techniques, and also compare to the best-performing single backbone (\Best). We present the improvement \upg{} and deterioration \downr{} of accuracy performance for each method when we compare it against the \Best{} backbone. Mean, Max and Min $\Delta$ summarize the difference in performance across datasets.}


\label{tab:zs-baselines}
\vspace{-5mm}
\end{table*}

\begin{table}[t]
\centering
\resizebox{0.49\textwidth}{!}{
\begin{tabular}{rllllll}
\toprule
 \textbf{Dataset }& RN50 & RN101 & ViT-B-32 & ViT-B-16 & ViT-L-14 & Oracle \\
\midrule

\Pets & 85.53 \downr{-0.27} & 89.13 \upg{2.26} & 87.71 \upg{0.25} & 91.66 \upg{2.59} & 94.63 \upg{1.04} & 98.23 \upg{0.16} \\
\Cars & 75.15 \upg{20.92} & 80.91 \upg{19.79} & 76.77 \upg{17.04} & 83.24 \upg{18.63} & 89.13 \upg{11.38} & 95.77 \upg{4.92} \\
\CUB & 65.64 \upg{19.07} & 70.69 \upg{21.06} & 70.87 \upg{17.88} & 76.49 \upg{21.21} & 83.29 \upg{21.23} & 92.56 \upg{11.36} \\
\DTD & 68.99 \upg{27.77} & 69.68 \upg{26.01} & 71.76 \upg{27.77} & 74.95 \upg{29.84} & 78.30 \upg{22.98} & 89.57 \upg{19.95} \\
\FGVC & 39.87 \upg{22.8} & 41.91 \upg{23.28} & 41.34 \upg{21.69} & 49.17 \upg{24.78} & 60.88 \upg{29.16} & 77.53 \upg{25.44} \\
\Food & 81.73 \upg{3.82} & 84.49 \upg{2.63} & 83.25 \upg{0.67} & 88.07 \upg{0.15} & 92.42 \upg{0.1} & 97.80 \upg{1.2} \\
\Flowers & 90.91 \upg{24.78} & 92.73 \upg{27.53} & 92.65 \upg{26.17} & 94.57 \upg{23.14} & 98.36 \upg{19.3} & 99.25 \upg{12.93} \\
\Imagenet & 70.25 \upg{10.4} & 72.44 \upg{10.15} & 73.01 \upg{9.66} & 77.47 \upg{9.12} & 82.15 \upg{6.63} & 90.09 \upg{4.8} \\\midrule
Mean $\Delta$ & 16.16 & 16.59 & 15.14 & 16.07 & 13.98 & 10.10 \\
Max $\Delta$ & 27.77 & 27.53 & 27.77 & 29.84 & 29.16 & 25.44 \\
Min $\Delta$ & -0.27 & 2.26 & 0.25 & 0.15 & 0.10 & 0.16 \\
\bottomrule
\end{tabular}
}
\caption{Linear probe accuracy across multiple datasets and employing different backbones, showcasing the performance improvement achieved with the linear probe in comparison with the zero-shot version with L2-Norm. The last column presents the Oracle performance, indicating accuracy benchmarks when the optimal backbone for each image sample is known in advance. The final four rows present the statistics of performance improvement.}
\label{tab:lp-table}
\vspace{-5mm}
\end{table}

\begin{table*}[t]
\centering
\resizebox{0.92\textwidth}{!}{
\begin{tabular}{rcccccccccc}
\toprule
\multirow{3}{*}{\bf Dataset} & \multirow{3}{*}{\Best} & \multicolumn{4}{c}{\bf Non-Parametric} & \multicolumn{5}{c}{\bf Parametric} \\
\cmidrule(lr){3-6} \cmidrule(lr){7-11} 
 & & \VoteOne & \VoteThree & \Confidence & \LogitAvg & \CalibratedConfidence & \CLogitAvg & MoE & \GAC & \NNC \\
\midrule
\Pets & 94.63 & 93.13 \downr{-1.5} & 93.00 \downr{-1.64} & 90.32 \downr{-4.31} & 92.78 \downr{-1.85} & 92.80 \downr{-1.83} & 93.40 \downr{-1.23} & 92.17 \downr{-2.46} & 94.46 \downr{-0.17} & \textbf{94.99} \upg{0.36} \\
\Cars & 89.13 & 87.94 \downr{-1.19} & 88.07 \downr{-1.06} & 84.39 \downr{-4.74} & 89.14 \upg{0.01} & 88.67 \downr{-0.46} & 89.08 \downr{-0.05} & 88.51 \downr{-0.62} & 90.00 \upg{0.87} & \textbf{90.19} \upg{1.06} \\
\CUB & 83.29 & 82.05 \downr{-1.24} & 81.79 \downr{-1.5} & 74.09 \downr{-9.2} & 82.72 \downr{-0.57} & 82.10 \downr{-1.19} & 83.34 \upg{0.05} & 76.80 \downr{-6.49} & 84.55 \upg{1.26} & \textbf{84.88} \upg{1.59} \\
\DTD & 78.30 & 78.35 \upg{0.05} & 78.40 \upg{0.11} & 73.03 \downr{-5.27} & 78.56 \upg{0.27} & 76.65 \downr{-1.65} & 79.10 \upg{0.8} & 72.44 \downr{-5.86} & \textbf{80.00} \upg{1.7} & 79.14 \upg{0.84} \\
\FGVC & 60.88 & 55.84 \downr{-5.04} & 56.08 \downr{-4.8} & 49.53 \downr{-11.34} & 58.36 \downr{-2.52} & 58.96 \downr{-1.92} & 59.35 \downr{-1.53} & 60.31 \downr{-0.57} & 61.51 \upg{0.63} & \textbf{62.20} \upg{1.32} \\
\Food & 92.42 & 93.03 \upg{0.61} & 92.94 \upg{0.52} & 84.21 \downr{-8.21} & 93.99 \upg{1.57} & 92.83 \upg{0.41} & 93.64 \upg{1.22} & \textbf{94.14} \upg{1.72} & 93.46 \upg{1.04} & 93.72 \upg{1.30} \\
\Flowers & 98.36 & 96.76 \downr{-1.59} & 96.81 \downr{-1.54} & 94.19 \downr{-4.16} & 96.99 \downr{-1.37} & 97.84 \downr{-0.52} & 97.17 \downr{-1.19} & 86.56 \downr{-11.8} & 98.06 \upg{-0.3} & \textbf{98.39} \upg{0.03} \\
\Imagenet & 82.15 & 77.20 \downr{-4.95} & 77.65 \downr{-4.5} & 75.36 \downr{-6.79} & 77.57 \downr{-4.59} & 81.01 \downr{-1.14} & 80.29 \downr{-1.87} & 78.81 \downr{-3.34} & 82.37 \upg{0.22} & \textbf{82.48} \upg{0.33} \\ \midrule
Mean $\Delta$ & - & -1.86 & -1.80 & -6.75  & -1.13 & -1.04 & -0.48 & -3.68  & 0.66 & 0.85 \\
Max  $\Delta$ & - & 0.61  & 0.52  & -4.16  & 1.57  &  0.41 & 1.22  & 1.72   & 1.7 & 1.59 \\
Min  $\Delta$ & - & -5.04 & -4.80 & -11.34 & -4.59 & -1.92 & -1.87 & -11.80 & -0.3 & 0.03 \\

\bottomrule
\end{tabular}
}
\vspace{-2mm}
\caption{Our results on combining the LinearProbe CLIP predictions with different backbones, which we group intro non-parametric and parametric techniques. We present the improvement \upg{} and deterioration \downr{} of accuracy performance for each method when compared with the best-performing single backbone (\Best{}). Mean, Max and Min $\Delta$ summarize the difference in performance across datasets.}
\label{tab:lp-baselines}
\vspace{-5mm}
\end{table*}
\section{Experimental Setup}
\label{sec:setup}


For our experiments, we utilize OpenCLIP \cite{ilharco_gabriel_2021_5143773}, an open source implementation of the original CLIP \cite{radford2019language}. Our research is predominantly centered on the context of zero-shot learning, where we operate with the fundamental constraint of lacking access to labeled data. As a result, our initial investigation primarily revolves around evaluating the model's performance in comparison to the baseline zero-shot CLIP paradigm. However, our study also encompasses experimental investigations into the differences in adapted CLIPs using linear probing and limited samples for combining the backbones.


To help characterize the improvements that can be achieved with our proposed approach, we consider common model ensembling techniques as baselines that are grouped into Non-Parametric and Parametric.

In the \textbf{non-parametric}, we consider four approaches commonly used in ensembles:

\paragraph{Logit averaging (\LogitAvg)}: We take the average of logits to produce a new logit that goes through a softmax to predict the class label for a certain sample. 
\paragraph{Voting (\VoteOne{} and \VoteThree)}: Voting enables us to combine conceptually different classifiers to predict the class label. It consists of using the majority of votes between the backbones to produce a final prediction. We use the top-1 (\VoteOne) prediction from each backbone, and the final prediction is determined by the label with the highest number of votes. In cases where multiple labels receive the same number of votes, we select the one with the highest probability. Additionally, we experiment with top-3 voting (\VoteThree), where we seek a consensus among backbones by considering the three most likely predictions from each of them. These predictions are weighted based on their position within the top-3 list.
\paragraph{Confidence (\Confidence)}: Using the Shannon entropy to evaluate the confidence of each backbone in each prediction, we select the backbone with the highest confidence for a prediction as the source of the final prediction.

We also consider two variations of \textbf{parametric} calibration where we perform model-wise calibration before combining models, as follows. 

\paragraph{Calibrated confidence (\CalibratedConfidence)} For this approach, we first calibrate the probabilities of each backbone independently using temperature scaling. Then follow the procedure above and utilize the Shannon entropy of each backbone to select the one with the highest confidence

\paragraph{Calibrated logit averaging (\CLogitAvg)} We first calibrate the probabilities of each backbone independently using temperature scaling. We then simply average the calibrated logits using the independently-obtained temperature parameters.

We also investigate the adaptability and \textit{orthogonality} exhibited by each backbone when subjected to adaptation for a particular dataset. This examination seeks to understand how effectively each backbone can be tailored to a target dataset's specific characteristics and determine if the orthogonality remains after adaptation.
This adaptation is facilitated through the utilization of \textbf{linear probing}. In contrast to a comprehensive fine-tuning of the pre-trained models, linear probing emerges as a more straightforward and effective methodology for tailoring foundational models to a target dataset. 
The linear probing technique employs a linear layer, wherein the weights of the pre-trained backbone remain frozen, allowing for the efficient learning of a classifier tailored to the attributes of the specific dataset.

We initialize the weights of the linear probing using language weights, a practice detailed in \cite{li2022elevater}. This initialization strategy proves notably superior to random initialization, exhibiting enhanced stability, particularly in scenarios involving few-shot learning. 
We use the output of each backbone $\phi_b$ normalized with L2 as an input to the linear probing. Each linear probe undergoes training with the target dataset's training set, allocating 90\% for the actual linear probe training and reserving the remaining 10\% for the integration of backbones through the parametric approaches.

Finally, we made use of \textbf{Mixture of Experts (MoE)}, one of the most popular techniques in machine learning to enhance model performance by combining the strengths of multiple specialized submodels \cite{rau2019moe, Zoph2022STMoEDS, lepikhin2020gshard, NEURIPS2022_91edff07}. MoEs divide the overall model into multiple experts, each adept at addressing specific regions of the input space. A gating network determines the relevance of each expert for a given input, orchestrating the collaboration. We use the implementation of Sparse MoE \cite{Zoph2022STMoEDS} by \cite{rau2019moe}, where the input to the MoE layer is the concatenation of the vision features $x_b$ from $\phi_b$, we use five experts and trained for the classification task with a cross-entropy loss. We use Adam \cite{kingma2014adam} with a learning rate of $2\times10^{-5}$ for 300 epochs. 

\subsection{Datasets and Backbones}

We evaluate our proposed approach on eight popular image classification datasets: Describable Texture Dataset (\DTD) \cite{cimpoi14describing}, FGVC-Aircraft (\FGVC) \cite{maji13fine-grained} Caltech-UCSD Birds-200-2011 (\CUB) \cite{WahCUB_200_2011}, Flowers (\Flowers) \cite{nilsback2008automated}, Pets (\Pets) \cite{parkhi2012cats}, Stanford Cars (\Cars) \cite{krause2013collecting}, Food (\Food) \cite{bossard2014food} and ImageNet-1k (\Imagenet) \cite{deng2009imagenet}. These datasets consist of 47, 100, 200, 102, 37, 196, 101 and 1,000 classes, respectively. Our objective is to evaluate the orthogonality of the predictions with respect to the different backbones used to train CLIP. Moreover, we want to evaluate probing and our method to fuse those backbones to improve the performance of the image classification task. We, therefore, utilize accuracy to evaluate the performance of each backbone of CLIP and the fusion mechanism.

We consider a broad selection of backbones. In particular, 
this work explores the ResNet \cite{He_2016_CVPR} family with its variants RN50 and RN101, and the ViT \cite{dosovitskiy2020vit} family with B-16, B-32 and L-14 variants for the image encoder $\phi_v$.

\section{Results}

\begin{table*}[t!]
\centering
\resizebox{0.67\textwidth}{!}{%
\begin{tabular}{rccccccc}
\toprule
\textbf{Dataset} & \Best & \NNC(1) & \NNC(2) & \NNC(4) & \NNC(8) & \NNC(16) & \NNC(32) \\
\midrule
\Pets & 93.59 & 94.09 \upg{0.50} & 94.03 \upg{0.44} & 93.81 \upg{0.22} & 94.18 \upg{0.59} & \textbf{94.19} \upg{0.6} & \textbf{94.19} \upg{0.60} \\
\Cars & 77.75 & 79.08 \upg{1.33} & 79.27 \upg{1.52} & 79.09 \upg{1.34} & 79.31 \upg{1.56} & 79.29 \upg{1.54} & \textbf{79.38} \upg{4.80} \\
\CUB & 62.06 & 66.76 \upg{4.70} & \textbf{66.97} \upg{4.91} & 66.81 \upg{4.75} & 66.91 \upg{4.85} & 66.95 \upg{4.89} & 66.86 \upg{2.02} \\
\DTD & 55.32 & 56.97 \upg{1.65} & 57.34 \upg{2.02} & 57.50 \upg{2.18} & 57.23 \upg{1.91} & \textbf{57.71} \upg{2.39} & 57.34 \upg{4.05} \\
\FGVC & 31.71 & 33.21 \upg{1.50} & 33.03 \upg{1.32} & 32.82 \upg{1.11} & 33.00 \upg{1.29} & 35.28 \upg{2.39} & \textbf{35.76} \upg{0.75} \\
\Food & 92.32 & 92.50 \upg{0.18} & 92.97 \upg{0.65} & 93.03 \upg{0.71} & 93.02 \upg{0.70} & \textbf{93.07} \upg{3.57} & \textbf{93.07} \upg{0.75} \\
\Flowers & 79.05 & 78.65 \downr{-0.40} & 78.95 \downr{-0.10} & 78.95 \downr{-0.10} & 79.05 \upg{0.00} & \textbf{81.80} \upg{2.75} & 81.80 \upg{2.75} \\
\Imagenet & 75.54 & 76.19 \upg{0.65} & \textbf{76.40} \upg{0.86} & 76.34 \upg{0.80} & 76.37 \upg{0.83} & 75.98 \upg{0.44} & 76.34 \upg{0.80}\\ 
\midrule
Mean $\Delta$ & - & 1.26 & 1.45 & 1.38 & 1.47 & 2.12 & 2.18 \\ 
Max $\Delta$  & - & 4.70 & 4.91 & 4.75 & 4.85 & 4.89 & 4.80 \\ 
Min $\Delta$  & - & -0.4 & -0.10 & -0.10 & 0.00 & 0.60 & 0.60 \\ 

\bottomrule
\end{tabular}%
}
\caption{Ablation of \NNC{} performance when changing the number of samples used to train. We use \NNC($n$) to denote \NNC{} with $n$ samples per class. \upg{} and \downr{} showcase the improvement in performance compared with \Best{} backbone in a zero-shot setting.}
\label{tab:NNC_limited_samples}
\vspace{-5mm}
\end{table*}
\paragraph{Combination of backbones.} 
The challenges of effectively fusing different backbones for improved predictions are evident in Table \ref{tab:zs-baselines}. In non-parametric baselines, leveraging the confidence of each backbone in its predictions consistently fails to enhance the overall performance beyond that of the best backbone. 
The lack of calibrated probabilities in the \Confidence{} contributes to overconfidence in some backbones, resulting in performance degradation when combined. Calibrating the confidence leads to improvements, although the performance still falls short of matching the best backbone, except for \DTD{}. This trend persists in the \LogitAvg{} approach, where averaging performance across backbones does not exploit their orthogonal prediction capabilities effectively, yielding an average delta accuracy of -0.85\%. Notably, the \LogitAvg{} approach shows substantial improvement for the \CUB{} dataset compared to the best backbone. When the backbones are calibrated \CLogitAvg, the \LogitAvg{} approach enhances its performances across all datasets except \CUB, \Food, and \Imagenet{} when compared to the non-calibrated version. Intriguingly, conventional ensemble techniques such as \textbf{Vote T-1} and \textbf{Vote T-3} prove ineffective in providing a significant boost to prediction accuracy beyond that of the best backbone.
In employing parametric methods, we utilize the entire training set of the target datasets. Our proposed approach \NNC{} demonstrates a noteworthy capability to enhance the performance of the best backbone, achieving a substantial improvement of up to 6.34\% in the case of the \CUB{} dataset. On average, our method exhibits a commendable improvement of 2.69\% when compared to the performance of the best backbone across the evaluated datasets.

Table \ref{tab:lp-table} illustrates the accuracy results of linear probing on each dataset and backbone. Notably, a substantial enhancement in performance is observed through this adaptation technique when compared to the zero-shot \LTwo Norm (Tab. \ref{tab:zs-all-table}). RN101 stands out as the backbone experiencing the most significant improvement, averaging a remarkable 16.59\% across all datasets, surpassing even ViT-B-32. Despite this, its performance falls short of outperforming ViT-B-16 and ViT-L-14. Even after linear probing, ViT-L-14 remains the top-performing backbone across all datasets. Noteworthy is the considerable room for improvement indicated by the \Oracle{} of linear probes, showcasing superior performance compared to any individual backbone, suggesting potential gains through a combination of backbones. Venn diagrams for the linear probe are also shown in the Appendix. 
Although the room for improvement between the Oracle and the best linear probe is consistent across datasets, we found that the space for improvement is much less than the zero-shot option. 

The outcomes presented in Table \ref{tab:lp-baselines} showcase the results of combining linear probing versions for each backbone, using both non-parametric and parametric approaches. Similar to the zero-shot setting, our proposed \NNC{} consistently demonstrates improvement across all datasets, achieving an average enhancement of 0.85\% compared to the \Best{} linear-probe backbone. Notably, this improvement is more modest compared to the zero-shot version, which records an average improvement of 2.69\%. Intriguingly, the MoE approach does not reach the performance level of the \Best{} backbone and exhibits a negative improvement of -3.68\%. This discrepancy suggests that MoE might face challenges in effectively partitioning the input space into distinct clusters specific to certain experts, potentially hindering its optimal functionality.

In our exploration of the effectiveness of the \NNC{} approach, we extend our analysis to a scenario where we limit the samples to combine the zero-shot CLIPs. This experiment allows us to assess the adaptability and performance of our proposed method under limited training data conditions. Table \ref{tab:NNC_limited_samples} presents the performance of \NNC{} by means of a limited number of samples. Although the performance of \NNC{} overall improves when it has more data available to combine the backbones, in the majority of cases just using one sample \NNC(1) per class is enough to improve its performance. Notably, there is a drop in the trend of improvement of performance when we use four samples available \NNC(4), we believe this is because of the use of the same hyperparameters in the training across a number of samples, and probably it needs a better fine-tuning.




\section{Conclusion}
\label{sec:conclusion}

In this paper, we have undertaken a comprehensive analysis of various backbones within the CLIP framework, specifically focusing on the image classification task. Unlike broader studies that span various downstream tasks, such as those by Goldblum et al. \cite{goldblum2023battle}, our emphasis lies in understanding and leveraging the unique contributions of each backbone within the CLIP context. Our research unveils a distinctive orthogonality in predictions between backbones of zero-shot CLIP and LinearProbe CLIP, presenting a valuable avenue for enhancing CLIP's performance through a synergistic combination of these backbones. The introduction of an \Oracle{} highlights this orthogonality, emphasizing the potential of mixing different backbones to optimize performance in image classification tasks. We propose an innovative approach that learns a set of temperatures, refining performance by appropriately weighting the logits from each backbone to yield an enhanced prediction. Notably, our proposed method introduces two efficient techniques, a genetic algorithm and a multi-layer perceptron (MLP), for learning these temperatures, with the MLP showcasing the ability to combine backbones effectively even with limited samples, \eg one per class. This novel methodology provides a streamlined means of capitalizing on the individual strengths of diverse backbones, ultimately boosting predictive accuracy in the realm of image classification. Our findings contribute to a fine understanding of CLIP's backbone behaviour and offer a practical strategy for optimizing its performance in real-world applications.

\newpage
{\small
\bibliographystyle{ieee_fullname}
\bibliography{egbib}
}
\clearpage
\newpage
\appendix
\section{Venn diagrams ZeroShot CLIP}
\label{sec:venn_zs_clip}
\begin{figure*}[b!]
    \centering
    \underline{\textbf{\Cars}}
    \includegraphics[width=0.99\textwidth]{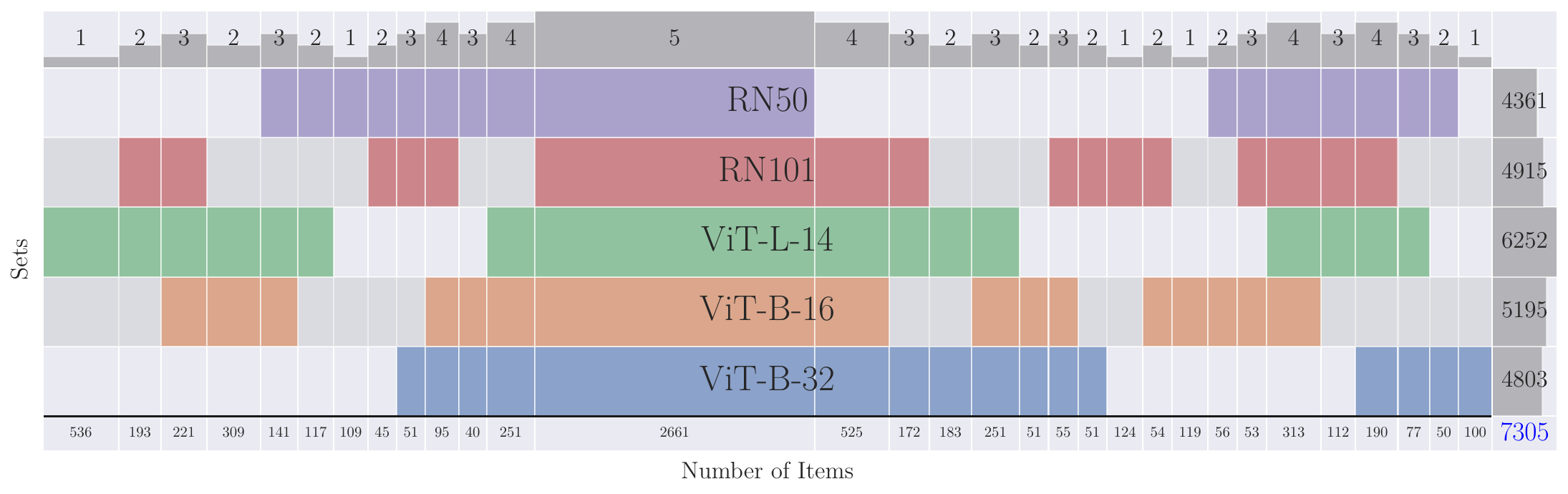}
    \underline{\textbf{\CUB}}
    \includegraphics[width=0.99\textwidth]{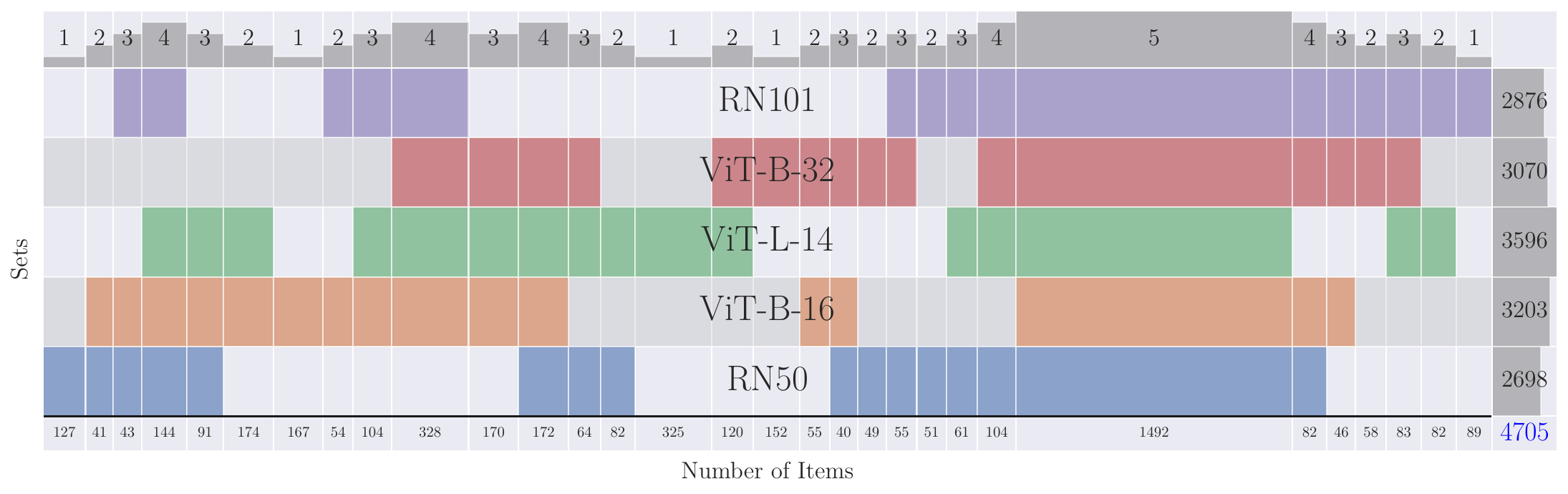}
    \underline{\textbf{\DTD}}
    \includegraphics[width=0.99\textwidth]{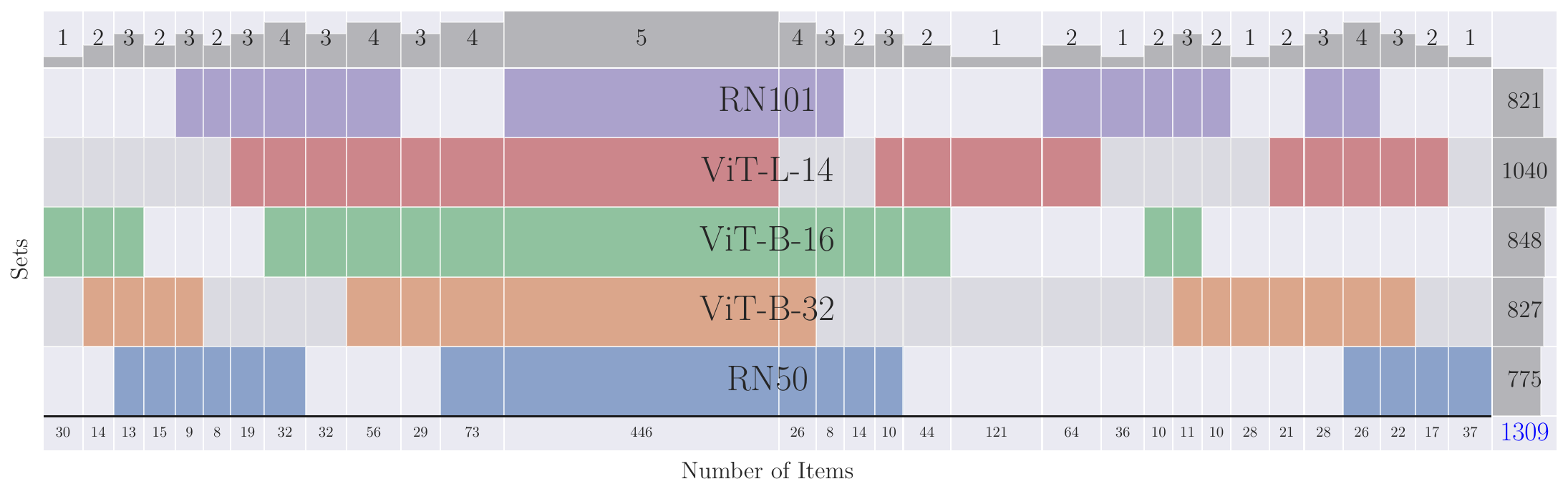}\vspace{-3mm}
    \caption{Venn diagram with the correct prediction of each backbone. The Top part of the Venn diagram shows the number of backbones that are predicting correctly a set of images. Each column represents a set of image instances that are predicted correctly by some group of backbones. Each row in the diagram shows in colour the backbone that correctly predicts a certain set of image instances, in grey when the backbone is not correctly predicting those instances. The bottom part of the Venn diagram shows the number of images in a certain set. The right part is the total amount of correctly predicted images per backbone.}
    \label{fig:venn_diagram_1_supp}
    \vspace{-4mm}
\end{figure*}
\begin{figure*}[b!]
    \centering
    \underline{\textbf{\FGVC}}
    \includegraphics[width=0.99\textwidth]{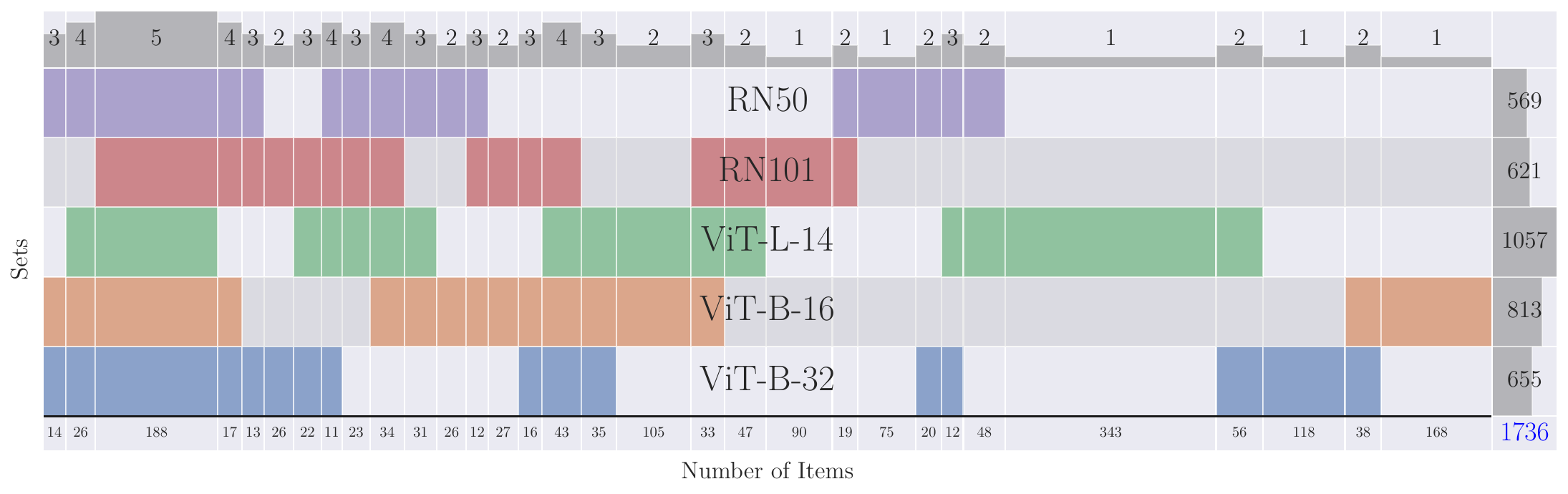}
    \underline{\textbf{\Flowers}}
    \includegraphics[width=0.99\textwidth]{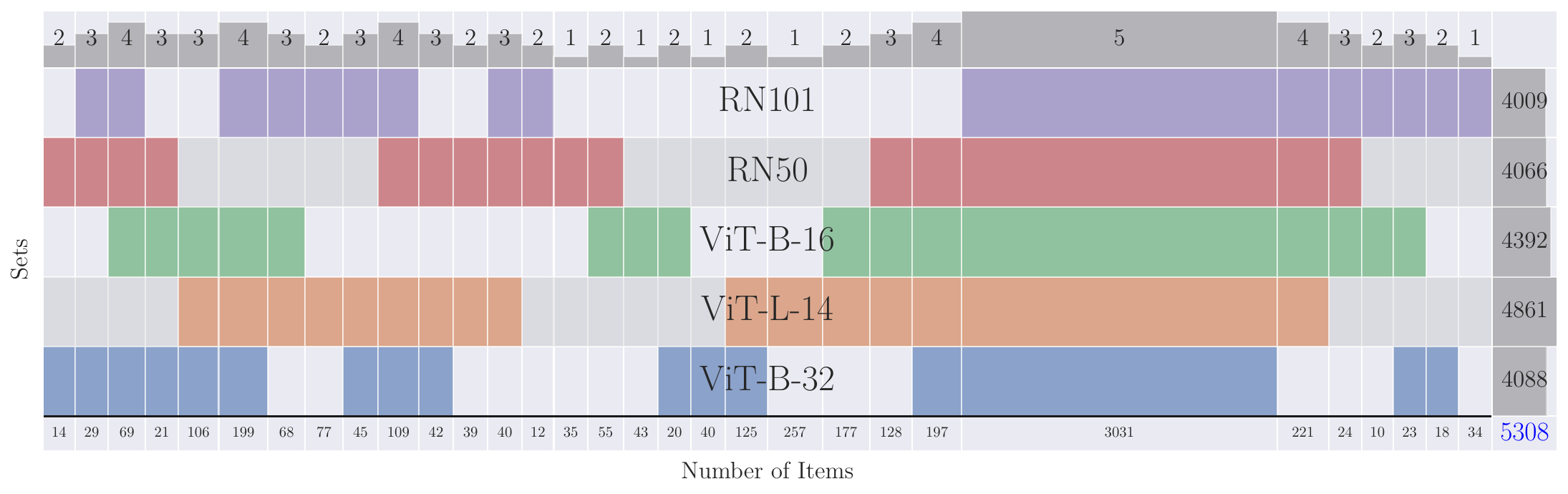}
    \underline{\textbf{\Food}}
    \includegraphics[width=0.99\textwidth]{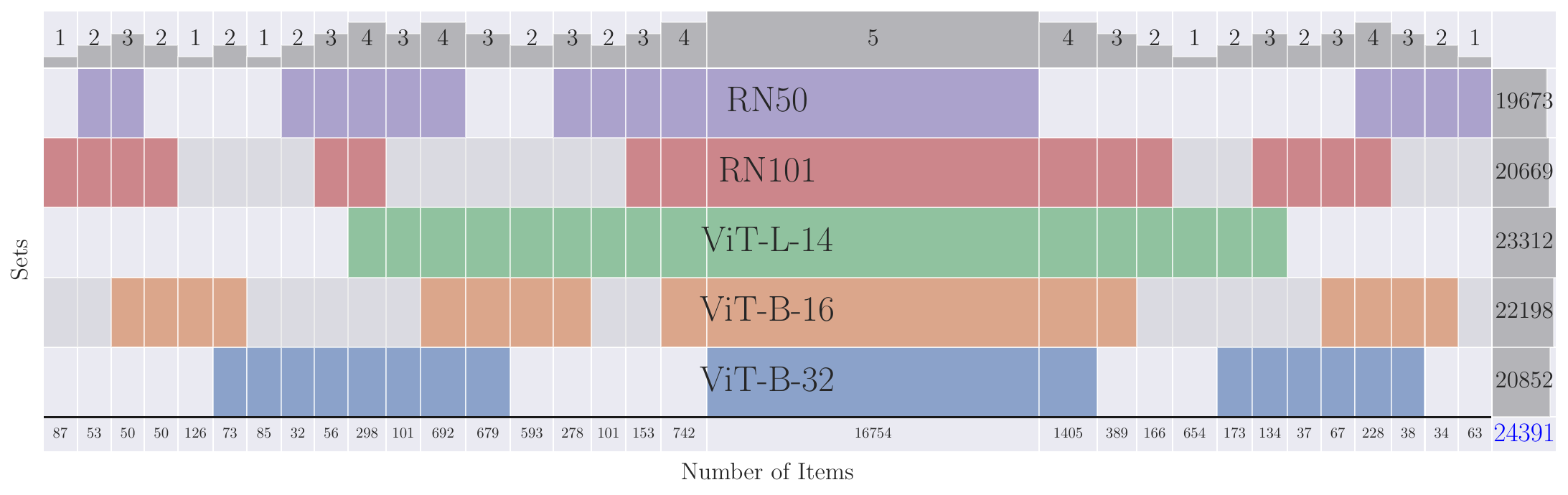}\vspace{-3mm}
    \caption{Venn diagram with the correct prediction of each backbone. The Top part of the Venn diagram shows the number of backbones that are predicting correctly a set of images. Each column represents a set of image instances that are predicted correctly by some group of backbones. Each row in the diagram shows in colour the backbone that correctly predicts a certain set of image instances, in grey when the backbone is not correctly predicting those instances. The bottom part of the Venn diagram shows the number of images in a certain set. The right part is the total amount of correctly predicted images per backbone.}
    \label{fig:venn_diagram_2_supp}
    \vspace{-4mm}
\end{figure*}

\begin{figure*}[b!]
    \centering
    \underline{\textbf{\Pets}}
    \includegraphics[width=0.99\textwidth]{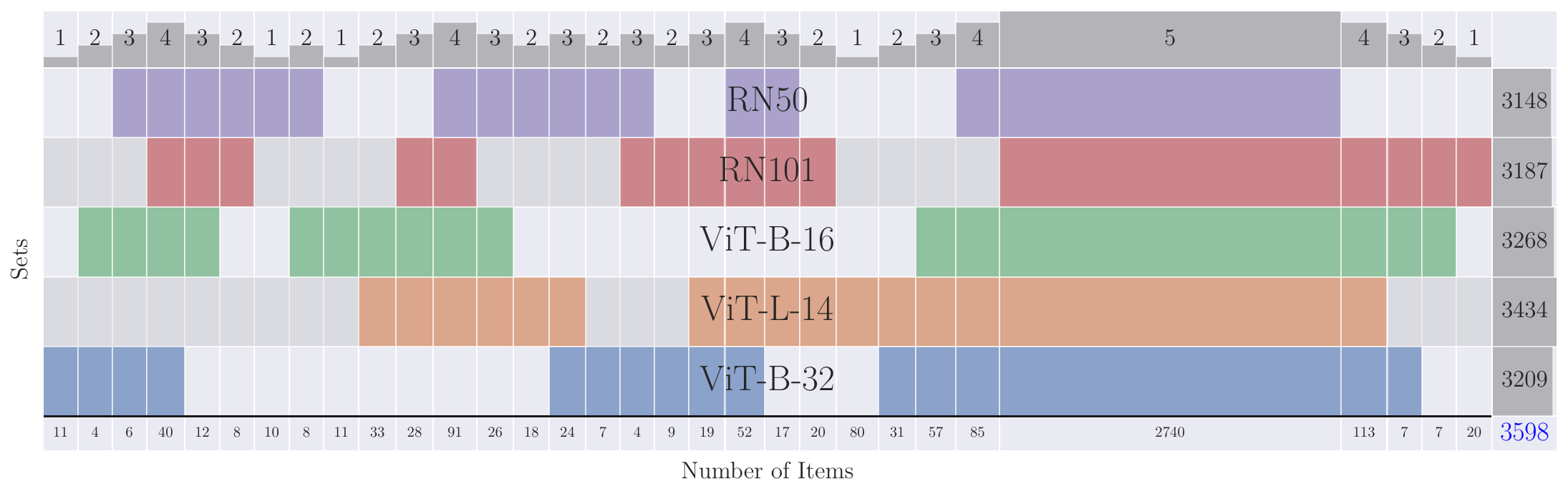}\vspace{-3mm}
    \caption{Venn diagram with the correct prediction of each backbone. The Top part of the Venn diagram shows the number of backbones that are predicting correctly a set of images. Each column represents a set of image instances that are predicted correctly by some group of backbones. Each row in the diagram shows in colour the backbone that correctly predicts a certain set of image instances, in grey when the backbone is not correctly predicting those instances. The bottom part of the Venn diagram shows the number of images in a certain set. The right part is the total amount of correctly predicted images per backbone.}
    \label{fig:venn_diagram_3_supp}
    \vspace{-4mm}
\end{figure*}
Figures \ref{fig:venn_diagram_1_supp}, \ref{fig:venn_diagram_2_supp}, and \ref{fig:venn_diagram_3_supp} visually represent the Venn diagrams for selected benchmark datasets utilizing predictions from ZeroShot CLIP. Across these instances, a consistent divergence in predictions is observed among different backbone families. Particularly noteworthy are cases such as \Cars{}, \CUB{}, \DTD{}, and \FGVC{}, where less than 50\% of correct predictions align across all four backbones. Specifically, \Cars{} exhibits a 36.43\% agreement, \CUB{} with 31.71\%, \DTD{} with 34.07\%, and \FGVC{} with 10.83\%.

Furthermore, considering the agreement within the same backbone family, \Cars{} demonstrates a 52.48\% agreement within the ViT family and a 61.15\% agreement within the ResNet family. Similarly, \CUB{} records a 48.72\% agreement within the ViT family and a 57.37\% agreement within the ResNet family. In the case of \DTD{}, there is a 49.19\% agreement within the ViT family and a 56.16\% agreement within the ResNet family. \FGVC{} shows an agreement of 17.14\% within the ViT family and 36.31\% within the ResNet family. Importantly, the overall agreement within the ResNet family appears to be larger than that within the ViT family, suggesting a potentially lesser scope for improvement given the diversity of predictions. These insights highlight the distinct predictive characteristics within and between backbone families, showcasing the potential for leveraging this diversity to enhance overall model performance.
\section{All possible combinations per dataset}
Tables \ref{tab:all_possible_combinations_pets}, \ref{tab:all_possible_combinations_cars}, \ref{tab:all_possible_combinations_cub}, \ref{tab:all_possible_combinations_dtd}, \ref{tab:all_possible_combinations_fgvc}, \ref{tab:all_possible_combinations_food}, \ref{tab:all_possible_combinations_flowers}, and \ref{tab:all_possible_combinations_inet} present the results of all possible combinations of backbones using the non-parametric and parametric approaches proposed in the paper. Notably, the performance of \NNC{} consistently emerges as the best across various backbone combinations and datasets when compared to other methods.

It's noteworthy that instances exist where the combination of specific backbones yields a more substantial performance boost than utilizing all backbones together. For instance, in the \Pets{} dataset, combining ResNet 50, 101, and ViT-B-32 results in a delta improvement of 2.37\%, surpassing the 0.99\% improvement achieved by using all backbones. This phenomenon is consistent across datasets with different backbone combinations. In \Cars{}, there is a boost of 5.71\% when combining ResNet-101 and ViT-B-32, compared to the 2.55\% boost when using all five different backbones. Similarly, in \DTD{}, there is a 3.62\% improvement when all backbones are combined, whereas combining ResNet 50, 101, and ViT-B-16 yields a higher improvement of 5.11\%. While the best delta improvement among backbones may not necessarily come from combining all backbones, the best overall accuracy is consistently obtained when using the combination of all backbones.
\begin{table*}[b]
\centering
\Pets \\
\qquad
\resizebox{0.9\textwidth}{!}{
\begin{tabular}{ccccccccccccccc}
\toprule
\multicolumn{2}{c}{ResNet} & \multicolumn{3}{c}{ViT} & \multirow{2}{*}{\Best}  & \multicolumn{4}{c}{Non-Parametric} & \multicolumn{4}{c}{Parametric} & \multirow{2}{*}{\Oracle}\\
\cmidrule(lr){1-2} \cmidrule(lr){3-5} \cmidrule(lr){7-10} \cmidrule(lr){11-14}
50 & 101 & B-32 & B-16 & L-14 & & \VoteOne & \VoteThree & \Confidence & \LogitAvg & \CalibratedConfidence & \CLogitAvg & \GAC & \NNC &  \\
\midrule
\Checkmark &   &   &   &   &  \multicolumn{10}{c}{\xfill{.1em}  85.80 \samey{0.00} \xfill{.1em}} \\ 
  & \Checkmark &   &   &   & \multicolumn{10}{c}{\xfill{.1em}  86.86 \samey{0.00} \xfill{.1em}} \\ 
  &   & \Checkmark &   &   & \multicolumn{10}{c}{\xfill{.1em}  87.46 \samey{0.00} \xfill{.1em}} \\ 
  &   &   & \Checkmark &   & \multicolumn{10}{c}{\xfill{.1em}  89.07 \samey{0.00} \xfill{.1em}} \\ 
  &   &   &   & \Checkmark & \multicolumn{10}{c}{\xfill{.1em}  93.59 \samey{0.00} \xfill{.1em}} \\ \midrule
\Checkmark & \Checkmark &   &   &   & 86.86 & 87.84 \upg{0.98} & 87.90 \upg{1.04} & 88.20 \upg{1.34} & 88.03 \upg{1.17} & 87.93 \upg{1.06} & 87.84 \upg{0.98} & 87.14 \upg{0.27} & 88.09 \upg{1.23} & 91.88 \upg{5.01} \\ \midrule
\Checkmark &   & \Checkmark &   &   & 87.46 & 88.50 \upg{1.04} & 89.02 \upg{1.55} & 85.25 \downr{-2.21} & 89.40 \upg{1.94} & 87.98 \upg{0.52} & 89.21 \upg{1.74} & 89.48 \upg{2.02} & 89.45 \upg{1.99} & 92.64 \upg{5.18} \\ \midrule
\Checkmark &   &   & \Checkmark &   & 89.07 & 89.83 \upg{0.76} & 89.92 \upg{0.84} & 89.67 \upg{0.6} & 89.83 \upg{0.76} & 89.48 \upg{0.41} & 89.72 \upg{0.65} & 89.86 \upg{0.79} & 90.27 \upg{1.2} & 92.89 \upg{3.82} \\ \midrule
\Checkmark &   &   &   & \Checkmark & 93.59 & 94.03 \upg{0.44} & 94.00 \upg{0.41} & 94.00 \upg{0.41} & 93.81 \upg{0.22} & 93.73 \upg{0.14} & 93.98 \upg{0.38} & 93.73 \upg{0.14} & 94.19 \upg{0.6} & 96.18 \upg{2.59} \\ \midrule
  & \Checkmark & \Checkmark &   &   & 87.46 & 88.72 \upg{1.25} & 88.99 \upg{1.53} & 85.91 \downr{-1.55} & 89.07 \upg{1.61} & 88.55 \upg{1.09} & 89.07 \upg{1.61} & 89.10 \upg{1.64} & 89.04 \upg{1.58} & 93.00 \upg{5.53} \\ \midrule
  & \Checkmark &   & \Checkmark &   & 89.07 & 90.00 \upg{0.93} & 90.11 \upg{1.04} & 89.75 \upg{0.68} & 89.89 \upg{0.82} & 89.81 \upg{0.74} & 90.11 \upg{1.04} & 89.45 \upg{0.38} & 90.30 \upg{1.23} & 93.13 \upg{4.06} \\ \midrule
  & \Checkmark &   &   & \Checkmark & 93.59 & 93.32 \downr{-0.27} & 93.32 \downr{-0.27} & 93.13 \downr{-0.46} & 93.24 \downr{-0.35} & 93.27 \downr{-0.33} & 93.43 \downr{-0.16} & 93.70 \upg{0.11} & 93.70 \upg{0.11} & 96.51 \upg{2.92} \\ \midrule
  &   & \Checkmark & \Checkmark &   & 89.07 & 89.86 \upg{0.79} & 90.16 \upg{1.09} & 87.35 \downr{-1.72} & 90.41 \upg{1.34} & 89.45 \upg{0.38} & 90.38 \upg{1.31} & 90.27 \upg{1.2} & 90.30 \upg{1.23} & 93.35 \upg{4.28} \\ \midrule
  &   & \Checkmark &   & \Checkmark & 93.59 & 93.79 \upg{0.19} & 93.76 \upg{0.16} & 93.00 \downr{-0.6} & 93.38 \downr{-0.22} & 93.00 \downr{-0.6} & 93.73 \upg{0.14} & 93.89 \upg{0.3} & 93.87 \upg{0.27} & 95.99 \upg{2.4} \\ \midrule
  &   &   & \Checkmark & \Checkmark & 93.59 & 93.98 \upg{0.38} & 93.95 \upg{0.35} & 93.92 \upg{0.33} & 94.06 \upg{0.46} & 93.84 \upg{0.25} & 93.98 \upg{0.38} & 93.73 \upg{0.14} & 93.92 \upg{0.33} & 96.18 \upg{2.59} \\ \midrule
\Checkmark & \Checkmark & \Checkmark &   &   & 87.46 & 88.85 \upg{1.39} & 89.29 \upg{1.83} & 84.93 \downr{-2.53} & 89.72 \upg{2.26} & 88.42 \upg{0.95} & 89.45 \upg{1.99} & 89.48 \upg{2.02} & 89.83 \upg{2.37} & 94.69 \upg{7.22} \\ \midrule
\Checkmark & \Checkmark &   & \Checkmark &   & 89.07 & 89.29 \upg{0.22} & 89.81 \upg{0.74} & 90.02 \upg{0.95} & 90.11 \upg{1.04} & 89.86 \upg{0.79} & 90.08 \upg{1.01} & 90.11 \upg{1.04} & 90.38 \upg{1.31} & 94.74 \upg{5.67} \\ \midrule
\Checkmark & \Checkmark &   &   & \Checkmark & 93.59 & 91.58 \downr{-2.02} & 92.97 \downr{-0.63} & 93.43 \downr{-0.16} & 93.19 \downr{-0.41} & 93.30 \downr{-0.3} & 92.86 \downr{-0.74} & 93.70 \upg{0.11} & 94.06 \upg{0.46} & 97.36 \upg{3.76} \\ \midrule
\Checkmark &   & \Checkmark & \Checkmark &   & 89.07 & 90.16 \upg{1.09} & 90.46 \upg{1.39} & 86.73 \downr{-2.34} & 90.62 \upg{1.55} & 89.15 \upg{0.08} & 90.46 \upg{1.39} & 90.35 \upg{1.28} & 90.79 \upg{1.72} & 94.79 \upg{5.72} \\ \midrule
\Checkmark &   & \Checkmark &   & \Checkmark & 93.59 & 92.50 \downr{-1.09} & 93.08 \downr{-0.52} & 92.86 \downr{-0.74} & 93.02 \downr{-0.57} & 92.86 \downr{-0.74} & 93.19 \downr{-0.41} & 93.32 \downr{-0.27} & 94.25 \upg{0.65} & 97.03 \upg{3.43} \\ \midrule
\Checkmark &   &   & \Checkmark & \Checkmark & 93.59 & 92.97 \downr{-0.63} & 93.70 \upg{0.11} & 93.98 \upg{0.38} & 93.87 \upg{0.27} & 93.84 \upg{0.25} & 93.84 \upg{0.25} & 94.19 \upg{0.6} & 94.47 \upg{0.87} & 96.97 \upg{3.38} \\ \midrule
  & \Checkmark & \Checkmark & \Checkmark &   & 89.07 & 89.83 \upg{0.76} & 90.16 \upg{1.09} & 86.54 \downr{-2.53} & 90.65 \upg{1.58} & 89.83 \upg{0.76} & 90.41 \upg{1.34} & 90.35 \upg{1.28} & 90.71 \upg{1.64} & 95.12 \upg{6.05} \\ \midrule
  & \Checkmark & \Checkmark &   & \Checkmark & 93.59 & 92.20 \downr{-1.39} & 93.02 \downr{-0.57} & 92.56 \downr{-1.04} & 92.75 \downr{-0.84} & 92.70 \downr{-0.9} & 92.94 \downr{-0.65} & 94.00 \upg{0.41} & 93.98 \upg{0.38} & 97.27 \upg{3.68} \\ \midrule
  & \Checkmark &   & \Checkmark & \Checkmark & 93.59 & 92.91 \downr{-0.68} & 93.46 \downr{-0.14} & 93.46 \downr{-0.14} & 93.43 \downr{-0.16} & 93.57 \downr{-0.03} & 93.49 \downr{-0.11} & 93.84 \upg{0.25} & 93.92 \upg{0.33} & 97.30 \upg{3.71} \\ \midrule
  &   & \Checkmark & \Checkmark & \Checkmark & 93.59 & 92.97 \downr{-0.63} & 93.27 \downr{-0.33} & 92.89 \downr{-0.71} & 93.51 \downr{-0.08} & 93.13 \downr{-0.46} & 93.40 \downr{-0.19} & 94.03 \upg{0.44} & 94.17 \upg{0.57} & 97.03 \upg{3.43} \\ \midrule
\Checkmark & \Checkmark & \Checkmark & \Checkmark &   & 89.07 & 90.27 \upg{1.2} & 90.41 \upg{1.34} & 86.21 \downr{-2.86} & 90.57 \upg{1.5} & 89.48 \upg{0.41} & 90.54 \upg{1.47} & 90.35 \upg{1.28} & 90.90 \upg{1.83} & 95.88 \upg{6.81} \\ \midrule
\Checkmark & \Checkmark & \Checkmark &   & \Checkmark & 93.59 & 92.20 \downr{-1.39} & 92.80 \downr{-0.79} & 92.10 \downr{-1.5} & 92.91 \downr{-0.68} & 92.64 \downr{-0.95} & 92.80 \downr{-0.79} & 93.16 \downr{-0.44} & 94.17 \upg{0.57} & 97.77 \upg{4.17} \\ \midrule
\Checkmark & \Checkmark &   & \Checkmark & \Checkmark & 93.59 & 92.56 \downr{-1.04} & 92.94 \downr{-0.65} & 93.59 \samey{0.0} & 93.19 \downr{-0.41} & 93.54 \downr{-0.05} & 92.94 \downr{-0.65} & 93.81 \upg{0.22} & 94.52 \upg{0.93} & 97.77 \upg{4.17} \\ \midrule
\Checkmark &   & \Checkmark & \Checkmark & \Checkmark & 93.59 & 92.91 \downr{-0.68} & 93.27 \downr{-0.33} & 92.78 \downr{-0.82} & 93.19 \downr{-0.41} & 93.05 \downr{-0.55} & 93.27 \downr{-0.33} & 93.79 \upg{0.19} & 94.33 \upg{0.74} & 97.52 \upg{3.92} \\ \midrule
  & \Checkmark & \Checkmark & \Checkmark & \Checkmark & 93.59 & 92.61 \downr{-0.98} & 93.00 \downr{-0.6} & 92.12 \downr{-1.47} & 92.94 \downr{-0.65} & 92.91 \downr{-0.68} & 93.05 \downr{-0.55} & 93.68 \upg{0.08} & 94.03 \upg{0.44} & 97.79 \upg{4.2} \\ \midrule
\Checkmark & \Checkmark & \Checkmark & \Checkmark & \Checkmark & 93.59 & 91.63 \downr{-1.96} & 93.00 \downr{-0.59} & 92.26 \downr{-1.34} & 92.86 \downr{-0.74} & 92.89 \downr{-0.71} & 92.94 \downr{-0.65} & 93.30 \downr{-0.29} & 94.58 \upg{0.99} & 98.06 \upg{4.47} \\ \midrule

\multicolumn{5}{c}{Mean $\Delta$} & & -0.05 & 0.35 & -0.77 & 0.42 & 0.06 & 0.40 & 0.58 & 0.98 & 4.31 \\ 
\multicolumn{5}{c}{Max $\Delta$} & & 1.39 & 1.83 & 1.34 & 2.26 & 1.09 & 1.99 & 2.02 & 2.37 & 7.22 \\ 
\multicolumn{5}{c}{Min $\Delta$} & & -2.02 & -0.79 & -2.86 & -0.84 & -0.95 & -0.79 & -0.44 & 0.11 & 2.40 \\ 
\bottomrule
\end{tabular}
}
\caption{Our results on \Pets{} dataset for all the possible combinations of combining the zero-shot predictions of CLIP backbones, which we group intro non-parametric and parametric techniques. Also, the best-performing single backbone (\Best) and the \Oracle performance. We present, for each combination of backbones, the improvement \upg{}, constancy \samey{} and deterioration \downr{} of accuracy performance for each method when we compare it against the \Best{} backbone. Mean, Max, and Min $\Delta$ summarize the difference in performance across methods and backbone combinations.}
\label{tab:all_possible_combinations_pets}
\end{table*}

\begin{table*}[]
\centering
\Cars \\
\qquad
\resizebox{0.9\textwidth}{!}{

\begin{tabular}{ccccccccccccccc}
\toprule
\multicolumn{2}{c}{ResNet} & \multicolumn{3}{c}{ViT} & \multirow{2}{*}{\Best}  & \multicolumn{4}{c}{Non-Parametric} & \multicolumn{4}{c}{Parametric} & \multirow{2}{*}{\Oracle}\\
\cmidrule(lr){1-2} \cmidrule(lr){3-5} \cmidrule(lr){7-10} \cmidrule(lr){11-14}
50 & 101 & B-32 & B-16 & L-14 & & \VoteOne & \VoteThree & \Confidence & \LogitAvg & \CalibratedConfidence & \CLogitAvg & \GAC & \NNC &  \\
\midrule

\Checkmark &   &   &   &   &  \multicolumn{10}{c}{\xfill{.1em}  54.23 \samey{0.00} \xfill{.1em}} \\ 
  & \Checkmark &   &   &   & \multicolumn{10}{c}{\xfill{.1em}  61.12 \samey{0.00} \xfill{.1em}} \\ 
  &   & \Checkmark &   &   & \multicolumn{10}{c}{\xfill{.1em}  59.73 \samey{0.00} \xfill{.1em}} \\ 
  &   &   & \Checkmark &   & \multicolumn{10}{c}{\xfill{.1em}  64.61 \samey{0.00} \xfill{.1em}} \\ 
  &   &   &   & \Checkmark & \multicolumn{10}{c}{\xfill{.1em}  77.75 \samey{0.00} \xfill{.1em}} \\ \midrule
\Checkmark & \Checkmark &   &   &   & 61.12 & 62.41 \upg{1.28} & 62.67 \upg{1.54} & 53.41 \downr{-7.71} & 63.03 \upg{1.9} & 61.81 \upg{0.68} & 62.83 \upg{1.7} & 63.52 \upg{2.4} & 63.59 \upg{2.46} & 71.58 \upg{10.46} \\ \midrule
\Checkmark &   & \Checkmark &   &   & 59.73 & 61.30 \upg{1.57} & 62.26 \upg{2.52} & 60.84 \upg{1.11} & 63.39 \upg{3.66} & 61.14 \upg{1.41} & 62.52 \upg{2.79} & 63.41 \upg{3.68} & 63.40 \upg{3.67} & 71.50 \upg{11.76} \\ \midrule
\Checkmark &   &   & \Checkmark &   & 64.61 & 65.27 \upg{0.66} & 65.55 \upg{0.95} & 53.94 \downr{-10.67} & 66.01 \upg{1.41} & 64.97 \upg{0.36} & 65.89 \upg{1.28} & 66.55 \upg{1.94} & 66.80 \upg{2.19} & 73.95 \upg{9.34} \\ \midrule
\Checkmark &   &   &   & \Checkmark & 77.75 & 77.58 \downr{-0.17} & 77.76 \upg{0.01} & 77.69 \downr{-0.06} & 77.24 \downr{-0.51} & 77.47 \downr{-0.29} & 77.64 \downr{-0.11} & 78.51 \upg{0.76} & 78.57 \upg{0.82} & 83.96 \upg{6.21} \\ \midrule
  & \Checkmark & \Checkmark &   &   & 61.12 & 64.42 \upg{3.3} & 64.99 \upg{3.87} & 63.41 \upg{2.29} & 66.89 \upg{5.77} & 63.71 \upg{2.59} & 65.66 \upg{4.54} & 66.86 \upg{5.73} & 66.83 \upg{5.71} & 73.60 \upg{12.47} \\ \midrule
  & \Checkmark &   & \Checkmark &   & 64.61 & 67.16 \upg{2.55} & 67.74 \upg{3.13} & 59.28 \downr{-5.32} & 67.99 \upg{3.38} & 66.66 \upg{2.05} & 68.11 \upg{3.51} & 68.03 \upg{3.42} & 68.20 \upg{3.59} & 76.27 \upg{11.67} \\ \midrule
  & \Checkmark &   &   & \Checkmark & 77.75 & 77.47 \downr{-0.29} & 77.78 \upg{0.02} & 77.42 \downr{-0.34} & 77.73 \downr{-0.02} & 77.44 \downr{-0.31} & 78.10 \upg{0.35} & 78.92 \upg{1.17} & 79.11 \upg{1.36} & 84.32 \upg{6.57} \\ \midrule
  &   & \Checkmark & \Checkmark &   & 64.61 & 66.34 \upg{1.73} & 66.84 \upg{2.24} & 65.41 \upg{0.81} & 67.95 \upg{3.35} & 65.87 \upg{1.27} & 67.26 \upg{2.65} & 68.00 \upg{3.4} & 67.77 \upg{3.16} & 75.48 \upg{10.87} \\ \midrule
  &   & \Checkmark &   & \Checkmark & 77.75 & 77.55 \downr{-0.2} & 77.86 \upg{0.11} & 77.55 \downr{-0.2} & 77.42 \downr{-0.34} & 77.52 \downr{-0.24} & 77.80 \upg{0.05} & 78.68 \upg{0.93} & 78.86 \upg{1.11} & 83.88 \upg{6.13} \\ \midrule
  &   &   & \Checkmark & \Checkmark & 77.75 & 77.54 \downr{-0.21} & 77.45 \downr{-0.3} & 77.32 \downr{-0.44} & 77.79 \upg{0.04} & 77.18 \downr{-0.57} & 77.63 \downr{-0.12} & 78.88 \upg{1.13} & 78.91 \upg{1.16} & 84.26 \upg{6.5} \\ \midrule
\Checkmark & \Checkmark & \Checkmark &   &   & 61.12 & 64.63 \upg{3.51} & 65.23 \upg{4.1} & 57.36 \downr{-3.77} & 66.56 \upg{5.43} & 63.69 \upg{2.56} & 66.01 \upg{4.89} & 66.76 \upg{5.63} & 66.76 \upg{5.63} & 78.86 \upg{17.73} \\ \midrule
\Checkmark & \Checkmark &   & \Checkmark &   & 64.61 & 66.34 \upg{1.73} & 67.35 \upg{2.75} & 59.66 \downr{-4.95} & 67.60 \upg{3.0} & 66.65 \upg{2.04} & 67.77 \upg{3.16} & 67.95 \upg{3.35} & 68.20 \upg{3.59} & 80.66 \upg{16.06} \\ \midrule
\Checkmark & \Checkmark &   &   & \Checkmark & 77.75 & 73.80 \downr{-3.95} & 76.23 \downr{-1.52} & 76.10 \downr{-1.65} & 76.59 \downr{-1.16} & 77.25 \downr{-0.5} & 76.86 \downr{-0.9} & 79.06 \upg{1.31} & 78.88 \upg{1.13} & 87.49 \upg{9.74} \\ \midrule
\Checkmark &   & \Checkmark & \Checkmark &   & 64.61 & 66.24 \upg{1.63} & 67.06 \upg{2.45} & 56.24 \downr{-8.37} & 67.94 \upg{3.33} & 66.07 \upg{1.47} & 67.62 \upg{3.01} & 68.46 \upg{3.86} & 68.29 \upg{3.68} & 80.24 \upg{15.63} \\ \midrule
\Checkmark &   & \Checkmark &   & \Checkmark & 77.75 & 75.04 \downr{-2.71} & 77.17 \downr{-0.58} & 77.50 \downr{-0.25} & 76.71 \downr{-1.04} & 77.34 \downr{-0.41} & 77.24 \downr{-0.51} & 78.73 \upg{0.98} & 78.71 \upg{0.96} & 87.15 \upg{9.4} \\ \midrule
\Checkmark &   &   & \Checkmark & \Checkmark & 77.75 & 74.53 \downr{-3.22} & 76.28 \downr{-1.47} & 75.90 \downr{-1.85} & 76.76 \downr{-0.99} & 76.97 \downr{-0.78} & 76.94 \downr{-0.81} & 78.40 \upg{0.65} & 79.01 \upg{1.26} & 87.43 \upg{9.68} \\ \midrule
  & \Checkmark & \Checkmark & \Checkmark &   & 64.61 & 67.86 \upg{3.26} & 68.95 \upg{4.34} & 60.40 \downr{-4.2} & 69.36 \upg{4.75} & 67.04 \upg{2.44} & 69.12 \upg{4.51} & 69.16 \upg{4.55} & 69.34 \upg{4.74} & 81.37 \upg{16.76} \\ \midrule
  & \Checkmark & \Checkmark &   & \Checkmark & 77.75 & 75.04 \downr{-2.71} & 76.96 \downr{-0.8} & 77.17 \downr{-0.58} & 76.88 \downr{-0.87} & 77.18 \downr{-0.57} & 77.14 \downr{-0.61} & 79.24 \upg{1.49} & 79.26 \upg{1.5} & 87.32 \upg{9.56} \\ \midrule
  & \Checkmark &   & \Checkmark & \Checkmark & 77.75 & 75.60 \downr{-2.15} & 76.55 \downr{-1.21} & 75.70 \downr{-2.05} & 77.25 \downr{-0.5} & 77.15 \downr{-0.6} & 77.37 \downr{-0.39} & 79.02 \upg{1.27} & 79.06 \upg{1.31} & 87.63 \upg{9.87} \\ \midrule
  &   & \Checkmark & \Checkmark & \Checkmark & 77.75 & 75.23 \downr{-2.52} & 76.71 \downr{-1.04} & 77.28 \downr{-0.47} & 77.35 \downr{-0.4} & 77.19 \downr{-0.56} & 77.15 \downr{-0.6} & 78.42 \upg{0.67} & 79.04 \upg{1.29} & 87.39 \upg{9.64} \\ \midrule
\Checkmark & \Checkmark & \Checkmark & \Checkmark &   & 64.61 & 67.72 \upg{3.11} & 68.54 \upg{3.93} & 60.59 \downr{-4.02} & 68.77 \upg{4.17} & 66.98 \upg{2.38} & 68.56 \upg{3.95} & 69.06 \upg{4.45} & 69.16 \upg{4.55} & 84.18 \upg{19.57} \\ \midrule
\Checkmark & \Checkmark & \Checkmark &   & \Checkmark & 77.75 & 73.81 \downr{-3.94} & 75.87 \downr{-1.88} & 75.97 \downr{-1.78} & 76.04 \downr{-1.72} & 77.08 \downr{-0.67} & 76.43 \downr{-1.32} & 78.56 \upg{0.81} & 79.12 \upg{1.37} & 89.37 \upg{11.62} \\ \midrule
\Checkmark & \Checkmark &   & \Checkmark & \Checkmark & 77.75 & 74.69 \downr{-3.06} & 76.06 \downr{-1.69} & 75.66 \downr{-2.09} & 76.41 \downr{-1.34} & 76.93 \downr{-0.82} & 76.47 \downr{-1.28} & 79.07 \upg{1.32} & 79.22 \upg{1.47} & 89.60 \upg{11.85} \\ \midrule
\Checkmark &   & \Checkmark & \Checkmark & \Checkmark & 77.75 & 74.51 \downr{-3.25} & 75.95 \downr{-1.8} & 75.90 \downr{-1.85} & 76.57 \downr{-1.18} & 77.01 \downr{-0.75} & 76.05 \downr{-1.7} & 78.83 \upg{1.08} & 79.19 \upg{1.44} & 89.30 \upg{11.55} \\ \midrule
  & \Checkmark & \Checkmark & \Checkmark & \Checkmark & 77.75 & 75.28 \downr{-2.47} & 76.48 \downr{-1.27} & 75.54 \downr{-2.21} & 76.61 \downr{-1.14} & 76.99 \downr{-0.76} & 76.91 \downr{-0.85} & 79.07 \upg{1.32} & 79.33 \upg{1.58} & 89.49 \upg{11.74} \\ \midrule
\Checkmark & \Checkmark & \Checkmark & \Checkmark & \Checkmark & 77.75 & 73.67 \downr{-4.08} & 75.70 \downr{-2.05} & 75.56 \downr{-2.19} & 75.75 \downr{-2.0} & 76.78 \downr{-0.97} & 75.96 \downr{-1.79} & 78.96 \upg{1.21} & 80.30 \upg{2.55} & 90.85 \upg{13.1} \\ \midrule

\multicolumn{5}{c}{Mean $\Delta$} & & -0.41 & 0.63 & -2.42 & 1.04 & 0.40 & 0.98 & 2.25 & 2.43 & 11.36 \\ 
\multicolumn{5}{c}{Max $\Delta$} & & 3.51 & 4.34 & 2.29 & 5.77 & 2.59 & 4.89 & 5.73 & 5.71 & 19.57 \\ 
\multicolumn{5}{c}{Min $\Delta$} & & -4.08 & -2.05 & -10.67 & -2.00 & -0.97 & -1.79 & 0.65 & 0.82 & 6.13 \\ 
\bottomrule
\end{tabular}
}
\caption{Our results on \Cars{} dataset for all the possible combinations of combining the zero-shot predictions of CLIP backbones, which we group intro non-parametric and parametric techniques. Also, the best-performing single backbone (\Best) and the \Oracle performance. We present, for each combination of backbones, the improvement \upg{}, constancy \samey{} and deterioration \downr{} of accuracy performance for each method when we compare it against the \Best{} backbone. Mean, Max, and Min $\Delta$ summarize the difference in performance across methods and backbone combinations.}
\label{tab:all_possible_combinations_cars}
\end{table*}

\begin{table*}[]
\centering
\CUB \\
\qquad
\resizebox{0.9\textwidth}{!}{

\begin{tabular}{ccccccccccccccc}
\toprule
\multicolumn{2}{c}{ResNet} & \multicolumn{3}{c}{ViT} & \multirow{2}{*}{\Best}  & \multicolumn{4}{c}{Non-Parametric} & \multicolumn{4}{c}{Parametric} & \multirow{2}{*}{\Oracle}\\
\cmidrule(lr){1-2} \cmidrule(lr){3-5} \cmidrule(lr){7-10} \cmidrule(lr){11-14}
50 & 101 & B-32 & B-16 & L-14 & & \VoteOne & \VoteThree & \Confidence & \LogitAvg & \CalibratedConfidence & \CLogitAvg & \GAC & \NNC &  \\
\midrule

\Checkmark &   &   &   &   &  \multicolumn{10}{c}{\xfill{.1em} 46.57 \samey{0.00} \xfill{.1em}} \\ 
  & \Checkmark &   &   &   & \multicolumn{10}{c}{\xfill{.1em}  49.64 \samey{0.00} \xfill{.1em}} \\ 
  &   & \Checkmark &   &   & \multicolumn{10}{c}{\xfill{.1em}  52.99 \samey{0.00} \xfill{.1em}} \\ 
  &   &   & \Checkmark &   & \multicolumn{10}{c}{\xfill{.1em}  55.28 \samey{0.00} \xfill{.1em}} \\ 
  &   &   &   & \Checkmark & \multicolumn{10}{c}{\xfill{.1em}  62.06 \samey{0.00} \xfill{.1em}} \\ \midrule
\Checkmark & \Checkmark &   &   &   & 49.64 & 51.28 \upg{1.64} & 52.33 \upg{2.69} & 45.41 \downr{-4.23} & 55.06 \upg{5.42} & 50.72 \upg{1.09} & 53.56 \upg{3.92} & 55.06 \upg{5.42} & 55.13 \upg{5.49} & 61.13 \upg{11.49} \\ \midrule
\Checkmark &   & \Checkmark &   &   & 52.99 & 54.92 \upg{1.93} & 55.70 \upg{2.71} & 53.94 \upg{0.95} & 57.77 \upg{4.78} & 53.90 \upg{0.91} & 56.51 \upg{3.52} & 57.68 \upg{4.69} & 57.75 \upg{4.76} & 64.03 \upg{11.05} \\ \midrule
\Checkmark &   &   & \Checkmark &   & 55.28 & 56.61 \upg{1.33} & 57.51 \upg{2.23} & 55.82 \upg{0.54} & 59.39 \upg{4.11} & 55.49 \upg{0.21} & 58.44 \upg{3.16} & 59.73 \upg{4.45} & 59.68 \upg{4.4} & 65.52 \upg{10.23} \\ \midrule
\Checkmark &   &   &   & \Checkmark & 62.06 & 61.65 \downr{-0.41} & 62.32 \upg{0.26} & 61.72 \downr{-0.35} & 63.96 \upg{1.9} & 61.01 \downr{-1.05} & 62.77 \upg{0.71} & 64.12 \upg{2.05} & 64.26 \upg{2.19} & 70.49 \upg{8.42} \\ \midrule
  & \Checkmark & \Checkmark &   &   & 52.99 & 54.94 \upg{1.95} & 55.49 \upg{2.5} & 48.72 \downr{-4.26} & 57.61 \upg{4.63} & 54.04 \upg{1.05} & 56.51 \upg{3.52} & 57.42 \upg{4.44} & 57.58 \upg{4.59} & 63.82 \upg{10.84} \\ \midrule
  & \Checkmark &   & \Checkmark &   & 55.28 & 56.56 \upg{1.28} & 57.16 \upg{1.88} & 56.25 \upg{0.97} & 58.78 \upg{3.5} & 56.27 \upg{0.98} & 58.08 \upg{2.8} & 58.80 \upg{3.52} & 59.25 \upg{3.97} & 65.34 \upg{10.06} \\ \midrule
  & \Checkmark &   &   & \Checkmark & 62.06 & 62.01 \downr{-0.05} & 62.70 \upg{0.64} & 62.03 \downr{-0.03} & 63.58 \upg{1.52} & 61.67 \downr{-0.4} & 63.24 \upg{1.17} & 63.69 \upg{1.62} & 63.93 \upg{1.86} & 70.31 \upg{8.25} \\ \midrule
  &   & \Checkmark & \Checkmark &   & 55.28 & 57.87 \upg{2.59} & 58.58 \upg{3.3} & 51.05 \downr{-4.23} & 60.87 \upg{5.59} & 57.40 \upg{2.12} & 59.63 \upg{4.35} & 60.72 \upg{5.44} & 60.74 \upg{5.45} & 67.10 \upg{11.82} \\ \midrule
  &   & \Checkmark &   & \Checkmark & 62.06 & 62.58 \upg{0.52} & 63.29 \upg{1.23} & 62.62 \upg{0.55} & 64.34 \upg{2.28} & 62.32 \upg{0.26} & 64.07 \upg{2.0} & 64.58 \upg{2.52} & 64.50 \upg{2.43} & 71.33 \upg{9.27} \\ \midrule
  &   &   & \Checkmark & \Checkmark & 62.06 & 62.63 \upg{0.57} & 63.36 \upg{1.29} & 54.87 \downr{-7.2} & 64.98 \upg{2.92} & 62.29 \upg{0.22} & 63.91 \upg{1.85} & 65.07 \upg{3.0} & 65.00 \upg{2.93} & 71.18 \upg{9.11} \\ \midrule
\Checkmark & \Checkmark & \Checkmark &   &   & 52.99 & 56.25 \upg{3.26} & 57.21 \upg{4.23} & 47.93 \downr{-5.06} & 59.32 \upg{6.33} & 54.50 \upg{1.52} & 58.18 \upg{5.2} & 59.60 \upg{6.61} & 59.25 \upg{6.27} & 69.71 \upg{16.72} \\ \midrule
\Checkmark & \Checkmark &   & \Checkmark &   & 55.28 & 57.34 \upg{2.05} & 58.42 \upg{3.14} & 53.31 \downr{-1.97} & 60.20 \upg{4.92} & 56.08 \upg{0.79} & 59.42 \upg{4.14} & 60.58 \upg{5.3} & 60.68 \upg{5.4} & 70.90 \upg{15.62} \\ \midrule
\Checkmark & \Checkmark &   &   & \Checkmark & 62.06 & 61.25 \downr{-0.81} & 62.81 \upg{0.74} & 59.89 \downr{-2.17} & 64.01 \upg{1.95} & 60.80 \downr{-1.26} & 63.41 \upg{1.35} & 65.08 \upg{3.02} & 65.03 \upg{2.97} & 74.75 \upg{12.69} \\ \midrule
\Checkmark &   & \Checkmark & \Checkmark &   & 55.28 & 58.94 \upg{3.66} & 60.11 \upg{4.83} & 51.86 \downr{-3.42} & 61.48 \upg{6.2} & 57.46 \upg{2.17} & 60.99 \upg{5.71} & 61.22 \upg{5.94} & 62.01 \upg{6.73} & 72.64 \upg{17.36} \\ \midrule
\Checkmark &   & \Checkmark &   & \Checkmark & 62.06 & 61.96 \downr{-0.1} & 63.88 \upg{1.81} & 62.39 \upg{0.33} & 65.36 \upg{3.3} & 61.60 \downr{-0.47} & 64.36 \upg{2.3} & 65.67 \upg{3.61} & 65.64 \upg{3.57} & 75.85 \upg{13.79} \\ \midrule
\Checkmark &   &   & \Checkmark & \Checkmark & 62.06 & 62.67 \upg{0.6} & 64.19 \upg{2.12} & 55.44 \downr{-6.63} & 65.67 \upg{3.61} & 61.72 \downr{-0.35} & 64.96 \upg{2.9} & 66.02 \upg{3.95} & 66.34 \upg{4.28} & 76.04 \upg{13.98} \\ \midrule
  & \Checkmark & \Checkmark & \Checkmark &   & 55.28 & 58.97 \upg{3.69} & 59.92 \upg{4.64} & 49.83 \downr{-5.45} & 61.24 \upg{5.95} & 57.73 \upg{2.45} & 60.68 \upg{5.4} & 61.18 \upg{5.9} & 61.65 \upg{6.37} & 71.99 \upg{16.71} \\ \midrule
  & \Checkmark & \Checkmark &   & \Checkmark & 62.06 & 62.15 \upg{0.09} & 63.44 \upg{1.38} & 60.25 \downr{-1.81} & 65.00 \upg{2.93} & 61.82 \downr{-0.24} & 63.98 \upg{1.92} & 65.36 \upg{3.3} & 65.31 \upg{3.24} & 75.42 \upg{13.36} \\ \midrule
  & \Checkmark &   & \Checkmark & \Checkmark & 62.06 & 62.43 \upg{0.36} & 64.03 \upg{1.97} & 55.47 \downr{-6.59} & 65.33 \upg{3.26} & 62.12 \upg{0.05} & 64.91 \upg{2.85} & 66.24 \upg{4.18} & 66.05 \upg{3.99} & 75.54 \upg{13.48} \\ \midrule
  &   & \Checkmark & \Checkmark & \Checkmark & 62.06 & 63.12 \upg{1.05} & 64.46 \upg{2.4} & 54.45 \downr{-7.61} & 65.93 \upg{3.87} & 62.77 \upg{0.71} & 64.79 \upg{2.73} & 66.05 \upg{3.99} & 66.05 \upg{3.99} & 76.60 \upg{14.53} \\ \midrule
\Checkmark & \Checkmark & \Checkmark & \Checkmark &   & 55.28 & 59.44 \upg{4.16} & 60.61 \upg{5.33} & 49.57 \downr{-5.71} & 61.81 \upg{6.52} & 57.49 \upg{2.21} & 61.39 \upg{6.11} & 61.96 \upg{6.68} & 62.50 \upg{7.21} & 75.60 \upg{20.31} \\ \midrule
\Checkmark & \Checkmark & \Checkmark &   & \Checkmark & 62.06 & 62.00 \downr{-0.07} & 63.62 \upg{1.55} & 59.37 \downr{-2.69} & 64.96 \upg{2.9} & 61.22 \downr{-0.85} & 64.10 \upg{2.04} & 65.38 \upg{3.31} & 65.91 \upg{3.85} & 78.32 \upg{16.26} \\ \midrule
\Checkmark & \Checkmark &   & \Checkmark & \Checkmark & 62.06 & 62.98 \upg{0.91} & 64.41 \upg{2.35} & 54.45 \downr{-7.61} & 65.22 \upg{3.16} & 61.49 \downr{-0.57} & 64.65 \upg{2.59} & 66.66 \upg{4.59} & 66.59 \upg{4.52} & 78.58 \upg{16.52} \\ \midrule
\Checkmark &   & \Checkmark & \Checkmark & \Checkmark & 62.06 & 63.63 \upg{1.57} & 65.15 \upg{3.09} & 54.73 \downr{-7.34} & 65.96 \upg{3.9} & 62.20 \upg{0.14} & 65.59 \upg{3.52} & 66.21 \upg{4.14} & 66.90 \upg{4.83} & 79.67 \upg{17.6} \\ \midrule
  & \Checkmark & \Checkmark & \Checkmark & \Checkmark & 62.06 & 63.19 \upg{1.12} & 64.38 \upg{2.31} & 54.18 \downr{-7.89} & 65.69 \upg{3.62} & 62.29 \upg{0.22} & 65.21 \upg{3.14} & 66.28 \upg{4.21} & 66.57 \upg{4.5} & 79.01 \upg{16.95} \\ \midrule
\Checkmark & \Checkmark & \Checkmark & \Checkmark & \Checkmark & 62.06 & 63.32 \upg{1.26} & 64.38 \upg{2.31} & 53.71 \downr{-8.35} & 65.53 \upg{3.47} & 61.79 \downr{-0.28} & 65.01 \upg{2.95} & 66.29 \upg{4.23} & 68.40 \upg{6.34} & 81.20 \upg{19.14} \\ \midrule

\multicolumn{5}{c}{Mean $\Delta$} & & 1.31 & 2.42 & -3.74 & 3.94 & 0.45 & 3.15 & 4.24 & 4.47 & 13.68 \\ 
\multicolumn{5}{c}{Max $\Delta$} & & 4.16 & 5.33 & 0.97 & 6.52 & 2.45 & 6.11 & 6.68 & 7.21 & 20.31 \\ 
\multicolumn{5}{c}{Min $\Delta$} & & -0.81 & 0.26 & -8.35 & 1.52 & -1.26 & 0.71 & 1.62 & 1.86 & 8.25 \\ 
\bottomrule
\end{tabular}
}
\caption{Our results on \CUB{} dataset for all the possible combinations of combining the zero-shot predictions of CLIP backbones, which we group intro non-parametric and parametric techniques. Also, the best-performing single backbone (\Best) and the \Oracle performance. We present, for each combination of backbones, the improvement \upg{}, constancy \samey{} and deterioration \downr{} of accuracy performance for each method when we compare it against the \Best{} backbone. Mean, Max, and Min $\Delta$ summarize the difference in performance across methods and backbone combinations.}
\label{tab:all_possible_combinations_cub}
\end{table*}

\begin{table*}[]
\centering
\DTD \\
\qquad
\resizebox{0.9\textwidth}{!}{

\begin{tabular}{ccccccccccccccc}
\toprule
\multicolumn{2}{c}{ResNet} & \multicolumn{3}{c}{ViT} & \multirow{2}{*}{\Best}  & \multicolumn{4}{c}{Non-Parametric} & \multicolumn{4}{c}{Parametric} & \multirow{2}{*}{\Oracle}\\
\cmidrule(lr){1-2} \cmidrule(lr){3-5} \cmidrule(lr){7-10} \cmidrule(lr){11-14}
50 & 101 & B-32 & B-16 & L-14 & & \VoteOne & \VoteThree & \Confidence & \LogitAvg & \CalibratedConfidence & \CLogitAvg & \GAC & \NNC &  \\
\midrule

\Checkmark &   &   &   &   &  \multicolumn{10}{c}{\xfill{.1em} 41.22 \samey{0.00} \xfill{.1em}} \\ 
  & \Checkmark &   &   &   & \multicolumn{10}{c}{\xfill{.1em}  43.67 \samey{0.00} \xfill{.1em}} \\ 
  &   & \Checkmark &   &   & \multicolumn{10}{c}{\xfill{.1em}  43.99 \samey{0.00} \xfill{.1em}} \\ 
  &   &   & \Checkmark &   & \multicolumn{10}{c}{\xfill{.1em}  45.11 \samey{0.00} \xfill{.1em}} \\ 
  &   &   &   & \Checkmark & \multicolumn{10}{c}{\xfill{.1em}  55.32 \samey{0.00} \xfill{.1em}} \\ \midrule
\Checkmark & \Checkmark &   &   &   & 43.67 & 47.13 \upg{3.46} & 47.29 \upg{3.62} & 46.44 \upg{2.77} & 47.71 \upg{4.04} & 45.96 \upg{2.29} & 47.34 \upg{3.67} & 47.61 \upg{3.94} & 47.71 \upg{4.04} & 54.36 \upg{10.69} \\ \midrule
\Checkmark &   & \Checkmark &   &   & 43.99 & 44.63 \upg{0.64} & 45.00 \upg{1.01} & 41.22 \downr{-2.77} & 46.01 \upg{2.02} & 43.78 \downr{-0.21} & 45.43 \upg{1.44} & 45.74 \upg{1.76} & 45.90 \upg{1.91} & 51.70 \upg{7.71} \\ \midrule
\Checkmark &   &   & \Checkmark &   & 45.11 & 46.86 \upg{1.76} & 47.13 \upg{2.02} & 40.37 \downr{-4.73} & 47.39 \upg{2.29} & 45.85 \upg{0.74} & 47.23 \upg{2.13} & 47.87 \upg{2.77} & 47.39 \upg{2.29} & 53.24 \upg{8.14} \\ \midrule
\Checkmark &   &   &   & \Checkmark & 55.32 & 55.64 \upg{0.32} & 56.12 \upg{0.8} & 41.33 \downr{-13.99} & 55.64 \upg{0.32} & 55.05 \downr{-0.27} & 56.22 \upg{0.9} & 55.43 \upg{0.11} & 56.44 \upg{1.12} & 62.23 \upg{6.91} \\ \midrule
  & \Checkmark & \Checkmark &   &   & 43.99 & 47.02 \upg{3.03} & 47.34 \upg{3.35} & 41.60 \downr{-2.39} & 48.46 \upg{4.47} & 47.13 \upg{3.14} & 47.93 \upg{3.94} & 47.87 \upg{3.88} & 47.98 \upg{3.99} & 55.11 \upg{11.12} \\ \midrule
  & \Checkmark &   & \Checkmark &   & 45.11 & 48.46 \upg{3.35} & 48.94 \upg{3.83} & 48.30 \upg{3.19} & 49.15 \upg{4.04} & 48.35 \upg{3.24} & 48.67 \upg{3.56} & 48.78 \upg{3.67} & 48.78 \upg{3.67} & 55.74 \upg{10.64} \\ \midrule
  & \Checkmark &   &   & \Checkmark & 55.32 & 55.21 \downr{-0.11} & 55.59 \upg{0.27} & 43.35 \downr{-11.97} & 55.69 \upg{0.37} & 55.11 \downr{-0.21} & 56.33 \upg{1.01} & 56.76 \upg{1.44} & 56.70 \upg{1.38} & 61.60 \upg{6.28} \\ \midrule
  &   & \Checkmark & \Checkmark &   & 45.11 & 46.65 \upg{1.54} & 46.86 \upg{1.76} & 43.24 \downr{-1.86} & 47.71 \upg{2.61} & 45.48 \upg{0.37} & 47.23 \upg{2.13} & 47.07 \upg{1.97} & 47.61 \upg{2.5} & 53.56 \upg{8.46} \\ \midrule
  &   & \Checkmark &   & \Checkmark & 55.32 & 55.74 \upg{0.43} & 56.33 \upg{1.01} & 43.56 \downr{-11.76} & 56.22 \upg{0.9} & 55.32 \samey{0.0} & 56.22 \upg{0.9} & 56.49 \upg{1.17} & 56.70 \upg{1.38} & 62.02 \upg{6.7} \\ \midrule
  &   &   & \Checkmark & \Checkmark & 55.32 & 57.02 \upg{1.7} & 57.39 \upg{2.07} & 43.88 \downr{-11.44} & 56.28 \upg{0.96} & 56.60 \upg{1.28} & 56.76 \upg{1.44} & 56.81 \upg{1.49} & 56.86 \upg{1.54} & 62.02 \upg{6.7} \\ \midrule
\Checkmark & \Checkmark & \Checkmark &   &   & 43.99 & 47.50 \upg{3.51} & 47.71 \upg{3.72} & 40.37 \downr{-3.62} & 48.99 \upg{5.0} & 46.97 \upg{2.98} & 47.82 \upg{3.83} & 49.26 \upg{5.27} & 49.10 \upg{5.11} & 59.26 \upg{15.27} \\ \midrule
\Checkmark & \Checkmark &   & \Checkmark &   & 45.11 & 48.35 \upg{3.24} & 49.41 \upg{4.31} & 44.68 \downr{-0.43} & 49.95 \upg{4.84} & 48.30 \upg{3.19} & 49.95 \upg{4.84} & 50.32 \upg{5.21} & 50.21 \upg{5.11} & 60.59 \upg{15.48} \\ \midrule
\Checkmark & \Checkmark &   &   & \Checkmark & 55.32 & 54.26 \downr{-1.06} & 56.22 \upg{0.9} & 41.70 \downr{-13.62} & 55.32 \samey{0.0} & 54.89 \downr{-0.43} & 56.76 \upg{1.44} & 57.02 \upg{1.7} & 57.18 \upg{1.86} & 65.80 \upg{10.48} \\ \midrule
\Checkmark &   & \Checkmark & \Checkmark &   & 45.11 & 46.86 \upg{1.76} & 46.97 \upg{1.86} & 44.95 \downr{-0.16} & 48.19 \upg{3.09} & 45.53 \upg{0.43} & 47.34 \upg{2.23} & 47.77 \upg{2.66} & 48.24 \upg{3.14} & 57.87 \upg{12.77} \\ \midrule
\Checkmark &   & \Checkmark &   & \Checkmark & 55.32 & 53.19 \downr{-2.13} & 55.43 \upg{0.11} & 41.86 \downr{-13.46} & 54.73 \downr{-0.59} & 55.11 \downr{-0.21} & 54.73 \downr{-0.59} & 56.06 \upg{0.74} & 56.49 \upg{1.17} & 65.59 \upg{10.27} \\ \midrule
\Checkmark &   &   & \Checkmark & \Checkmark & 55.32 & 53.99 \downr{-1.33} & 56.33 \upg{1.01} & 42.87 \downr{-12.45} & 56.01 \upg{0.69} & 55.80 \upg{0.48} & 56.60 \upg{1.28} & 57.34 \upg{2.02} & 57.18 \upg{1.86} & 65.69 \upg{10.37} \\ \midrule
  & \Checkmark & \Checkmark & \Checkmark &   & 45.11 & 48.40 \upg{3.3} & 49.68 \upg{4.57} & 41.76 \downr{-3.35} & 49.26 \upg{4.15} & 48.62 \upg{3.51} & 49.68 \upg{4.57} & 49.95 \upg{4.84} & 49.20 \upg{4.1} & 60.32 \upg{15.21} \\ \midrule
  & \Checkmark & \Checkmark &   & \Checkmark & 55.32 & 54.47 \downr{-0.85} & 55.69 \upg{0.37} & 44.57 \downr{-10.74} & 55.27 \downr{-0.05} & 55.32 \samey{0.0} & 56.01 \upg{0.69} & 56.86 \upg{1.54} & 57.18 \upg{1.86} & 65.32 \upg{10.0} \\ \midrule
  & \Checkmark &   & \Checkmark & \Checkmark & 55.32 & 54.89 \downr{-0.43} & 56.76 \upg{1.44} & 43.56 \downr{-11.76} & 55.74 \upg{0.43} & 56.22 \upg{0.9} & 56.81 \upg{1.49} & 56.70 \upg{1.38} & 57.77 \upg{2.45} & 65.37 \upg{10.05} \\ \midrule
  &   & \Checkmark & \Checkmark & \Checkmark & 55.32 & 54.63 \downr{-0.69} & 56.44 \upg{1.12} & 44.47 \downr{-10.85} & 55.74 \upg{0.43} & 55.74 \upg{0.43} & 56.49 \upg{1.17} & 57.23 \upg{1.91} & 57.07 \upg{1.76} & 65.32 \upg{10.0} \\ \midrule
\Checkmark & \Checkmark & \Checkmark & \Checkmark &   & 45.11 & 48.67 \upg{3.56} & 49.52 \upg{4.41} & 43.30 \downr{-1.81} & 49.63 \upg{4.52} & 48.46 \upg{3.35} & 49.79 \upg{4.68} & 50.27 \upg{5.16} & 49.41 \upg{4.31} & 63.19 \upg{18.09} \\ \midrule
\Checkmark & \Checkmark & \Checkmark &   & \Checkmark & 55.32 & 54.36 \downr{-0.96} & 55.16 \downr{-0.16} & 42.50 \downr{-12.82} & 54.73 \downr{-0.59} & 55.00 \downr{-0.32} & 54.95 \downr{-0.37} & 56.22 \upg{0.9} & 57.34 \upg{2.02} & 68.03 \upg{12.71} \\ \midrule
\Checkmark & \Checkmark &   & \Checkmark & \Checkmark & 55.32 & 54.89 \downr{-0.43} & 56.38 \upg{1.06} & 42.93 \downr{-12.39} & 55.00 \downr{-0.32} & 55.53 \upg{0.21} & 56.81 \upg{1.49} & 57.87 \upg{2.55} & 57.71 \upg{2.39} & 68.14 \upg{12.82} \\ \midrule
\Checkmark &   & \Checkmark & \Checkmark & \Checkmark & 55.32 & 53.72 \downr{-1.6} & 55.96 \upg{0.64} & 42.50 \downr{-12.82} & 54.57 \downr{-0.74} & 55.48 \upg{0.16} & 55.48 \upg{0.16} & 56.38 \upg{1.06} & 56.97 \upg{1.65} & 67.71 \upg{12.39} \\ \midrule
  & \Checkmark & \Checkmark & \Checkmark & \Checkmark & 55.32 & 54.68 \downr{-0.64} & 56.12 \upg{0.8} & 44.47 \downr{-10.85} & 55.37 \upg{0.05} & 55.96 \upg{0.64} & 56.06 \upg{0.74} & 56.81 \upg{1.49} & 57.50 \upg{2.18} & 67.66 \upg{12.34} \\ \midrule
\Checkmark & \Checkmark & \Checkmark & \Checkmark & \Checkmark & 55.32 & 53.72 \downr{-1.6} & 55.69 \upg{0.37} & 42.55 \downr{-12.77} & 54.36 \downr{-0.96} & 55.59 \upg{0.27} & 55.21 \downr{-0.11} & 56.12 \upg{0.8} & 58.94 \upg{3.62} & 69.63 \upg{14.31} \\ \midrule

\multicolumn{5}{c}{Mean $\Delta$} & & 0.76 & 1.78 & -7.65 & 1.61 & 1.00 & 1.87 & 2.36 & 2.63 & 11.00 \\ 
\multicolumn{5}{c}{Max $\Delta$} & & 3.56 & 4.57 & 3.19 & 5.00 & 3.51 & 4.84 & 5.27 & 5.11 & 18.09 \\ 
\multicolumn{5}{c}{Min $\Delta$} & & -2.13 & -0.16 & -13.99 & -0.96 & -0.43 & -0.59 & 0.11 & 1.12 & 6.28 \\ 
\bottomrule
\end{tabular}
}
\caption{Our results on \DTD{} dataset for all the possible combinations of combining the zero-shot predictions of CLIP backbones, which we group intro non-parametric and parametric techniques. Also, the best-performing single backbone (\Best) and the \Oracle performance. We present, for each combination of backbones, the improvement \upg{}, constancy \samey{} and deterioration \downr{} of accuracy performance for each method when we compare it against the \Best{} backbone. Mean, Max, and Min $\Delta$ summarize the difference in performance across methods and backbone combinations.}
\label{tab:all_possible_combinations_dtd}
\end{table*}

\begin{table*}[]
\centering
\FGVC \\
\qquad
\resizebox{0.9\textwidth}{!}{

\begin{tabular}{ccccccccccccccc}
\toprule
\multicolumn{2}{c}{ResNet} & \multicolumn{3}{c}{ViT} & \multirow{2}{*}{\Best}  & \multicolumn{4}{c}{Non-Parametric} & \multicolumn{4}{c}{Parametric} & \multirow{2}{*}{\Oracle}\\
\cmidrule(lr){1-2} \cmidrule(lr){3-5} \cmidrule(lr){7-10} \cmidrule(lr){11-14}
50 & 101 & B-32 & B-16 & L-14 & & \VoteOne & \VoteThree & \Confidence & \LogitAvg & \CalibratedConfidence & \CLogitAvg & \GAC & \NNC &  \\
\midrule

\Checkmark &   &   &   &   &  \multicolumn{10}{c}{\xfill{.1em} 17.07 \samey{0.00} \xfill{.1em}} \\ 
  & \Checkmark &   &   &   & \multicolumn{10}{c}{\xfill{.1em}  18.63 \samey{0.00} \xfill{.1em}} \\ 
  &   & \Checkmark &   &   & \multicolumn{10}{c}{\xfill{.1em}  19.65 \samey{0.00} \xfill{.1em}} \\ 
  &   &   & \Checkmark &   & \multicolumn{10}{c}{\xfill{.1em}  24.39 \samey{0.00} \xfill{.1em}} \\ 
  &   &   &   & \Checkmark & \multicolumn{10}{c}{\xfill{.1em}  31.71 \samey{0.00} \xfill{.1em}} \\ \midrule
\Checkmark & \Checkmark &   &   &   & 18.63 & 18.87 \upg{0.24} & 19.02 \upg{0.39} & 18.60 \downr{-0.03} & 19.23 \upg{0.6} & 18.48 \downr{-0.15} & 19.05 \upg{0.42} & 19.59 \upg{0.96} & 19.89 \upg{1.26} & 26.19 \upg{7.56} \\ \midrule
\Checkmark &   & \Checkmark &   &   & 19.65 & 20.52 \upg{0.87} & 20.46 \upg{0.81} & 16.62 \downr{-3.03} & 21.36 \upg{1.71} & 19.80 \upg{0.15} & 21.18 \upg{1.53} & 21.00 \upg{1.35} & 22.17 \upg{2.52} & 27.69 \upg{8.04} \\ \midrule
\Checkmark &   &   & \Checkmark &   & 24.39 & 24.24 \downr{-0.15} & 24.33 \downr{-0.06} & 17.19 \downr{-7.2} & 24.78 \upg{0.39} & 23.97 \downr{-0.42} & 25.23 \upg{0.84} & 25.65 \upg{1.26} & 25.89 \upg{1.5} & 31.02 \upg{6.63} \\ \midrule
\Checkmark &   &   &   & \Checkmark & 31.71 & 31.47 \downr{-0.24} & 31.47 \downr{-0.24} & 31.47 \downr{-0.24} & 31.68 \downr{-0.03} & 31.59 \downr{-0.12} & 32.55 \upg{0.84} & 32.49 \upg{0.78} & 33.57 \upg{1.86} & 37.59 \upg{5.88} \\ \midrule
  & \Checkmark & \Checkmark &   &   & 19.65 & 21.00 \upg{1.35} & 21.51 \upg{1.86} & 20.70 \upg{1.05} & 21.69 \upg{2.04} & 20.73 \upg{1.08} & 22.02 \upg{2.37} & 21.21 \upg{1.56} & 22.41 \upg{2.76} & 28.20 \upg{8.55} \\ \midrule
  & \Checkmark &   & \Checkmark &   & 24.39 & 23.67 \downr{-0.72} & 23.97 \downr{-0.42} & 23.52 \downr{-0.87} & 23.82 \downr{-0.57} & 23.79 \downr{-0.6} & 24.36 \downr{-0.03} & 24.81 \upg{0.42} & 25.38 \upg{0.99} & 31.92 \upg{7.53} \\ \midrule
  & \Checkmark &   &   & \Checkmark & 31.71 & 30.81 \downr{-0.9} & 31.29 \downr{-0.42} & 30.93 \downr{-0.78} & 30.54 \downr{-1.17} & 31.20 \downr{-0.51} & 31.50 \downr{-0.21} & 31.35 \downr{-0.36} & 33.03 \upg{1.32} & 38.31 \upg{6.6} \\ \midrule
  &   & \Checkmark & \Checkmark &   & 24.39 & 24.18 \downr{-0.21} & 24.30 \downr{-0.09} & 20.22 \downr{-4.17} & 25.08 \upg{0.69} & 23.76 \downr{-0.63} & 24.84 \upg{0.45} & 25.02 \upg{0.63} & 25.95 \upg{1.56} & 32.73 \upg{8.34} \\ \midrule
  &   & \Checkmark &   & \Checkmark & 31.71 & 31.86 \upg{0.15} & 31.68 \downr{-0.03} & 31.59 \downr{-0.12} & 32.04 \upg{0.33} & 31.71 \samey{-0.0} & 32.79 \upg{1.08} & 32.61 \upg{0.9} & 33.33 \upg{1.62} & 39.57 \upg{7.86} \\ \midrule
  &   &   & \Checkmark & \Checkmark & 31.71 & 31.62 \downr{-0.09} & 32.19 \upg{0.48} & 31.95 \upg{0.24} & 32.76 \upg{1.05} & 31.92 \upg{0.21} & 32.88 \upg{1.17} & 33.30 \upg{1.59} & 33.69 \upg{1.98} & 41.25 \upg{9.54} \\ \midrule
\Checkmark & \Checkmark & \Checkmark &   &   & 19.65 & 20.79 \upg{1.14} & 21.51 \upg{1.86} & 18.12 \downr{-1.53} & 21.75 \upg{2.1} & 20.52 \upg{0.87} & 21.69 \upg{2.04} & 21.72 \upg{2.07} & 23.49 \upg{3.84} & 33.60 \upg{13.95} \\ \midrule
\Checkmark & \Checkmark &   & \Checkmark &   & 24.39 & 22.89 \downr{-1.5} & 23.28 \downr{-1.11} & 17.55 \downr{-6.84} & 23.67 \downr{-0.72} & 23.40 \downr{-0.99} & 24.03 \downr{-0.36} & 24.69 \upg{0.3} & 26.61 \upg{2.22} & 36.57 \upg{12.18} \\ \midrule
\Checkmark & \Checkmark &   &   & \Checkmark & 31.71 & 28.80 \downr{-2.91} & 30.00 \downr{-1.71} & 30.87 \downr{-0.84} & 30.21 \downr{-1.5} & 31.14 \downr{-0.57} & 31.53 \downr{-0.18} & 32.58 \upg{0.87} & 34.08 \upg{2.37} & 42.36 \upg{10.65} \\ \midrule
\Checkmark &   & \Checkmark & \Checkmark &   & 24.39 & 23.94 \downr{-0.45} & 24.57 \upg{0.18} & 22.53 \downr{-1.86} & 24.81 \upg{0.42} & 23.61 \downr{-0.78} & 24.99 \upg{0.6} & 25.71 \upg{1.32} & 26.76 \upg{2.37} & 37.68 \upg{13.29} \\ \midrule
\Checkmark &   & \Checkmark &   & \Checkmark & 31.71 & 30.36 \downr{-1.35} & 31.29 \downr{-0.42} & 30.93 \downr{-0.78} & 32.07 \upg{0.36} & 31.59 \downr{-0.12} & 32.73 \upg{1.02} & 32.25 \upg{0.54} & 34.74 \upg{3.03} & 43.53 \upg{11.82} \\ \midrule
\Checkmark &   &   & \Checkmark & \Checkmark & 31.71 & 30.63 \downr{-1.08} & 31.77 \upg{0.06} & 30.27 \downr{-1.44} & 32.40 \upg{0.69} & 31.80 \upg{0.09} & 32.97 \upg{1.26} & 33.06 \upg{1.35} & 34.80 \upg{3.09} & 45.06 \upg{13.35} \\ \midrule
  & \Checkmark & \Checkmark & \Checkmark &   & 24.39 & 23.79 \downr{-0.6} & 24.15 \downr{-0.24} & 20.55 \downr{-3.84} & 25.02 \upg{0.63} & 23.40 \downr{-0.99} & 24.72 \upg{0.33} & 25.50 \upg{1.11} & 26.31 \upg{1.92} & 38.10 \upg{13.71} \\ \midrule
  & \Checkmark & \Checkmark &   & \Checkmark & 31.71 & 30.24 \downr{-1.47} & 30.90 \downr{-0.81} & 30.90 \downr{-0.81} & 31.62 \downr{-0.09} & 31.20 \downr{-0.51} & 32.07 \upg{0.36} & 31.95 \upg{0.24} & 33.72 \upg{2.01} & 44.01 \upg{12.3} \\ \midrule
  & \Checkmark &   & \Checkmark & \Checkmark & 31.71 & 30.75 \downr{-0.96} & 31.62 \downr{-0.09} & 31.23 \downr{-0.48} & 31.77 \upg{0.06} & 31.44 \downr{-0.27} & 32.40 \upg{0.69} & 33.21 \upg{1.5} & 34.23 \upg{2.52} & 45.69 \upg{13.98} \\ \midrule
  &   & \Checkmark & \Checkmark & \Checkmark & 31.71 & 31.14 \downr{-0.57} & 31.59 \downr{-0.12} & 31.02 \downr{-0.69} & 32.91 \upg{1.2} & 31.77 \upg{0.06} & 33.39 \upg{1.68} & 33.18 \upg{1.47} & 34.50 \upg{2.79} & 46.56 \upg{14.85} \\ \midrule
\Checkmark & \Checkmark & \Checkmark & \Checkmark &   & 24.39 & 23.97 \downr{-0.42} & 23.91 \downr{-0.48} & 22.11 \downr{-2.28} & 24.72 \upg{0.33} & 23.31 \downr{-1.08} & 24.60 \upg{0.21} & 25.50 \upg{1.11} & 27.45 \upg{3.06} & 41.79 \upg{17.4} \\ \midrule
\Checkmark & \Checkmark & \Checkmark &   & \Checkmark & 31.71 & 28.65 \downr{-3.06} & 30.63 \downr{-1.08} & 30.45 \downr{-1.26} & 31.20 \downr{-0.51} & 31.14 \downr{-0.57} & 32.04 \upg{0.33} & 32.37 \upg{0.66} & 34.80 \upg{3.09} & 47.04 \upg{15.33} \\ \midrule
\Checkmark & \Checkmark &   & \Checkmark & \Checkmark & 31.71 & 29.46 \downr{-2.25} & 30.72 \downr{-0.99} & 29.85 \downr{-1.86} & 31.02 \downr{-0.69} & 31.38 \downr{-0.33} & 32.31 \upg{0.6} & 32.55 \upg{0.84} & 34.92 \upg{3.21} & 48.54 \upg{16.83} \\ \midrule
\Checkmark &   & \Checkmark & \Checkmark & \Checkmark & 31.71 & 30.06 \downr{-1.65} & 31.44 \downr{-0.27} & 31.02 \downr{-0.69} & 32.28 \upg{0.57} & 31.65 \downr{-0.06} & 33.21 \upg{1.5} & 33.63 \upg{1.92} & 35.37 \upg{3.66} & 49.38 \upg{17.67} \\ \midrule
  & \Checkmark & \Checkmark & \Checkmark & \Checkmark & 31.71 & 29.82 \downr{-1.89} & 31.20 \downr{-0.51} & 30.51 \downr{-1.2} & 32.58 \upg{0.87} & 31.35 \downr{-0.36} & 32.25 \upg{0.54} & 33.33 \upg{1.62} & 34.23 \upg{2.52} & 49.83 \upg{18.12} \\ \midrule
\Checkmark & \Checkmark & \Checkmark & \Checkmark & \Checkmark & 31.71 & 28.95 \downr{-2.76} & 30.96 \downr{-0.75} & 30.63 \downr{-1.08} & 31.53 \downr{-0.18} & 31.29 \downr{-0.42} & 31.89 \upg{0.18} & 33.18 \upg{1.47} & 35.88 \upg{4.17} & 52.09 \upg{20.37} \\ \midrule

\multicolumn{5}{c}{Mean $\Delta$} & & -0.83 & -0.16 & -1.64 & 0.33 & -0.27 & 0.74 & 1.06 & 2.43 & 12.01 \\ 
\multicolumn{5}{c}{Max $\Delta$} & & 1.35 & 1.86 & 1.05 & 2.10 & 1.08 & 2.37 & 2.07 & 4.17 & 20.37 \\ 
\multicolumn{5}{c}{Min $\Delta$} & & -3.06 & -1.71 & -7.20 & -1.50 & -1.08 & -0.36 & -0.36 & 0.99 & 5.88 \\ 
\bottomrule
\end{tabular}
}
\caption{Our results on \FGVC{} dataset for all the possible combinations of combining the zero-shot predictions of CLIP backbones, which we group intro non-parametric and parametric techniques. Also, the best-performing single backbone (\Best) and the \Oracle performance. We present, for each combination of backbones, the improvement \upg{}, constancy \samey{} and deterioration \downr{} of accuracy performance for each method when we compare it against the \Best{} backbone. Mean, Max, and Min $\Delta$ summarize the difference in performance across methods and backbone combinations.}
\label{tab:all_possible_combinations_fgvc}
\end{table*}

\begin{table*}[]
\centering
\Food \\
\qquad
\resizebox{0.9\textwidth}{!}{

\begin{tabular}{ccccccccccccccc}
\toprule
\multicolumn{2}{c}{ResNet} & \multicolumn{3}{c}{ViT} & \multirow{2}{*}{\Best}  & \multicolumn{4}{c}{Non-Parametric} & \multicolumn{4}{c}{Parametric} & \multirow{2}{*}{\Oracle}\\
\cmidrule(lr){1-2} \cmidrule(lr){3-5} \cmidrule(lr){7-10} \cmidrule(lr){11-14}
50 & 101 & B-32 & B-16 & L-14 & & \VoteOne & \VoteThree & \Confidence & \LogitAvg & \CalibratedConfidence & \CLogitAvg & \GAC & \NNC &  \\
\midrule

\Checkmark &   &   &   &   &  \multicolumn{10}{c}{\xfill{.1em} 77.91 \samey{0.00} \xfill{.1em}} \\ 
  & \Checkmark &   &   &   & \multicolumn{10}{c}{\xfill{.1em}  81.86 \samey{0.00} \xfill{.1em}} \\ 
  &   & \Checkmark &   &   & \multicolumn{10}{c}{\xfill{.1em}  82.58 \samey{0.00} \xfill{.1em}} \\ 
  &   &   & \Checkmark &   & \multicolumn{10}{c}{\xfill{.1em}  87.91 \samey{0.00} \xfill{.1em}} \\ 
  &   &   &   & \Checkmark & \multicolumn{10}{c}{\xfill{.1em}  92.32 \samey{0.00} \xfill{.1em}} \\ \midrule
\Checkmark & \Checkmark &   &   &   & 81.86 & 82.63 \upg{0.78} & 82.90 \upg{1.04} & 82.26 \upg{0.4} & 83.08 \upg{1.22} & 82.47 \upg{0.61} & 83.14 \upg{1.29} & 83.28 \upg{1.43} & 83.27 \upg{1.41} & 87.16 \upg{5.3} \\ \midrule
\Checkmark &   & \Checkmark &   &   & 82.58 & 83.83 \upg{1.25} & 84.26 \upg{1.68} & 76.96 \downr{-5.62} & 84.55 \upg{1.97} & 83.60 \upg{1.02} & 84.40 \upg{1.82} & 84.41 \upg{1.83} & 84.68 \upg{2.1} & 88.42 \upg{5.84} \\ \midrule
\Checkmark &   &   & \Checkmark &   & 87.91 & 87.56 \downr{-0.36} & 87.62 \downr{-0.3} & 87.49 \downr{-0.43} & 87.58 \downr{-0.33} & 87.52 \downr{-0.4} & 87.80 \downr{-0.11} & 88.13 \upg{0.22} & 88.55 \upg{0.64} & 91.31 \upg{3.39} \\ \midrule
\Checkmark &   &   &   & \Checkmark & 92.32 & 91.94 \downr{-0.38} & 92.04 \downr{-0.29} & 91.88 \downr{-0.44} & 91.75 \downr{-0.57} & 91.75 \downr{-0.57} & 91.93 \downr{-0.39} & 92.82 \upg{0.49} & 92.79 \upg{0.47} & 94.52 \upg{2.19} \\ \midrule
  & \Checkmark & \Checkmark &   &   & 82.58 & 84.97 \upg{2.39} & 85.31 \upg{2.73} & 84.48 \upg{1.9} & 85.70 \upg{3.11} & 84.62 \upg{2.04} & 85.54 \upg{2.96} & 85.55 \upg{2.97} & 85.69 \upg{3.11} & 89.28 \upg{6.69} \\ \midrule
  & \Checkmark &   & \Checkmark &   & 87.91 & 88.17 \upg{0.26} & 88.30 \upg{0.38} & 87.92 \upg{0.01} & 88.40 \upg{0.49} & 87.87 \downr{-0.05} & 88.34 \upg{0.43} & 88.72 \upg{0.81} & 88.68 \upg{0.76} & 91.81 \upg{3.9} \\ \midrule
  & \Checkmark &   &   & \Checkmark & 92.32 & 92.25 \downr{-0.07} & 92.34 \upg{0.02} & 92.04 \downr{-0.29} & 92.08 \downr{-0.24} & 91.85 \downr{-0.48} & 92.22 \downr{-0.11} & 92.78 \upg{0.45} & 92.80 \upg{0.48} & 94.81 \upg{2.49} \\ \midrule
  &   & \Checkmark & \Checkmark &   & 87.91 & 88.08 \upg{0.16} & 88.18 \upg{0.27} & 82.55 \downr{-5.36} & 88.25 \upg{0.33} & 88.10 \upg{0.19} & 88.31 \upg{0.4} & 88.79 \upg{0.88} & 88.74 \upg{0.83} & 91.54 \upg{3.63} \\ \midrule
  &   & \Checkmark &   & \Checkmark & 92.32 & 92.13 \downr{-0.2} & 92.21 \downr{-0.12} & 92.00 \downr{-0.33} & 92.04 \downr{-0.28} & 91.92 \downr{-0.4} & 92.14 \downr{-0.19} & 92.70 \upg{0.37} & 92.73 \upg{0.41} & 94.76 \upg{2.44} \\ \midrule
  &   &   & \Checkmark & \Checkmark & 92.32 & 92.58 \upg{0.25} & 92.67 \upg{0.34} & 92.45 \upg{0.12} & 92.57 \upg{0.24} & 92.31 \downr{-0.01} & 92.59 \upg{0.27} & 92.70 \upg{0.38} & 92.88 \upg{0.55} & 94.96 \upg{2.64} \\ \midrule
\Checkmark & \Checkmark & \Checkmark &   &   & 82.58 & 84.54 \upg{1.96} & 85.28 \upg{2.7} & 79.64 \downr{-2.95} & 85.59 \upg{3.01} & 84.62 \upg{2.04} & 85.38 \upg{2.8} & 85.82 \upg{3.24} & 85.84 \upg{3.26} & 91.16 \upg{8.58} \\ \midrule
\Checkmark & \Checkmark &   & \Checkmark &   & 87.91 & 86.54 \downr{-1.37} & 87.34 \downr{-0.57} & 87.57 \downr{-0.34} & 87.72 \downr{-0.19} & 87.51 \downr{-0.4} & 87.63 \downr{-0.28} & 88.62 \upg{0.71} & 88.73 \upg{0.82} & 92.99 \upg{5.07} \\ \midrule
\Checkmark & \Checkmark &   &   & \Checkmark & 92.32 & 88.86 \downr{-3.47} & 90.41 \downr{-1.91} & 91.66 \downr{-0.67} & 90.99 \downr{-1.34} & 91.45 \downr{-0.87} & 90.59 \downr{-1.74} & 92.79 \upg{0.47} & 92.84 \upg{0.52} & 95.47 \upg{3.15} \\ \midrule
\Checkmark &   & \Checkmark & \Checkmark &   & 87.91 & 87.45 \downr{-0.47} & 87.98 \upg{0.07} & 80.07 \downr{-7.85} & 87.97 \upg{0.06} & 87.77 \downr{-0.15} & 88.09 \upg{0.17} & 88.84 \upg{0.93} & 88.85 \upg{0.94} & 93.01 \upg{5.09} \\ \midrule
\Checkmark &   & \Checkmark &   & \Checkmark & 92.32 & 89.70 \downr{-2.63} & 90.90 \downr{-1.43} & 90.01 \downr{-2.32} & 91.18 \downr{-1.15} & 91.54 \downr{-0.79} & 90.93 \downr{-1.39} & 92.71 \upg{0.38} & 92.83 \upg{0.5} & 95.56 \upg{3.23} \\ \midrule
\Checkmark &   &   & \Checkmark & \Checkmark & 92.32 & 91.38 \downr{-0.94} & 92.00 \downr{-0.33} & 92.11 \downr{-0.21} & 91.98 \downr{-0.34} & 91.93 \downr{-0.4} & 92.00 \downr{-0.33} & 92.77 \upg{0.44} & 93.02 \upg{0.69} & 95.77 \upg{3.45} \\ \midrule
  & \Checkmark & \Checkmark & \Checkmark &   & 87.91 & 87.78 \downr{-0.13} & 88.18 \upg{0.27} & 83.45 \downr{-4.46} & 88.40 \upg{0.48} & 87.96 \upg{0.05} & 88.32 \upg{0.4} & 89.02 \upg{1.1} & 88.95 \upg{1.04} & 93.36 \upg{5.45} \\ \midrule
  & \Checkmark & \Checkmark &   & \Checkmark & 92.32 & 90.17 \downr{-2.15} & 91.20 \downr{-1.13} & 91.75 \downr{-0.57} & 91.50 \downr{-0.82} & 91.54 \downr{-0.78} & 91.28 \downr{-1.05} & 92.90 \upg{0.57} & 92.93 \upg{0.61} & 95.71 \upg{3.39} \\ \midrule
  & \Checkmark &   & \Checkmark & \Checkmark & 92.32 & 91.54 \downr{-0.78} & 91.98 \downr{-0.34} & 92.06 \downr{-0.26} & 92.07 \downr{-0.25} & 91.87 \downr{-0.45} & 92.02 \downr{-0.3} & 92.93 \upg{0.6} & 92.97 \upg{0.65} & 95.89 \upg{3.56} \\ \midrule
  &   & \Checkmark & \Checkmark & \Checkmark & 92.32 & 91.24 \downr{-1.08} & 91.85 \downr{-0.48} & 90.73 \downr{-1.59} & 92.02 \downr{-0.31} & 92.04 \downr{-0.28} & 91.92 \downr{-0.4} & 92.93 \upg{0.61} & 92.98 \upg{0.66} & 95.79 \upg{3.47} \\ \midrule
\Checkmark & \Checkmark & \Checkmark & \Checkmark &   & 87.91 & 87.65 \downr{-0.26} & 87.98 \upg{0.07} & 81.00 \downr{-6.91} & 88.11 \upg{0.2} & 87.66 \downr{-0.25} & 88.10 \upg{0.19} & 89.07 \upg{1.16} & 89.03 \upg{1.12} & 94.01 \upg{6.1} \\ \midrule
\Checkmark & \Checkmark & \Checkmark &   & \Checkmark & 92.32 & 89.57 \downr{-2.75} & 90.21 \downr{-2.12} & 90.17 \downr{-2.15} & 90.61 \downr{-1.71} & 91.24 \downr{-1.09} & 90.24 \downr{-2.09} & 92.49 \upg{0.17} & 92.90 \upg{0.58} & 96.10 \upg{3.77} \\ \midrule
\Checkmark & \Checkmark &   & \Checkmark & \Checkmark & 92.32 & 91.10 \downr{-1.22} & 91.29 \downr{-1.03} & 91.83 \downr{-0.5} & 91.38 \downr{-0.95} & 91.60 \downr{-0.73} & 91.26 \downr{-1.07} & 92.93 \upg{0.61} & 93.03 \upg{0.7} & 96.26 \upg{3.94} \\ \midrule
\Checkmark &   & \Checkmark & \Checkmark & \Checkmark & 92.32 & 90.99 \downr{-1.33} & 91.49 \downr{-0.84} & 89.53 \downr{-2.79} & 91.53 \downr{-0.8} & 91.73 \downr{-0.59} & 91.49 \downr{-0.84} & 92.93 \upg{0.6} & 93.06 \upg{0.74} & 96.25 \upg{3.93} \\ \midrule
  & \Checkmark & \Checkmark & \Checkmark & \Checkmark & 92.32 & 91.26 \downr{-1.06} & 91.56 \downr{-0.76} & 90.61 \downr{-1.72} & 91.62 \downr{-0.7} & 91.67 \downr{-0.65} & 91.45 \downr{-0.88} & 92.86 \upg{0.54} & 93.03 \upg{0.71} & 96.35 \upg{4.02} \\ \midrule
\Checkmark & \Checkmark & \Checkmark & \Checkmark & \Checkmark & 92.32 & 90.08 \downr{-2.25} & 90.88 \downr{-1.45} & 89.62 \downr{-2.7} & 91.09 \downr{-1.24} & 91.43 \downr{-0.9} & 90.91 \downr{-1.41} & 92.91 \upg{0.59} & 93.07 \upg{0.75} & 96.60 \upg{4.27} \\ \midrule

\multicolumn{5}{c}{Mean $\Delta$} & & -0.61 & -0.14 & -1.85 & 0.00 & -0.17 & -0.07 & 0.87 & 0.96 & 4.19 \\ 
\multicolumn{5}{c}{Max $\Delta$} & & 2.39 & 2.73 & 1.90 & 3.11 & 2.04 & 2.96 & 3.24 & 3.26 & 8.58 \\ 
\multicolumn{5}{c}{Min $\Delta$} & & -3.47 & -2.12 & -7.85 & -1.71 & -1.09 & -2.09 & 0.17 & 0.41 & 2.19 \\ 
\bottomrule
\end{tabular}
}

\caption{Our results on \Food{} dataset for all the possible combinations of combining the zero-shot predictions of CLIP backbones, which we group intro non-parametric and parametric techniques. Also, the best-performing single backbone (\Best) and the \Oracle performance. We present, for each combination of backbones, the improvement \upg{}, constancy \samey{} and deterioration \downr{} of accuracy performance for each method when we compare it against the \Best{} backbone. Mean, Max, and Min $\Delta$ summarize the difference in performance across methods and backbone combinations.}
\label{tab:all_possible_combinations_food}
\end{table*}

\begin{table*}[]
\centering
\Flowers \\
\qquad
\resizebox{0.9\textwidth}{!}{

\begin{tabular}{ccccccccccccccc}
\toprule
\multicolumn{2}{c}{ResNet} & \multicolumn{3}{c}{ViT} & \multirow{2}{*}{\Best}  & \multicolumn{4}{c}{Non-Parametric} & \multicolumn{4}{c}{Parametric} & \multirow{2}{*}{\Oracle}\\
\cmidrule(lr){1-2} \cmidrule(lr){3-5} \cmidrule(lr){7-10} \cmidrule(lr){11-14}
50 & 101 & B-32 & B-16 & L-14 & & \VoteOne & \VoteThree & \Confidence & \LogitAvg & \CalibratedConfidence & \CLogitAvg & \GAC & \NNC &  \\
\midrule

\Checkmark &   &   &   &   &  \multicolumn{10}{c}{\xfill{.1em} 66.12 \samey{0.00} \xfill{.1em}} \\ 
  & \Checkmark &   &   &   & \multicolumn{10}{c}{\xfill{.1em}  65.20 \samey{0.00} \xfill{.1em}} \\ 
  &   & \Checkmark &   &   & \multicolumn{10}{c}{\xfill{.1em}  66.48 \samey{0.00} \xfill{.1em}} \\ 
  &   &   & \Checkmark &   & \multicolumn{10}{c}{\xfill{.1em}  71.43 \samey{0.00} \xfill{.1em}} \\ 
  &   &   &   & \Checkmark & \multicolumn{10}{c}{\xfill{.1em}  79.05 \samey{0.00} \xfill{.1em}} \\ \midrule
 \Checkmark & \Checkmark &   &   &   & 66.12 & 68.01 \upg{1.89} & 67.95 \upg{1.82} & 67.36 \upg{1.24} & 68.63 \upg{2.5} & 67.25 \upg{1.12} & 68.43 \upg{2.31} & 68.40 \upg{2.28} & 68.84 \upg{2.72} & 73.83 \upg{7.71} \\ \midrule
\Checkmark &   & \Checkmark &   &   & 66.48 & 69.21 \upg{2.73} & 69.77 \upg{3.29} & 68.68 \upg{2.2} & 69.17 \upg{2.68} & 68.76 \upg{2.28} & 69.77 \upg{3.29} & 69.28 \upg{2.8} & 70.00 \upg{3.51} & 75.49 \upg{9.01} \\ \midrule
\Checkmark &   &   & \Checkmark &   & 71.43 & 72.45 \upg{1.02} & 72.68 \upg{1.25} & 72.03 \upg{0.6} & 72.48 \upg{1.06} & 71.87 \upg{0.44} & 72.69 \upg{1.27} & 72.42 \upg{0.99} & 73.12 \upg{1.69} & 76.63 \upg{5.2} \\ \midrule
\Checkmark &   &   &   & \Checkmark & 79.05 & 78.29 \downr{-0.76} & 78.22 \downr{-0.83} & 79.02 \downr{-0.03} & 77.33 \downr{-1.72} & 78.78 \downr{-0.28} & 78.19 \downr{-0.86} & 78.91 \downr{-0.15} & 79.41 \upg{0.36} & 83.27 \upg{4.21} \\ \midrule
  & \Checkmark & \Checkmark &   &   & 66.48 & 68.56 \upg{2.08} & 68.74 \upg{2.26} & 68.35 \upg{1.87} & 68.87 \upg{2.39} & 68.32 \upg{1.84} & 68.71 \upg{2.23} & 68.74 \upg{2.26} & 69.21 \upg{2.73} & 74.39 \upg{7.9} \\ \midrule
  & \Checkmark &   & \Checkmark &   & 71.43 & 71.98 \upg{0.55} & 72.32 \upg{0.89} & 71.70 \upg{0.28} & 72.35 \upg{0.93} & 71.41 \downr{-0.02} & 72.37 \upg{0.94} & 73.10 \upg{1.68} & 73.02 \upg{1.59} & 77.35 \upg{5.92} \\ \midrule
  & \Checkmark &   &   & \Checkmark & 79.05 & 77.87 \downr{-1.19} & 77.74 \downr{-1.32} & 78.34 \downr{-0.72} & 77.23 \downr{-1.82} & 77.87 \downr{-1.19} & 77.52 \downr{-1.53} & 77.83 \downr{-1.22} & 79.23 \upg{0.18} & 82.62 \upg{3.56} \\ \midrule
  &   & \Checkmark & \Checkmark &   & 71.43 & 72.91 \upg{1.48} & 72.69 \upg{1.27} & 65.15 \downr{-6.28} & 72.16 \upg{0.73} & 72.52 \upg{1.09} & 72.69 \upg{1.27} & 71.88 \upg{0.46} & 72.99 \upg{1.56} & 78.29 \upg{6.86} \\ \midrule
  &   & \Checkmark &   & \Checkmark & 79.05 & 78.50 \downr{-0.55} & 78.31 \downr{-0.75} & 78.45 \downr{-0.6} & 77.43 \downr{-1.63} & 78.35 \downr{-0.7} & 77.88 \downr{-1.17} & 78.94 \downr{-0.11} & 79.07 \upg{0.02} & 82.86 \upg{3.81} \\ \midrule
  &   &   & \Checkmark & \Checkmark & 79.05 & 78.01 \downr{-1.04} & 78.01 \downr{-1.04} & 78.45 \downr{-0.6} & 77.90 \downr{-1.15} & 78.37 \downr{-0.68} & 78.29 \downr{-0.76} & 78.14 \downr{-0.91} & 79.66 \upg{0.6} & 83.36 \upg{4.31} \\ \midrule
\Checkmark & \Checkmark & \Checkmark &   &   & 66.48 & 69.38 \upg{2.89} & 69.98 \upg{3.5} & 69.12 \upg{2.63} & 69.57 \upg{3.09} & 69.12 \upg{2.63} & 70.09 \upg{3.61} & 69.64 \upg{3.15} & 70.92 \upg{4.44} & 78.57 \upg{12.08} \\ \midrule
\Checkmark & \Checkmark &   & \Checkmark &   & 71.43 & 71.74 \upg{0.31} & 72.27 \upg{0.85} & 71.78 \upg{0.36} & 72.48 \upg{1.06} & 71.49 \upg{0.07} & 72.52 \upg{1.09} & 73.10 \upg{1.68} & 73.82 \upg{2.39} & 79.46 \upg{8.03} \\ \midrule
\Checkmark & \Checkmark &   &   & \Checkmark & 79.05 & 74.70 \downr{-4.36} & 76.03 \downr{-3.02} & 78.34 \downr{-0.72} & 75.59 \downr{-3.46} & 77.77 \downr{-1.28} & 75.95 \downr{-3.11} & 78.00 \downr{-1.06} & 79.90 \upg{0.85} & 84.65 \upg{5.59} \\ \midrule
\Checkmark &   & \Checkmark & \Checkmark &   & 71.43 & 72.69 \upg{1.27} & 73.20 \upg{1.77} & 66.29 \downr{-5.14} & 72.74 \upg{1.32} & 72.55 \upg{1.12} & 72.92 \upg{1.5} & 72.97 \upg{1.54} & 73.90 \upg{2.47} & 80.34 \upg{8.91} \\ \midrule
\Checkmark &   & \Checkmark &   & \Checkmark & 79.05 & 75.87 \downr{-3.19} & 76.89 \downr{-2.16} & 78.48 \downr{-0.57} & 76.34 \downr{-2.72} & 78.32 \downr{-0.73} & 76.74 \downr{-2.31} & 78.83 \downr{-0.23} & 80.08 \upg{1.02} & 84.91 \upg{5.85} \\ \midrule
\Checkmark &   &   & \Checkmark & \Checkmark & 79.05 & 76.84 \downr{-2.21} & 77.41 \downr{-1.64} & 78.48 \downr{-0.57} & 77.33 \downr{-1.72} & 78.35 \downr{-0.7} & 77.44 \downr{-1.61} & 78.09 \downr{-0.96} & 80.27 \upg{1.22} & 84.83 \upg{5.77} \\ \midrule
  & \Checkmark & \Checkmark & \Checkmark &   & 71.43 & 71.69 \upg{0.26} & 72.55 \upg{1.12} & 66.34 \downr{-5.09} & 72.09 \upg{0.67} & 72.43 \upg{1.01} & 72.82 \upg{1.4} & 71.74 \upg{0.31} & 73.83 \upg{2.41} & 80.94 \upg{9.51} \\ \midrule
  & \Checkmark & \Checkmark &   & \Checkmark & 79.05 & 75.15 \downr{-3.9} & 76.39 \downr{-2.67} & 77.92 \downr{-1.14} & 76.19 \downr{-2.86} & 77.67 \downr{-1.38} & 76.45 \downr{-2.6} & 78.53 \downr{-0.52} & 79.44 \upg{0.39} & 84.16 \upg{5.11} \\ \midrule
  & \Checkmark &   & \Checkmark & \Checkmark & 79.05 & 76.34 \downr{-2.72} & 77.13 \downr{-1.92} & 78.05 \downr{-1.01} & 77.20 \downr{-1.85} & 77.67 \downr{-1.38} & 77.36 \downr{-1.69} & 78.16 \downr{-0.89} & 80.16 \upg{1.11} & 84.88 \upg{5.82} \\ \midrule
  &   & \Checkmark & \Checkmark & \Checkmark & 79.05 & 77.04 \downr{-2.02} & 77.65 \downr{-1.4} & 76.11 \downr{-2.94} & 76.74 \downr{-2.31} & 78.11 \downr{-0.94} & 77.80 \downr{-1.25} & 78.57 \downr{-0.49} & 79.69 \upg{0.63} & 85.01 \upg{5.95} \\ \midrule
\Checkmark & \Checkmark & \Checkmark & \Checkmark &   & 71.43 & 72.52 \upg{1.09} & 72.95 \upg{1.53} & 67.02 \downr{-4.41} & 72.60 \upg{1.17} & 72.22 \upg{0.8} & 73.00 \upg{1.58} & 72.61 \upg{1.19} & 74.45 \upg{3.02} & 82.14 \upg{10.72} \\ \midrule
\Checkmark & \Checkmark & \Checkmark &   & \Checkmark & 79.05 & 75.22 \downr{-3.84} & 75.69 \downr{-3.37} & 77.92 \downr{-1.14} & 75.18 \downr{-3.87} & 77.59 \downr{-1.46} & 75.52 \downr{-3.53} & 78.63 \downr{-0.42} & 79.80 \upg{0.75} & 85.62 \upg{6.57} \\ \midrule
\Checkmark & \Checkmark &   & \Checkmark & \Checkmark & 79.05 & 76.01 \downr{-3.04} & 76.52 \downr{-2.54} & 78.00 \downr{-1.06} & 76.26 \downr{-2.8} & 77.57 \downr{-1.48} & 76.55 \downr{-2.5} & 78.24 \downr{-0.81} & 80.27 \upg{1.22} & 85.67 \upg{6.62} \\ \midrule
\Checkmark &   & \Checkmark & \Checkmark & \Checkmark & 79.05 & 76.71 \downr{-2.34} & 77.22 \downr{-1.84} & 76.21 \downr{-2.85} & 76.31 \downr{-2.75} & 78.14 \downr{-0.91} & 77.05 \downr{-2.0} & 78.18 \downr{-0.88} & 80.18 \upg{1.12} & 85.77 \upg{6.72} \\ \midrule
  & \Checkmark & \Checkmark & \Checkmark & \Checkmark & 79.05 & 76.81 \downr{-2.24} & 77.09 \downr{-1.97} & 75.80 \downr{-3.25} & 76.22 \downr{-2.83} & 77.69 \downr{-1.37} & 77.10 \downr{-1.95} & 78.19 \downr{-0.86} & 79.93 \upg{0.88} & 85.75 \upg{6.7} \\ \midrule
\Checkmark & \Checkmark & \Checkmark & \Checkmark & \Checkmark & 79.05 & 75.70 \downr{-3.35} & 76.28 \downr{-2.77} & 75.85 \downr{-3.20} & 75.54 \downr{-3.51} & 77.60 \downr{-1.45} & 76.25 \downr{-2.80} & 78.16 \downr{-0.89} & 81.10 \upg{2.05} & 86.32 \upg{7.27} \\ \midrule

\multicolumn{5}{c}{Mean $\Delta$} & & -0.81 & -0.37 & -1.24 & -0.75 & -0.14 & -0.35 & 0.30 & 1.57 & 6.76 \\ 
\multicolumn{5}{c}{Max $\Delta$} & & 2.89 & 3.50 & 2.63 & 3.09 & 2.63 & 3.61 & 3.15 & 4.44 & 12.08 \\ 
\multicolumn{5}{c}{Min $\Delta$} & & -4.36 & -3.37 & -6.28 & -3.87 & -1.48 & -3.53 & -1.22 & 0.02 & 3.56 \\ 
\bottomrule
\end{tabular}
}
\caption{Our results on \Flowers{} dataset for all the possible combinations of combining the zero-shot predictions of CLIP backbones, which we group intro non-parametric and parametric techniques. Also, the best-performing single backbone (\Best) and the \Oracle performance. We present, for each combination of backbones, the improvement \upg{}, constancy \samey{} and deterioration \downr{} of accuracy performance for each method when we compare it against the \Best{} backbone. Mean, Max, and Min $\Delta$ summarize the difference in performance across methods and backbone combinations.}
\label{tab:all_possible_combinations_flowers}
\end{table*}

\begin{table*}[]
\centering
\Imagenet \\
\qquad
\resizebox{0.9\textwidth}{!}{

\begin{tabular}{ccccccccccccccc}
\toprule
\multicolumn{2}{c}{ResNet} & \multicolumn{3}{c}{ViT} & \multirow{2}{*}{\Best}  & \multicolumn{4}{c}{Non-Parametric} & \multicolumn{4}{c}{Parametric} & \multirow{2}{*}{\Oracle}\\
\cmidrule(lr){1-2} \cmidrule(lr){3-5} \cmidrule(lr){7-10} \cmidrule(lr){11-14}
50 & 101 & B-32 & B-16 & L-14 & & \VoteOne & \VoteThree & \Confidence & \LogitAvg & \CalibratedConfidence & \CLogitAvg & \GAC & \NNC &  \\
\midrule

\Checkmark &   &   &   &   &  \multicolumn{10}{c}{\xfill{.1em} 59.84 \samey{0.00} \xfill{.1em}} \\ 
  & \Checkmark &   &   &   & \multicolumn{10}{c}{\xfill{.1em}  62.28 \samey{0.00} \xfill{.1em}} \\ 
  &   & \Checkmark &   &   & \multicolumn{10}{c}{\xfill{.1em}  63.35 \samey{0.00} \xfill{.1em}} \\ 
  &   &   & \Checkmark &   & \multicolumn{10}{c}{\xfill{.1em}  68.34 \samey{0.00} \xfill{.1em}} \\ 
  &   &   &   & \Checkmark & \multicolumn{10}{c}{\xfill{.1em}  75.54 \samey{0.00} \xfill{.1em}} \\ \midrule
\Checkmark & \Checkmark &   &   &   & 62.30 & 63.76 \upg{1.46} & 64.16 \upg{1.86} & 59.05 \downr{-3.25} & 64.61 \upg{2.31} & 63.22 \upg{0.91} & 64.47 \upg{2.17} & 64.71 \upg{2.41} & 65.14 \upg{2.84} & 70.75 \upg{8.45} \\ \midrule
\Checkmark &   & \Checkmark &   &   & 63.36 & 65.00 \upg{1.65} & 65.37 \upg{2.01} & 64.40 \upg{1.04} & 65.78 \upg{2.42} & 64.39 \upg{1.03} & 65.54 \upg{2.19} & 65.86 \upg{2.51} & 66.07 \upg{2.71} & 71.76 \upg{8.41} \\ \midrule
\Checkmark &   &   & \Checkmark &   & 68.34 & 68.47 \upg{0.12} & 68.72 \upg{0.37} & 68.15 \downr{-0.19} & 68.89 \upg{0.55} & 68.09 \downr{-0.26} & 68.85 \upg{0.51} & 69.60 \upg{1.26} & 69.86 \upg{1.52} & 74.51 \upg{6.17} \\ \midrule
\Checkmark &   &   &   & \Checkmark & 75.53 & 75.13 \downr{-0.4} & 75.27 \downr{-0.27} & 60.23 \downr{-15.3} & 75.09 \downr{-0.45} & 74.51 \downr{-1.02} & 74.97 \downr{-0.56} & 75.97 \upg{0.44} & 76.17 \upg{0.64} & 80.13 \upg{4.6} \\ \midrule
  & \Checkmark & \Checkmark &   &   & 63.36 & 66.03 \upg{2.67} & 66.41 \upg{3.05} & 60.29 \downr{-3.07} & 66.99 \upg{3.63} & 65.45 \upg{2.09} & 66.70 \upg{3.35} & 67.00 \upg{3.65} & 67.35 \upg{3.99} & 72.96 \upg{9.61} \\ \midrule
  & \Checkmark &   & \Checkmark &   & 68.34 & 69.09 \upg{0.75} & 69.29 \upg{0.95} & 68.54 \upg{0.2} & 69.58 \upg{1.24} & 68.45 \upg{0.11} & 69.40 \upg{1.05} & 69.90 \upg{1.56} & 70.29 \upg{1.95} & 75.15 \upg{6.8} \\ \midrule
  & \Checkmark &   &   & \Checkmark & 75.53 & 75.18 \downr{-0.35} & 75.35 \downr{-0.18} & 62.53 \downr{-13.0} & 75.16 \downr{-0.38} & 74.56 \downr{-0.97} & 75.10 \downr{-0.44} & 76.01 \upg{0.48} & 76.39 \upg{0.86} & 80.28 \upg{4.75} \\ \midrule
  &   & \Checkmark & \Checkmark &   & 68.34 & 68.90 \upg{0.56} & 69.17 \upg{0.82} & 63.17 \downr{-5.18} & 69.41 \upg{1.07} & 68.48 \upg{0.13} & 69.23 \upg{0.89} & 69.76 \upg{1.42} & 70.04 \upg{1.7} & 74.97 \upg{6.63} \\ \midrule
  &   & \Checkmark &   & \Checkmark & 75.53 & 75.37 \downr{-0.16} & 75.40 \downr{-0.13} & 63.53 \downr{-12.01} & 75.19 \downr{-0.35} & 74.79 \downr{-0.74} & 75.20 \downr{-0.33} & 75.98 \upg{0.45} & 76.31 \upg{0.78} & 80.49 \upg{4.96} \\ \midrule
  &   &   & \Checkmark & \Checkmark & 75.53 & 75.47 \downr{-0.06} & 75.61 \upg{0.08} & 68.42 \downr{-7.12} & 75.63 \upg{0.1} & 75.04 \downr{-0.5} & 75.50 \downr{-0.03} & 75.98 \upg{0.44} & 76.26 \upg{0.73} & 80.60 \upg{5.06} \\ \midrule
\Checkmark & \Checkmark & \Checkmark &   &   & 63.36 & 66.15 \upg{2.79} & 66.76 \upg{3.4} & 60.27 \downr{-3.08} & 67.33 \upg{3.97} & 65.51 \upg{2.16} & 67.08 \upg{3.72} & 67.47 \upg{4.11} & 67.87 \upg{4.51} & 76.39 \upg{13.03} \\ \midrule
\Checkmark & \Checkmark &   & \Checkmark &   & 68.34 & 68.22 \downr{-0.12} & 68.94 \upg{0.6} & 66.82 \downr{-1.53} & 69.23 \upg{0.89} & 68.17 \downr{-0.18} & 69.18 \upg{0.83} & 70.11 \upg{1.77} & 70.71 \upg{2.37} & 77.95 \upg{9.61} \\ \midrule
\Checkmark & \Checkmark &   &   & \Checkmark & 75.53 & 72.15 \downr{-3.38} & 74.09 \downr{-1.45} & 62.70 \downr{-12.83} & 74.22 \downr{-1.31} & 74.03 \downr{-1.5} & 74.08 \downr{-1.45} & 76.07 \upg{0.54} & 76.42 \upg{0.88} & 82.41 \upg{6.87} \\ \midrule
\Checkmark &   & \Checkmark & \Checkmark &   & 68.34 & 68.58 \upg{0.24} & 69.20 \upg{0.85} & 63.72 \downr{-4.62} & 69.45 \upg{1.11} & 68.35 \samey{0.0} & 69.45 \upg{1.11} & 70.18 \upg{1.84} & 70.60 \upg{2.26} & 78.06 \upg{9.72} \\ \midrule
\Checkmark &   & \Checkmark &   & \Checkmark & 75.53 & 72.80 \downr{-2.73} & 74.35 \downr{-1.18} & 63.26 \downr{-12.28} & 74.36 \downr{-1.17} & 74.24 \downr{-1.3} & 74.32 \downr{-1.21} & 76.14 \upg{0.61} & 76.35 \upg{0.82} & 82.61 \upg{7.08} \\ \midrule
\Checkmark &   &   & \Checkmark & \Checkmark & 75.53 & 73.95 \downr{-1.58} & 74.93 \downr{-0.61} & 67.33 \downr{-8.21} & 74.98 \downr{-0.55} & 74.58 \downr{-0.95} & 74.90 \downr{-0.63} & 76.16 \upg{0.63} & 76.54 \upg{1.01} & 82.76 \upg{7.23} \\ \midrule
  & \Checkmark & \Checkmark & \Checkmark &   & 68.34 & 69.15 \upg{0.81} & 69.73 \upg{1.39} & 62.46 \downr{-5.88} & 70.02 \upg{1.68} & 68.66 \upg{0.32} & 69.95 \upg{1.61} & 70.43 \upg{2.09} & 70.87 \upg{2.53} & 78.58 \upg{10.24} \\ \midrule
  & \Checkmark & \Checkmark &   & \Checkmark & 75.53 & 73.32 \downr{-2.21} & 74.61 \downr{-0.92} & 63.05 \downr{-12.49} & 74.63 \downr{-0.9} & 74.27 \downr{-1.27} & 74.58 \downr{-0.95} & 76.10 \upg{0.57} & 76.60 \upg{1.07} & 82.82 \upg{7.29} \\ \midrule
  & \Checkmark &   & \Checkmark & \Checkmark & 75.53 & 74.13 \downr{-1.4} & 75.04 \downr{-0.49} & 67.39 \downr{-8.14} & 75.11 \downr{-0.42} & 74.59 \downr{-0.94} & 75.14 \downr{-0.39} & 76.09 \upg{0.56} & 76.73 \upg{1.2} & 82.89 \upg{7.35} \\ \midrule
  &   & \Checkmark & \Checkmark & \Checkmark & 75.53 & 74.01 \downr{-1.52} & 74.90 \downr{-0.63} & 68.56 \downr{-6.97} & 75.02 \downr{-0.51} & 74.64 \downr{-0.9} & 74.98 \downr{-0.55} & 76.07 \upg{0.54} & 76.55 \upg{1.01} & 82.91 \upg{7.38} \\ \midrule
\Checkmark & \Checkmark & \Checkmark & \Checkmark &   & 68.34 & 69.10 \upg{0.75} & 69.51 \upg{1.17} & 62.51 \downr{-5.83} & 69.66 \upg{1.32} & 68.43 \upg{0.08} & 69.74 \upg{1.4} & 70.52 \upg{2.18} & 71.09 \upg{2.75} & 80.31 \upg{11.97} \\ \midrule
\Checkmark & \Checkmark & \Checkmark &   & \Checkmark & 75.53 & 72.37 \downr{-3.16} & 73.71 \downr{-1.82} & 63.11 \downr{-12.42} & 73.67 \downr{-1.86} & 73.94 \downr{-1.6} & 73.62 \downr{-1.92} & 76.12 \upg{0.58} & 76.69 \upg{1.16} & 84.08 \upg{8.55} \\ \midrule
\Checkmark & \Checkmark &   & \Checkmark & \Checkmark & 75.53 & 73.47 \downr{-2.06} & 74.46 \downr{-1.07} & 67.47 \downr{-8.06} & 74.46 \downr{-1.07} & 74.21 \downr{-1.32} & 74.41 \downr{-1.12} & 76.10 \upg{0.57} & 76.63 \upg{1.09} & 84.17 \upg{8.64} \\ \midrule
\Checkmark &   & \Checkmark & \Checkmark & \Checkmark & 75.53 & 73.54 \downr{-1.99} & 74.40 \downr{-1.13} & 67.54 \downr{-7.99} & 74.39 \downr{-1.14} & 74.32 \downr{-1.21} & 74.39 \downr{-1.14} & 76.14 \upg{0.61} & 76.65 \upg{1.11} & 84.26 \upg{8.73} \\ \midrule
  & \Checkmark & \Checkmark & \Checkmark & \Checkmark & 75.53 & 73.78 \downr{-1.75} & 74.64 \downr{-0.89} & 67.46 \downr{-8.07} & 74.58 \downr{-0.95} & 74.32 \downr{-1.21} & 74.59 \downr{-0.94} & 76.19 \upg{0.66} & 76.77 \upg{1.23} & 84.42 \upg{8.89} \\ \midrule
\Checkmark & \Checkmark & \Checkmark & \Checkmark & \Checkmark & 75.53 & 72.67 \downr{-2.86} & 73.95 \downr{-1.58} & 67.46 \downr{-8.07} & 73.89 \downr{-1.64} & 74.06 \downr{-1.47} & 73.88 \downr{-1.65} & 76.22 \upg{0.69} & 76.59 \upg{1.06} & 85.29 \upg{9.76} \\ \midrule

\multicolumn{5}{c}{Mean $\Delta$} & & -0.54 & 0.16 & -7.09 & 0.29 & -0.40 & 0.21 & 1.28 & 1.68 & 7.99 \\ 
\multicolumn{5}{c}{Max $\Delta$} & & 2.79 & 3.40 & 1.04 & 3.97 & 2.16 & 3.72 & 4.11 & 4.51 & 13.03 \\ 
\multicolumn{5}{c}{Min $\Delta$} & & -3.38 & -1.82 & -15.30 & -1.86 & -1.60 & -1.92 & 0.44 & 0.64 & 4.60 \\ 
\bottomrule
\end{tabular}
}
\caption{Our results on \Imagenet{} dataset for all the possible combinations of combining the zero-shot predictions of CLIP backbones, which we group intro non-parametric and parametric techniques. Also, the best-performing single backbone (\Best) and the \Oracle{} performance. We present, for each combination of backbones, the improvement \upg{}, constancy \samey{} and deterioration \downr{} of accuracy performance for each method when we compare it against the \Best{} backbone. Mean, Max, and Min $\Delta$ summarize the difference in performance across methods and backbone combinations.}
\label{tab:all_possible_combinations_inet}
\end{table*}

\section{Venn diagrams LinearProbe CLIP}
\begin{figure*}[ht!]
    \centering
    \underline{\textbf{\Cars}}
    \includegraphics[width=0.99\textwidth]{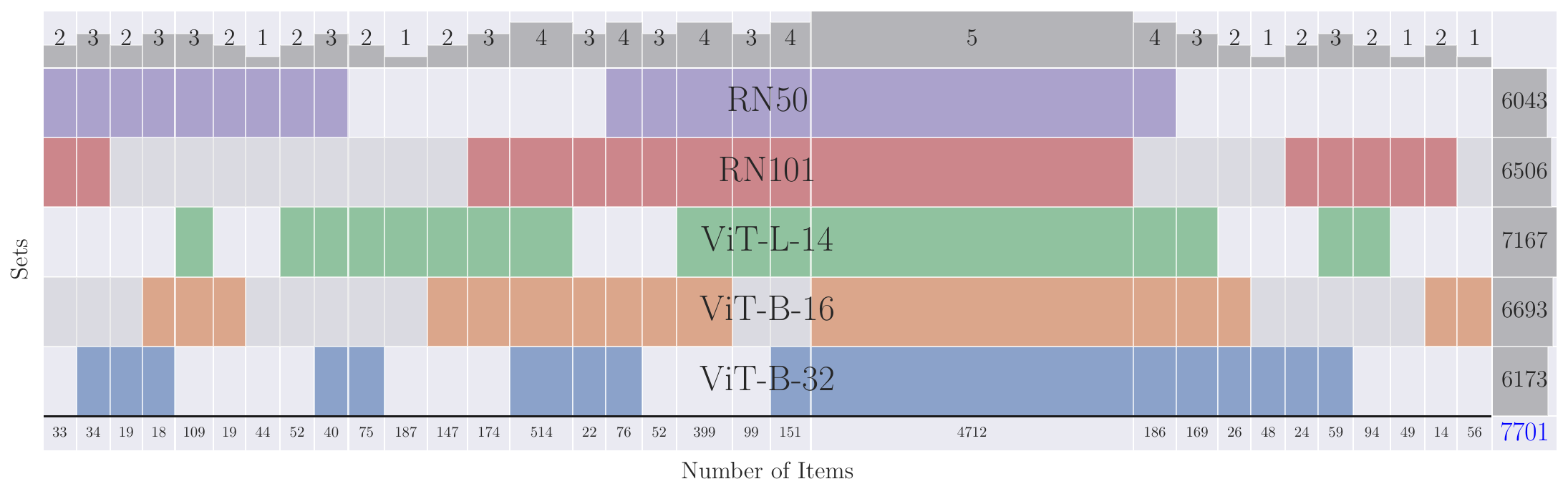}
    \underline{\textbf{\CUB}}
    \includegraphics[width=0.99\textwidth]{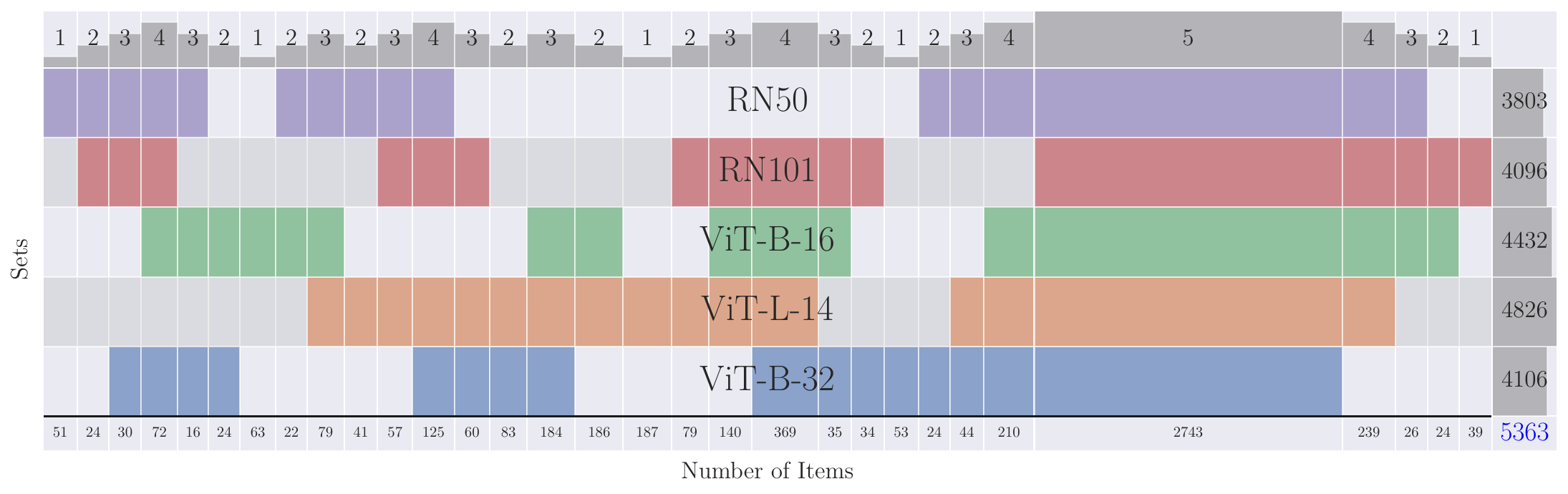}
    \underline{\textbf{\DTD}}
    \includegraphics[width=0.99\textwidth]{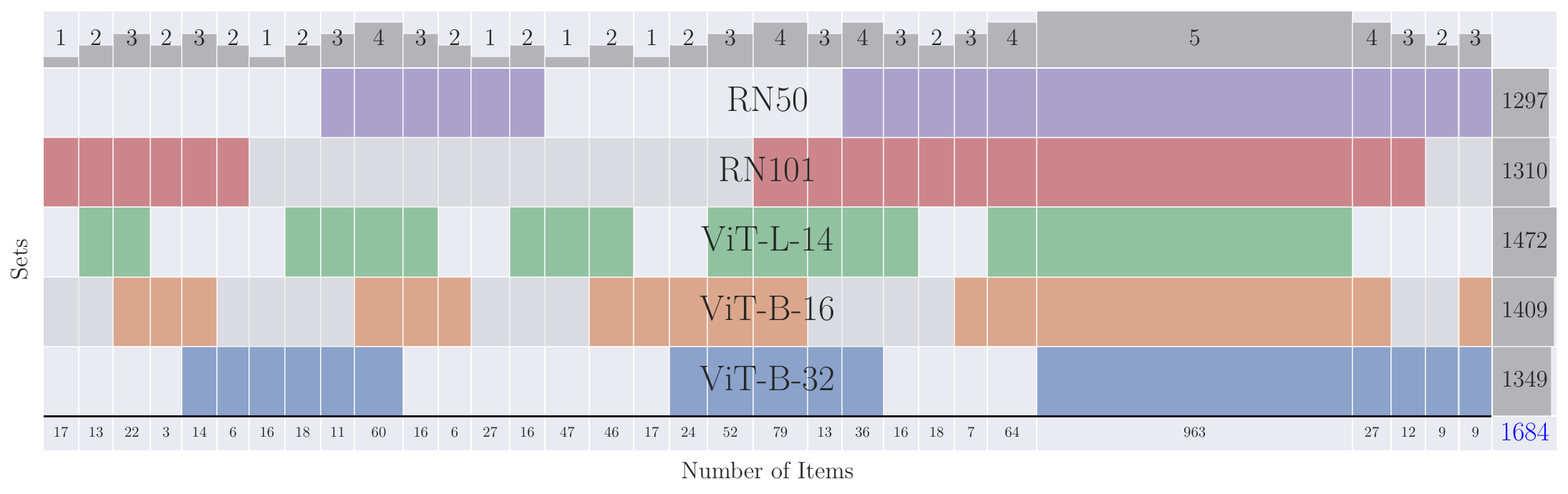}\vspace{-3mm}
    \caption{Venn diagram with the correct prediction of each backbone. The Top part of the Venn diagram shows the number of backbones that are predicting correctly a set of images. Each column represents a set of image instances that are predicted correctly by some group of backbones. Each row in the diagram shows in colour the backbone that correctly predicts a certain set of image instances, in grey when the backbone is not correctly predicting those instances. The bottom part of the Venn diagram shows the number of images in a certain set. The right part is the total amount of correctly predicted images per backbone.}
    \label{fig:venn_diagram_1_lp}
    \vspace{-4mm}
\end{figure*}

\begin{figure*}[ht!]
    \centering
    \underline{\textbf{\FGVC}}
    \includegraphics[width=0.99\textwidth]{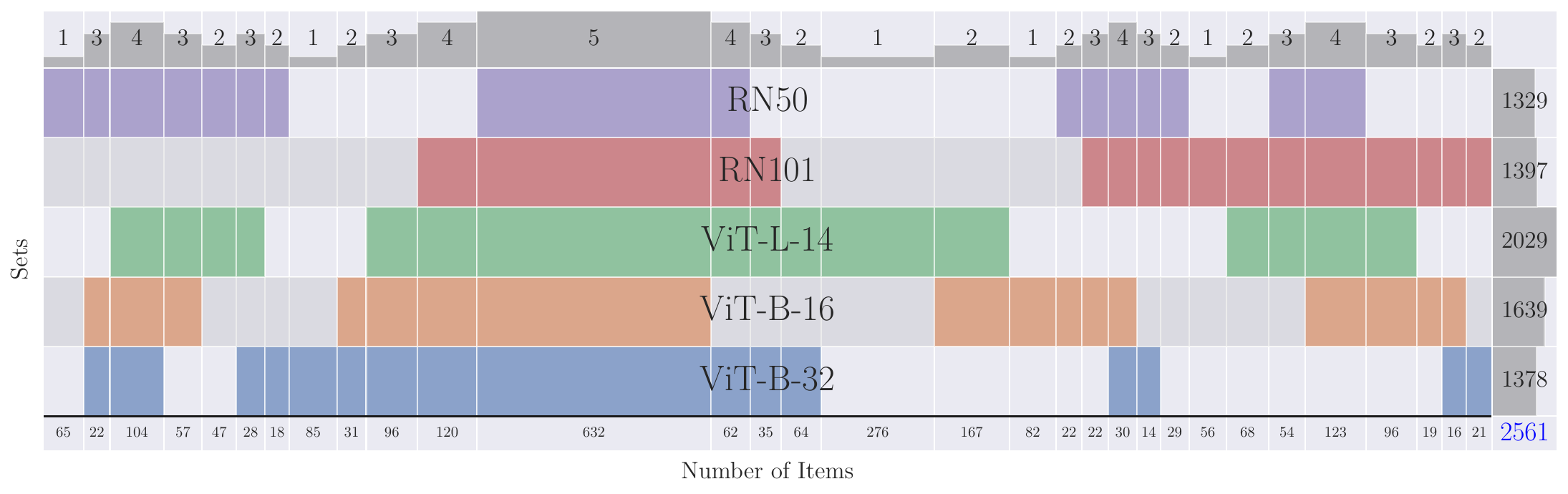}
    \underline{\textbf{\Flowers}}
    \includegraphics[width=0.99\textwidth]{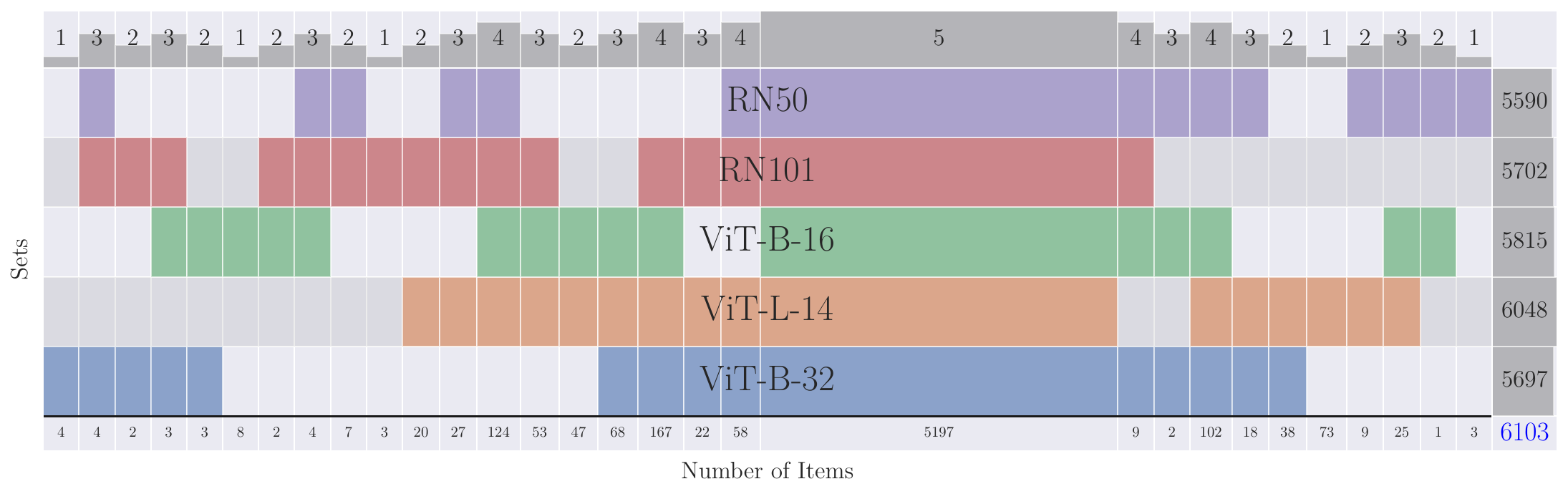}
    \underline{\textbf{\Food}}
    \includegraphics[width=0.99\textwidth]{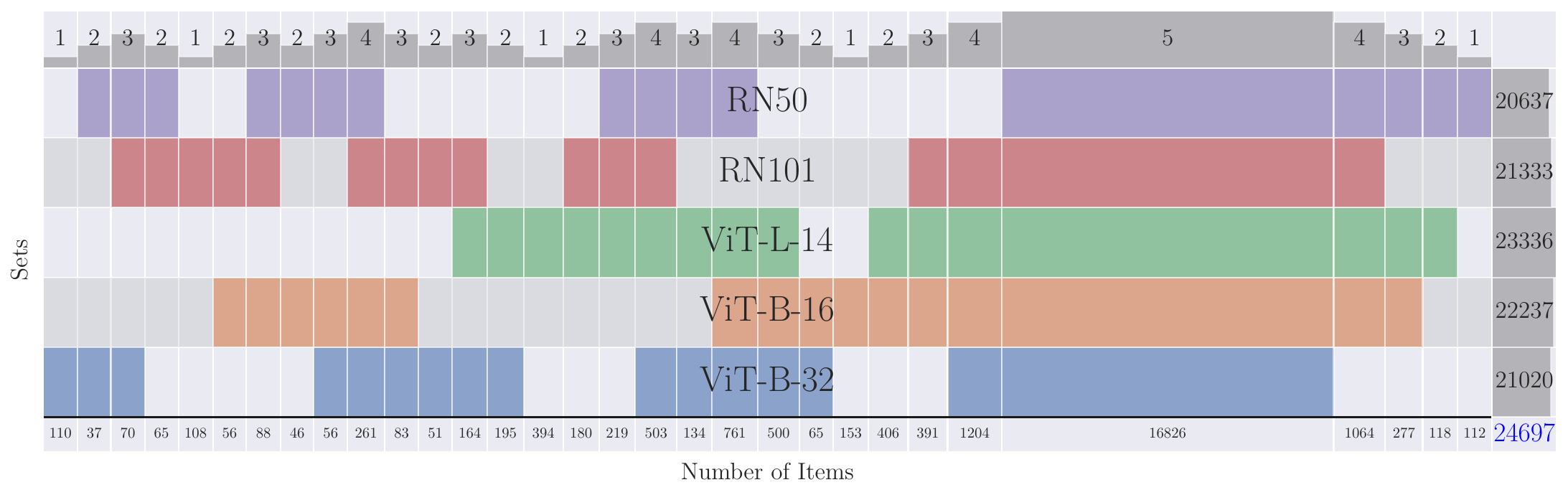}\vspace{-3mm}
    \caption{Venn diagram with the correct prediction of each backbone. The Top part of the Venn diagram shows the number of backbones that are predicting correctly a set of images. Each column represents a set of image instances that are predicted correctly by some group of backbones. Each row in the diagram shows in colour the backbone that correctly predicts a certain set of image instances, in grey when the backbone is not correctly predicting those instances. The bottom part of the Venn diagram shows the number of images in a certain set. The right part is the total amount of correctly predicted images per backbone.}
    \label{fig:venn_diagram_2_lp}
    \vspace{-4mm}
\end{figure*}

\begin{figure*}[ht!]
    \centering
    \underline{\textbf{\Pets}}
    \includegraphics[width=0.99\textwidth]{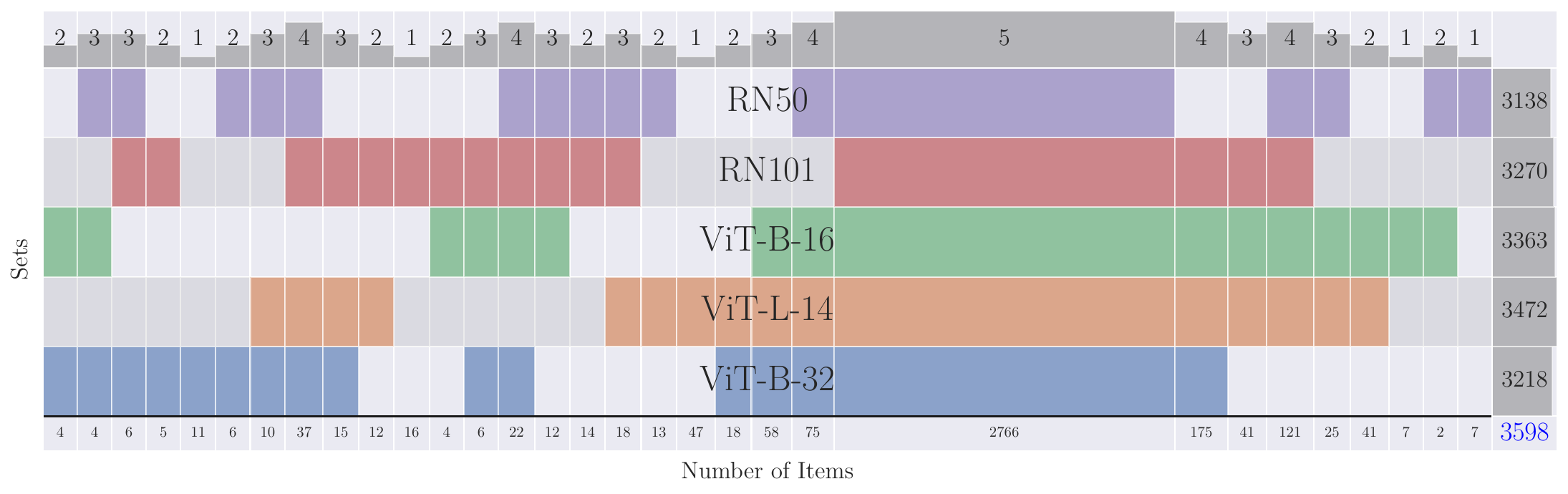}
    \underline{\textbf{\Imagenet}}
    \includegraphics[width=0.99\textwidth]{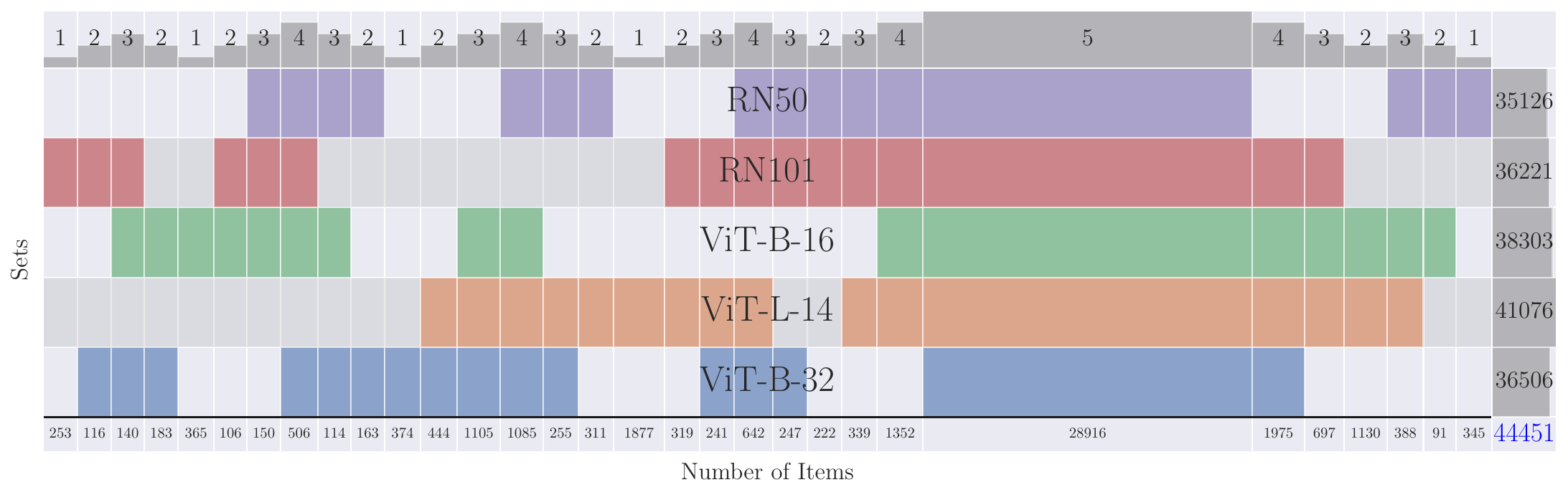}\vspace{-3mm}
    \caption{Venn diagram with the correct prediction of each backbone. The Top part of the Venn diagram shows the number of backbones that are predicting correctly a set of images. Each column represents a set of image instances that are predicted correctly by some group of backbones. Each row in the diagram shows in colour the backbone that correctly predicts a certain set of image instances, in grey when the backbone is not correctly predicting those instances. The bottom part of the Venn diagram shows the number of images in a certain set. The right part is the total amount of correctly predicted images per backbone.}
    \label{fig:venn_diagram_3_lp}
    \vspace{-4mm}
\end{figure*}

In Figures \ref{fig:venn_diagram_1_lp}, \ref{fig:venn_diagram_2_lp}, and \ref{fig:venn_diagram_3_lp}, the Venn diagrams for the LinearProbe versions of CLIP are presented. Similar to the ZeroShot CLIP, the diversity of predictions among backbones is preserved, albeit with a comparatively smaller potential for improvement. In comparison to the cases discussed in Section \ref{sec:venn_zs_clip}, \Cars{}, \CUB{}, \DTD{}, and \FGVC{} exhibit a higher overall agreement between different backbones in the LinearProbe context. Specifically, \Cars{} demonstrates a 61.19\% agreement, a notable increase from the 36.43\% observed in the ZeroShot case. Similarly, \CUB{} showcases a 51.34\% agreement in the LinearProbe case, compared to 31.71\% in the ZeroShot case. For \DTD{}, the LinearProbe case sees a 57.19\% agreement, a substantial improvement from the 34.07\% observed in the ZeroShot case. In the case of \FGVC{}, there is a 24.68\% agreement in the LinearProbe case, contrasting with the 10.83\% observed in the ZeroShot case. However, despite this increased agreement, there remains a notable scope for improvement, underscoring the continued orthogonality of predictions across different backbones in the LinearProbe versions.

\section{Alpha values}
\begin{figure*}
\centering
\begin{subfigure}{0.33\textwidth}
    \includegraphics[width=\textwidth]{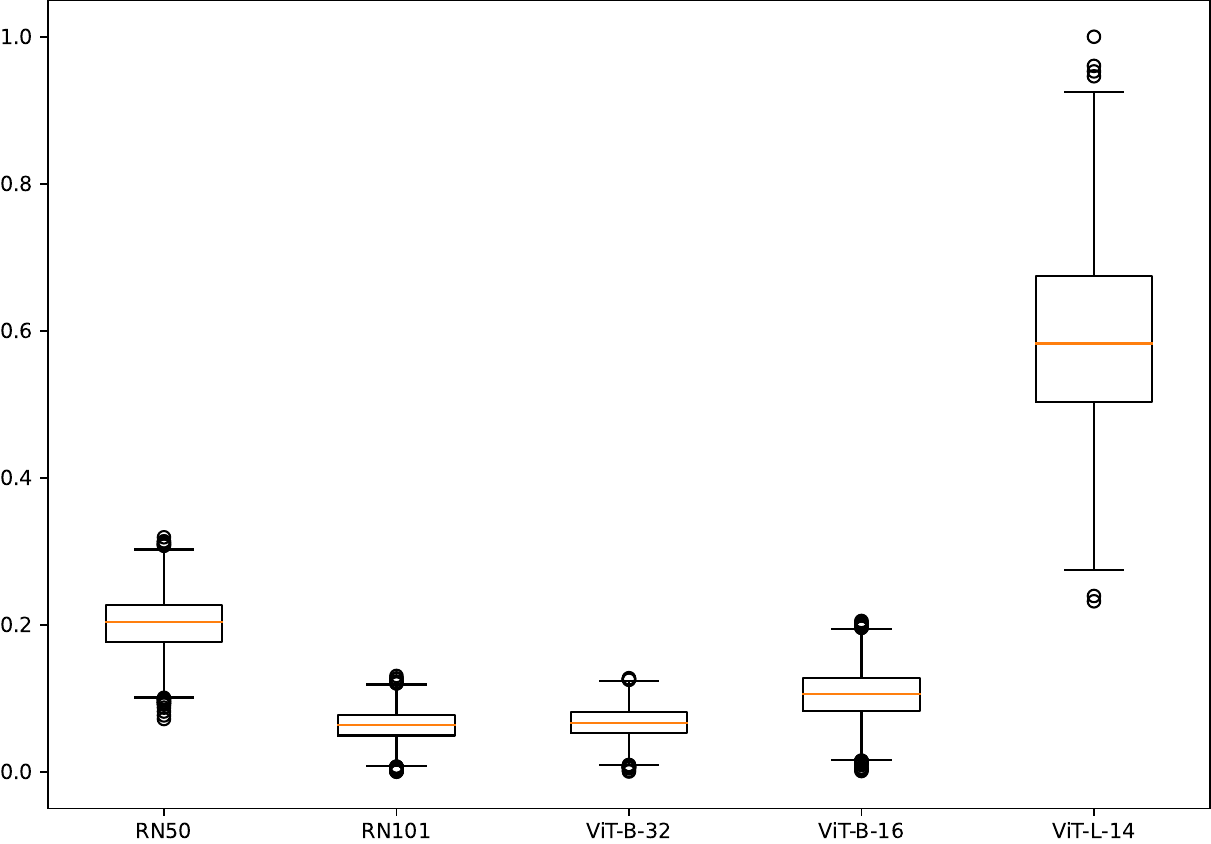}
    \caption{\Pets{}}
    \label{fig:alpha_box_plots_pets}
\end{subfigure}
\hfill
\begin{subfigure}{0.33\textwidth}
    \includegraphics[width=\textwidth]{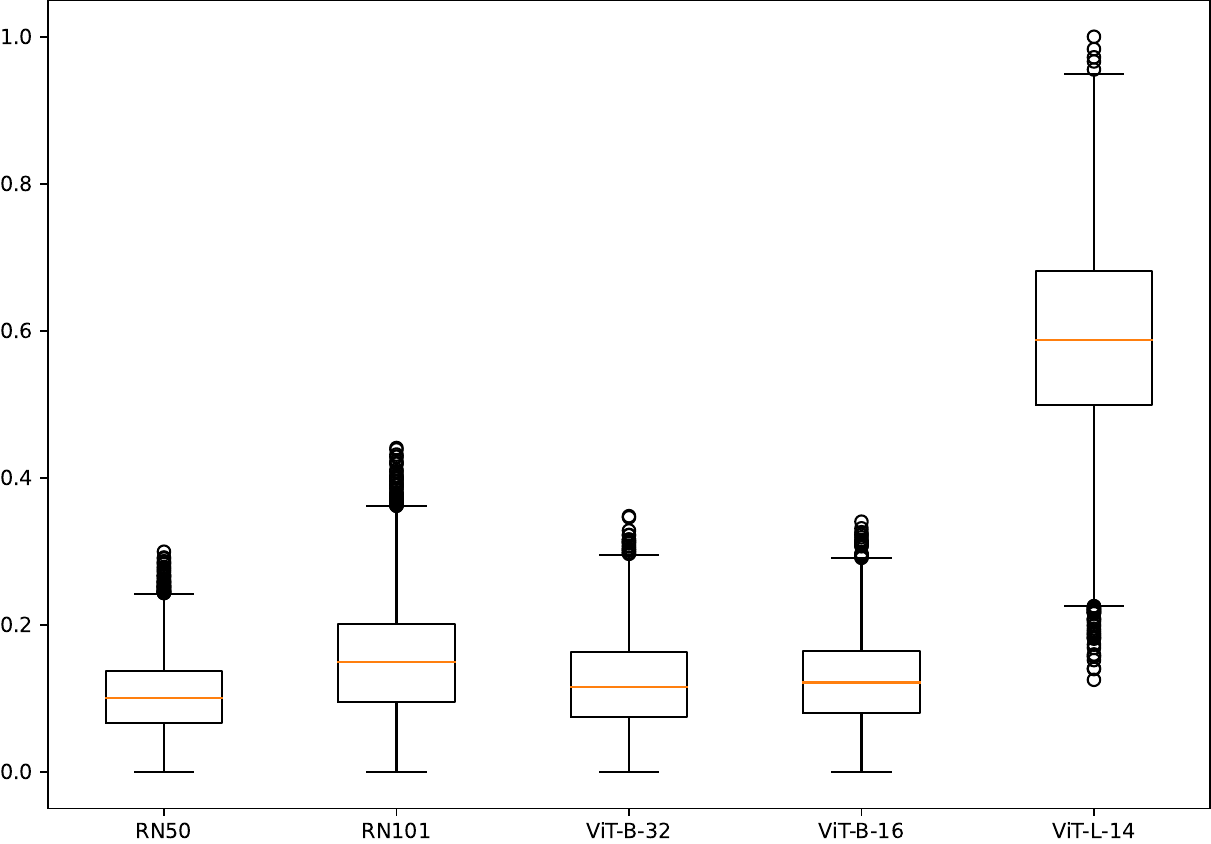}
    \caption{\Cars}
    \label{fig:alpha_box_plots_cars}
\end{subfigure}
\hfill
\begin{subfigure}{0.33\textwidth}
    \includegraphics[width=\textwidth]{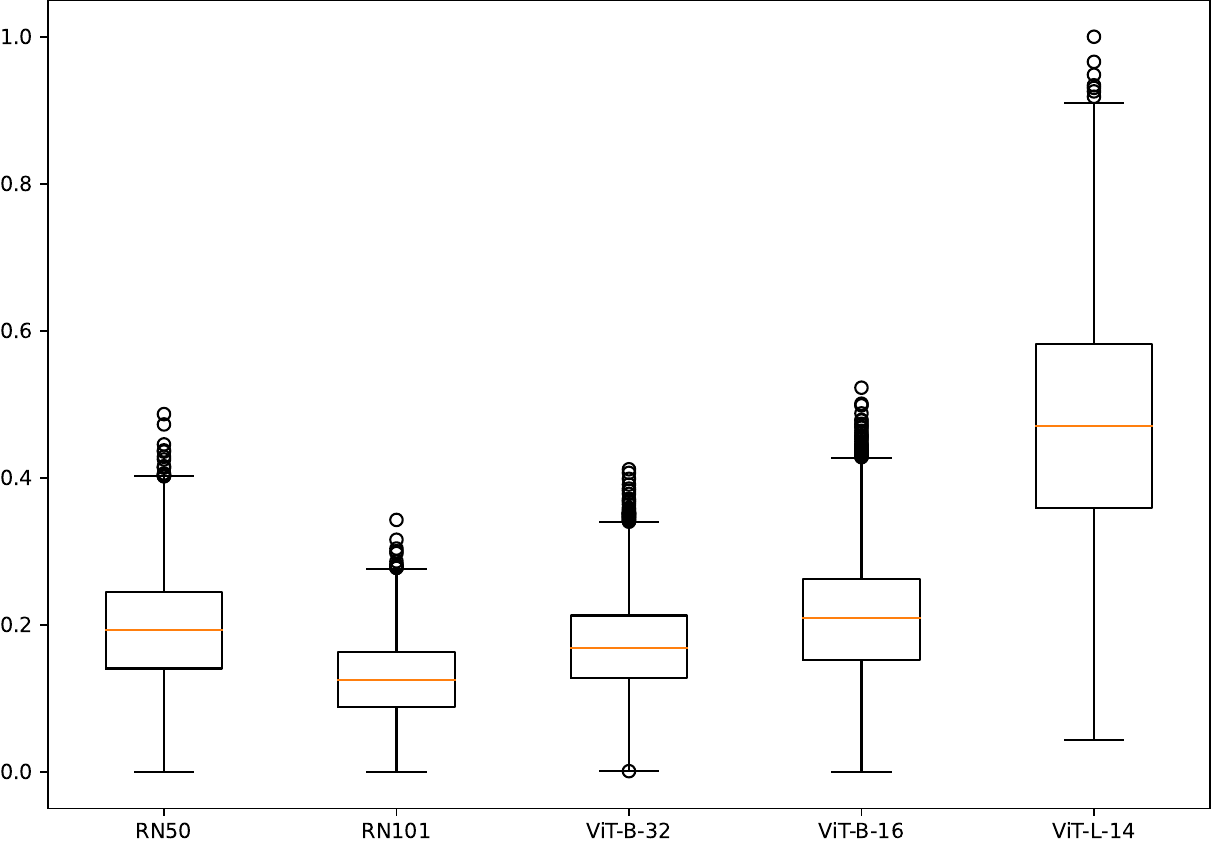}
    \caption{\CUB}
    \label{fig:alpha_box_plots_cub}
\end{subfigure}
\hfill
\begin{subfigure}{0.33\textwidth}
    \includegraphics[width=\textwidth]{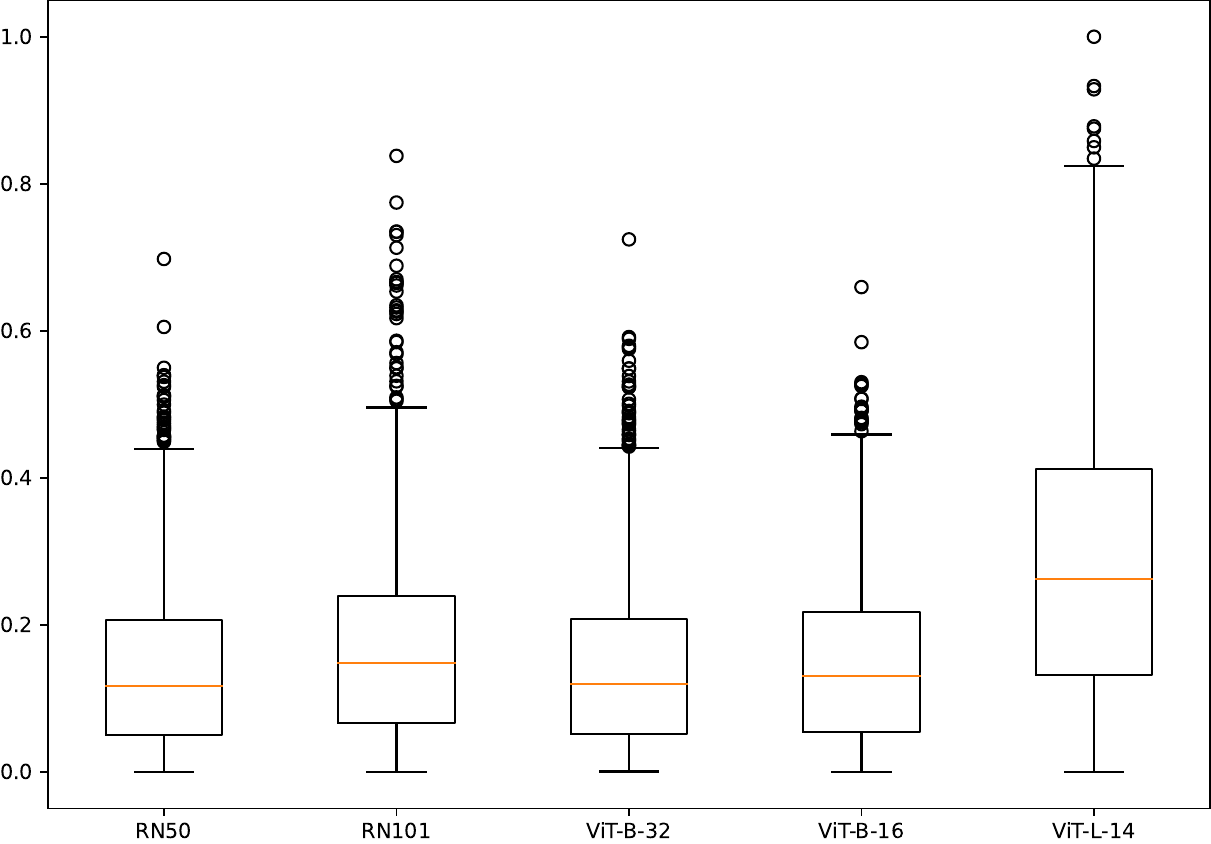}
    \caption{\DTD}
    \label{fig:alpha_box_plots_dtd}
\end{subfigure}
\hfill       
\begin{subfigure}{0.33\textwidth}
    \includegraphics[width=\textwidth]{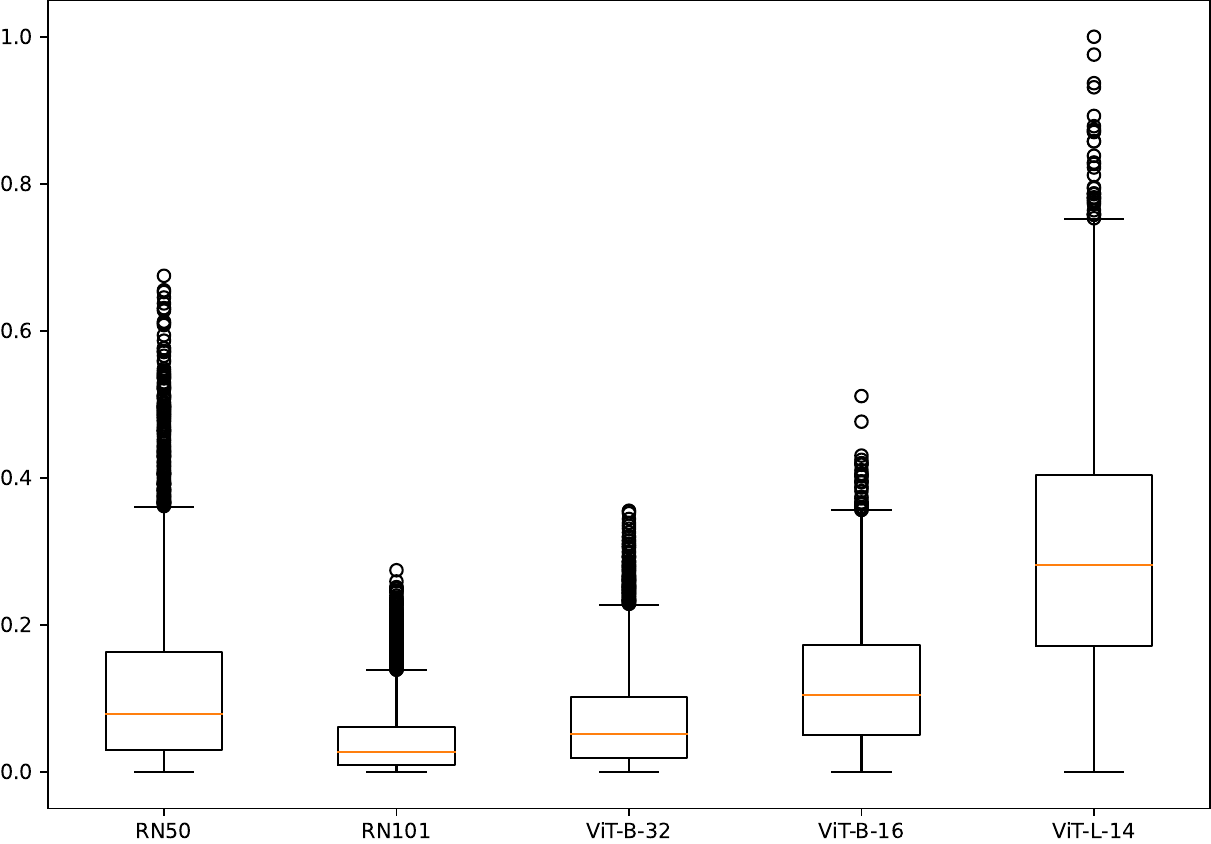}
    \caption{\FGVC{}}
    \label{fig:alpha_box_plots_fgvc}
\end{subfigure}
\hfill
\begin{subfigure}{0.33\textwidth}
    \includegraphics[width=\textwidth]{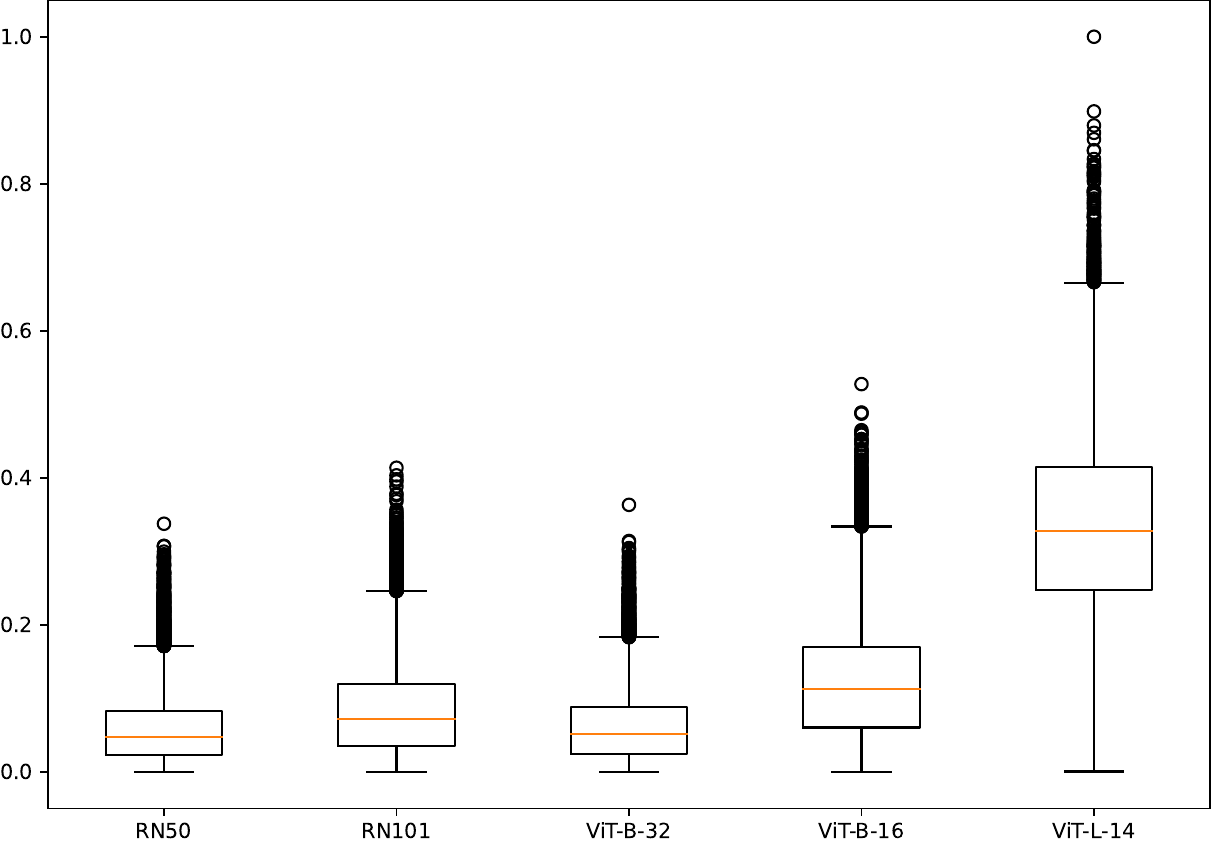}
    \caption{\Food}
    \label{fig:alpha_box_plots_food}
\end{subfigure}
\hfill
\begin{subfigure}{0.33\textwidth}
    \includegraphics[width=\textwidth]{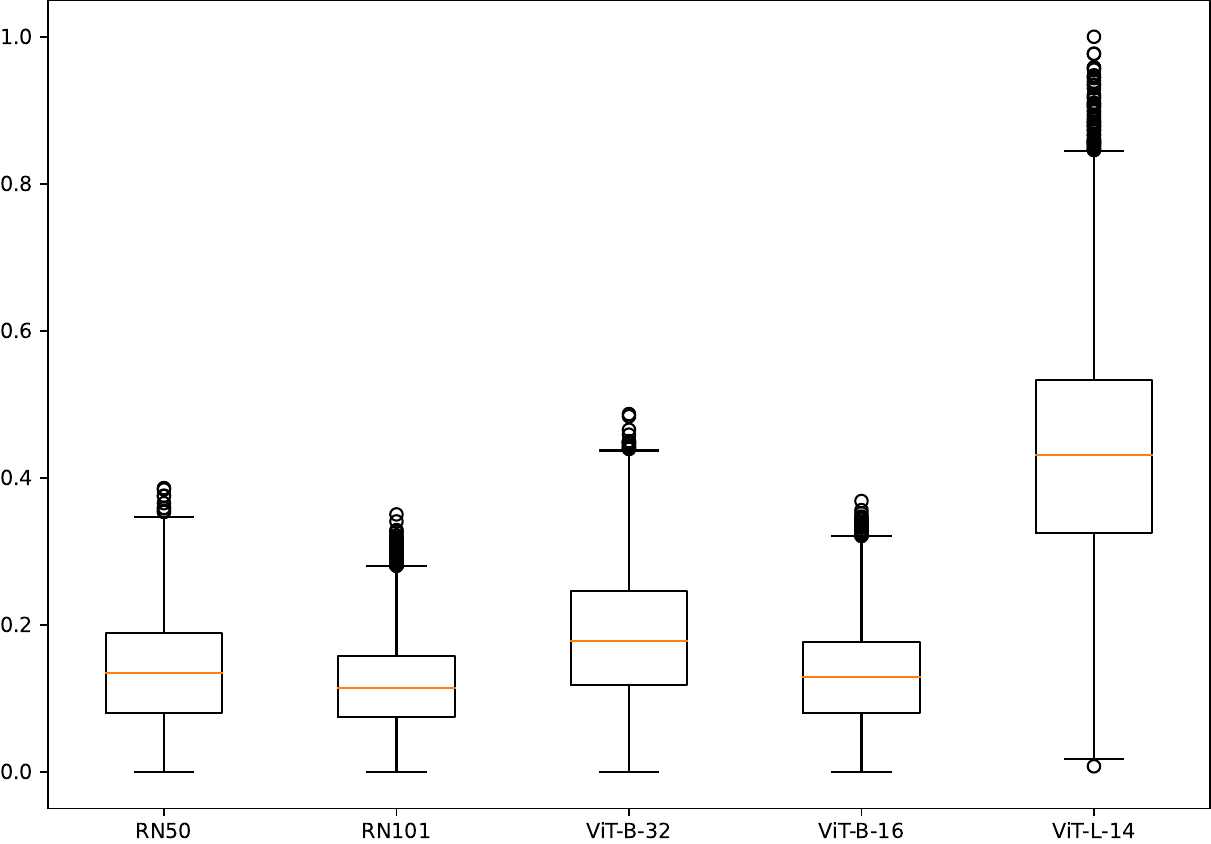}
    \caption{\Flowers}
    \label{fig:alpha_box_plots_flowers}
\end{subfigure}
\begin{subfigure}{0.33\textwidth}
    \includegraphics[width=\textwidth]{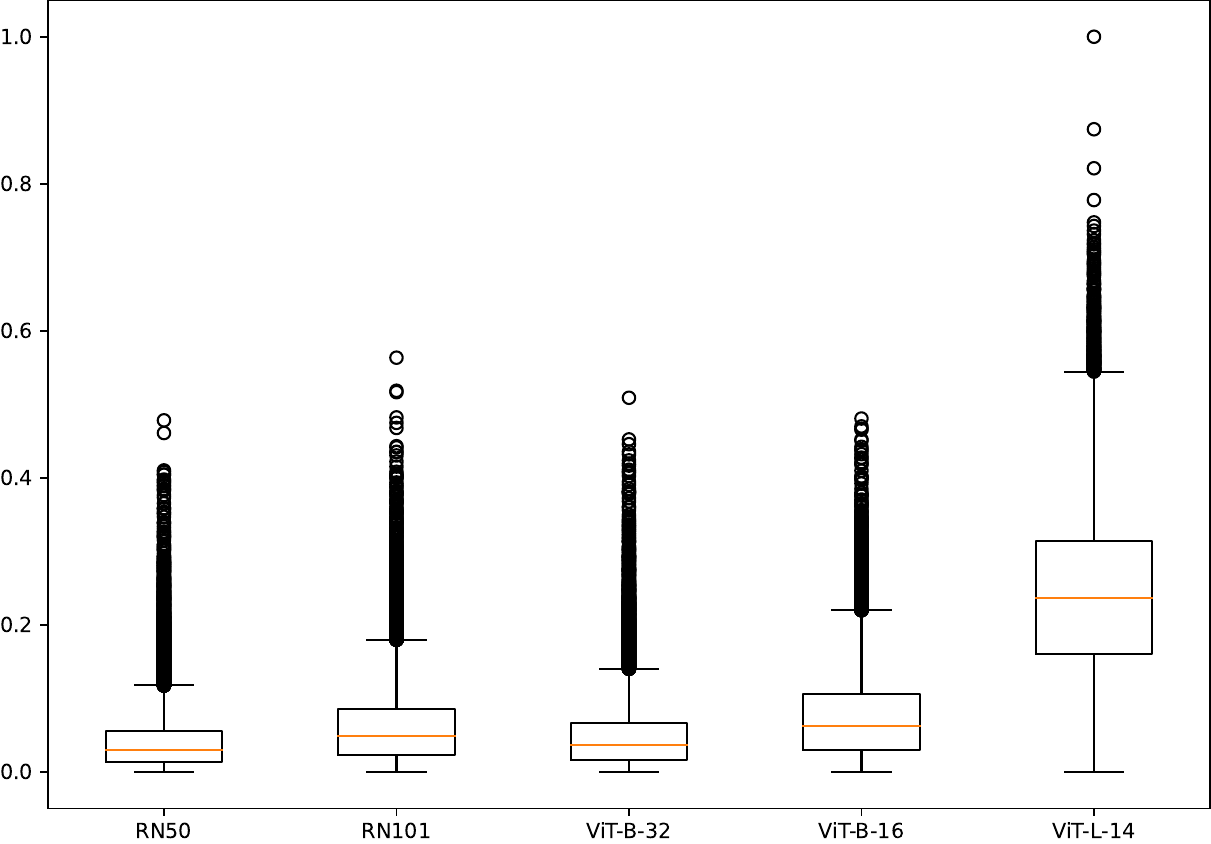}
    \caption{\Imagenet}
    \label{fig:alpha_box_plots_inet}
\end{subfigure}
\caption{Alpha values for each dataset using \NNC}
\label{fig:alphavalues_box_lots}
\end{figure*}

In Figure \ref{fig:alphavalues_box_lots}, we present the distribution of alpha values using a box plot for the \NNC{} method, normalized by their maximum value. Across all datasets, there is a notable dominance of ViT-L-14 in the weight distribution, particularly prominent in the \Pets{} and \Cars{} benchmarks, where ViT-L-14, being the largest backbone in the ViT family, holds significant influence. Interestingly, it is observed that the most weighted backbone within the ResNet family is not consistently ResNet-101, despite its deeper architecture. This observation is evident in datasets such as \Pets{}, \CUB{}, \FGVC{}, and \Flowers{}, where the mean value of alpha corresponding to ResNet-50 surpasses that of ResNet-101.

Furthermore, the distribution of alpha weights across backbones for datasets such as \Imagenet{}, \FGVC{}, and \DTD{} is more uniform compared to other datasets. This suggests that the \NNC{} method is effectively leveraging the strengths of each backbone to arrive at accurate labels for each sample, resulting in a more balanced distribution of weights across different backbones.
\end{document}